\documentclass[times,twocolumn,final]{elsarticle}

\usepackage{medima}
\usepackage{framed,multirow}

\usepackage{amssymb}
\usepackage{latexsym}
\usepackage{graphicx}
\usepackage{amsmath}
\usepackage{booktabs}
\usepackage{subcaption}
\usepackage{soul}
\usepackage{pifont}

\usepackage{float}

\usepackage{url}
\usepackage[table,xcdraw]{xcolor}
\newcommand{\xmark}{--}

\usepackage{pdfrender}
\newcommand*{\boldcheckmark}{%
  \textpdfrender{
    TextRenderingMode=FillStroke,
    LineWidth=.5pt, 
  }{\checkmark}%
}

\usepackage{hyperref}
\usepackage[normalem]{ulem}

\definecolor{newcolor}{rgb}{.8,.349,.1}

\begin{document}


\begin{frontmatter}


\title{Pixel-Wise Recognition for Holistic Surgical Scene Understanding}%

\author[1]{Nicolás \snm{Ayobi}\corref{cor1}}
\ead{n.ayobi@uniandes.edu.co}
\author[1]{Santiago \snm{Rodríguez} \fnref{fn1}}
\author[1]{Alejandra \snm{Pérez} \fnref{fn1}} 
\author[1]{Isabela \snm{Hernández} \fnref{fn1}}
\author[1]{Nicolás \snm{Aparicio}}
\author[1]{Eugénie \snm{Dessevres}}
\author[2]{Sebastián \snm{Peña}}
\author[2]{Jessica \snm{Santander}}
\author[2]{Juan Ignacio \snm{Caicedo}}
\author[3,4]{Nicolás \snm{Fernández}}
\author[1]{Pablo \snm{Arbeláez} \corref{cor1}}
\cortext[cor1]{Corresponding authors: Nicolás Ayobi and Pablo Arbeláez} 
\ead{pa.arbelaez@uniandes.edu.co}
\fntext[fn1]{Equal contribution}


\address[1]{Center for Research and Formation in Artificial Intelligence (CinfonIA), Universidad de los Andes, Carrera 1 No. 18a-12, 111711 Bogota, Colombia}
\address[2]{Department of Urology, Fundación Santa Fe de Bogotá, Carrera 7 No. 118 - 09, 111071 Bogota, Colombia}
\address[3]{Division of Urology, Seattle Children's Hospital, 4800 Sand Point Way NE, 98105 Seattle, Washington, USA}
\address[4]{Department of Urology, University of Washington, 1410 NE Campus Pkwy, 98195 Seattle, Washington, USA}

\begin{abstract}{}
This paper presents the Holistic and Multi-Granular Surgical Scene Understanding of Prostatectomies (GraSP) dataset, a curated benchmark that models surgical scene understanding as a hierarchy of complementary tasks with varying levels of granularity. Our approach encompasses long-term tasks, such as surgical phase and step recognition, and short-term tasks, including surgical instrument segmentation and atomic visual actions detection. To exploit our proposed benchmark, we introduce the Transformers for Actions, Phases, Steps, and Instrument Segmentation (TAPIS) model, a general architecture that combines a global video feature extractor with localized region proposals from an instrument segmentation model to tackle the multi-granularity of our benchmark. Through extensive experimentation in ours and alternative benchmarks, we demonstrate TAPIS's versatility and state-of-the-art performance across different tasks. This work represents a foundational step forward in Endoscopic Vision, offering a novel framework for future research towards holistic surgical scene understanding. \\

\end{abstract}

\end{frontmatter}


\section{Introduction} \begin{figure*}
    \centering
    \includegraphics[width=\linewidth]{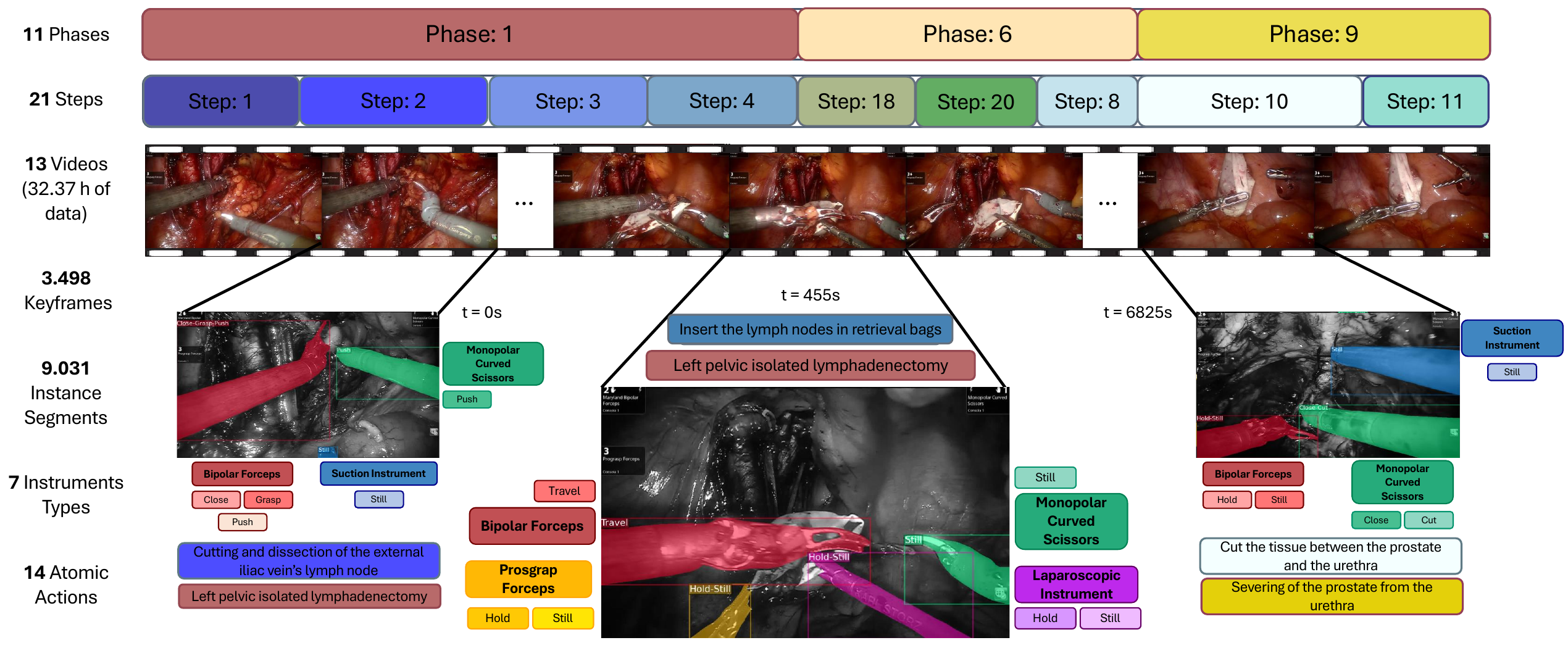}
    \caption{\textbf{The GraSP Dataset} formulates a holistic understanding of robot-assisted Radical Prostatectomy videos by studying four hierarchical tasks annotated in their highest granularity. These tasks include two long-term tasks, recognition of surgical phases and steps, and two short-term tasks, surgical instrument segmentation and atomic action detection. Figure best viewed in color. }
    \label{fig:grasp}
\end{figure*}

Robot-assisted surgery (RAS) has emerged as a major advancement in modern surgical practices~(\cite{goh2022rasevolution}), offering a minimally invasive alternative to open surgeries and providing enhanced precision over conventional laparoscopic techniques~(\cite{chochulo2023rasbetter}). 
Its enhanced instrumentation, visualization, and dexterity greatly benefit surgeons and significantly reduce patient complications~(\cite{lanfrance2004roboticsurgery,amsterdam2022rarp45}). 
Despite these benefits, robotic surgery faces several limitations, including limited automated assistance~(\cite{fiorini2021automated}), minimal image-based force feedback~(\cite{rassi2020haptic}), and a steep learning curve for specific surgeries~(\cite{Kim2019learningcurve,Vining2021learningcurve}). 
Nevertheless, the extensive data generated by surgical robots provide opportunities to develop data-based solutions that enrich the robot with surgical cognition and overcome current challenges~(\cite{surgical-data-science}). 

A crucial step towards exploiting the potential of surgical robotic systems is understanding the visual data captured by the robot's endoscope~(\cite{Mascagni2022cvsurgery}). Albeit, to fully interpret endoscopic data, it is necessary to go beyond simple object identification and adopt a holistic approach that thoroughly models the temporal and spatial dimensions of surgical videos. Such advanced understanding would enhance robotic systems by enabling augmented reality systems~(\cite{long2020augmentedreality, Tanzi2021augmentedreality}), context-aware assistance~(\cite{katic2013contextaware, kolbinger2023contextaware}), improved surgical training~(\cite{surgical-data-science}), and development of semi-automated surgical interventions.

In light of this, Endoscopic Vision (EV) is the area within Surgical Data Science that studies the essential components of image-guided surgical cognition like surgical environment geometry~(\cite{wang2022endonerf}), surgical procedural knowledge~(\cite{wagner2023heichole}), and tools-tissues interactions~(\cite{islam2020interactions,cholecT40}). However, each methodological approach in EV prioritizes different visual cues that lead to different requirements in visual recognition, temporal awareness, and spatial granularity. For example, surgical workflow analysis (SWA) studies tasks such as surgical phase recognition to understand the temporal progression of surgical procedures~(\cite{padoy2021surgicalworkflow}). Nevertheless, this objective implicitly requires a multirecognition methodology that identifies different agents and activities over time in surgical settings~(\cite{endonet-cholec80,nwoye2021rendezvous}). Consequently, SWA has expanded to multitask approaches that include recognizing surgical steps, gestures, actions, or instruments ~(\cite{demir2023surgicalworkflow}).

Despite the advancements in multi-task benchmarks for SWA, there remains a significant gap in the literature concerning a holistic understanding of the multi-level nature of surgical interventions. A fundamental limitation of most current SWA benchmarks is their constrained study of the spatial dimensions of endoscopic images. Specifically, current benchmarks often disregard the visual localization of surgical instruments, which are the constitutive acting agents in any endoscopic surgery~(\cite{nwoye2021rendezvous}). 
Instead, surgical instrument segmentation or detection tasks have mostly been treated as independent areas from SWA~(\cite{jin2018tooldetection}), and very few endoscopic benchmarks integrate these tasks with SWA due to the time-consuming nature of gathering such annotations.  
Nonetheless, integrating the visual locations of surgical instruments is primordial for robot-assisted surgery, as current robotic systems can track instrument usage and kinematics but cannot link these signals to the instruments' visual locations~(\cite{du2016tracking,su2018haptic,Guan2023multimode}).
On top of that, instrument segmentation, as the finest-grained level of visual recognition, would allow surgical robots to understand instrument shapes~(\cite{baby2023s3net}), poses~(\cite{du2018pose}), and visual boundaries~(\cite{garcia2017toolnet}).

A second area for improvement in current SWA benchmarks concerns the exploitation of the hierarchical complementarity among different visual perception tasks. 
Initially, some benchmarks treat single recognition tasks such as phases recognition~(\cite{stauder2016lapchole}) or action recognition~(\cite{bawa2021saras_esad,sharghi2021avos}) as isolated tasks rather than interdependent components of a cohesive surgical process. 
Similarly, multi-task benchmarks focus on a few levels of visual understanding, like one video-level task with several frame-level presence tasks~(\cite{endonet-cholec80, nwoye2021rendezvous}), or multiple video-level tasks with no spatial localization~(\cite{misaw}), rather than tackling multiple levels in the visual, temporal, and spatial dimensions.

\begin{figure}[]
    \centering
    \includegraphics[width=\linewidth]{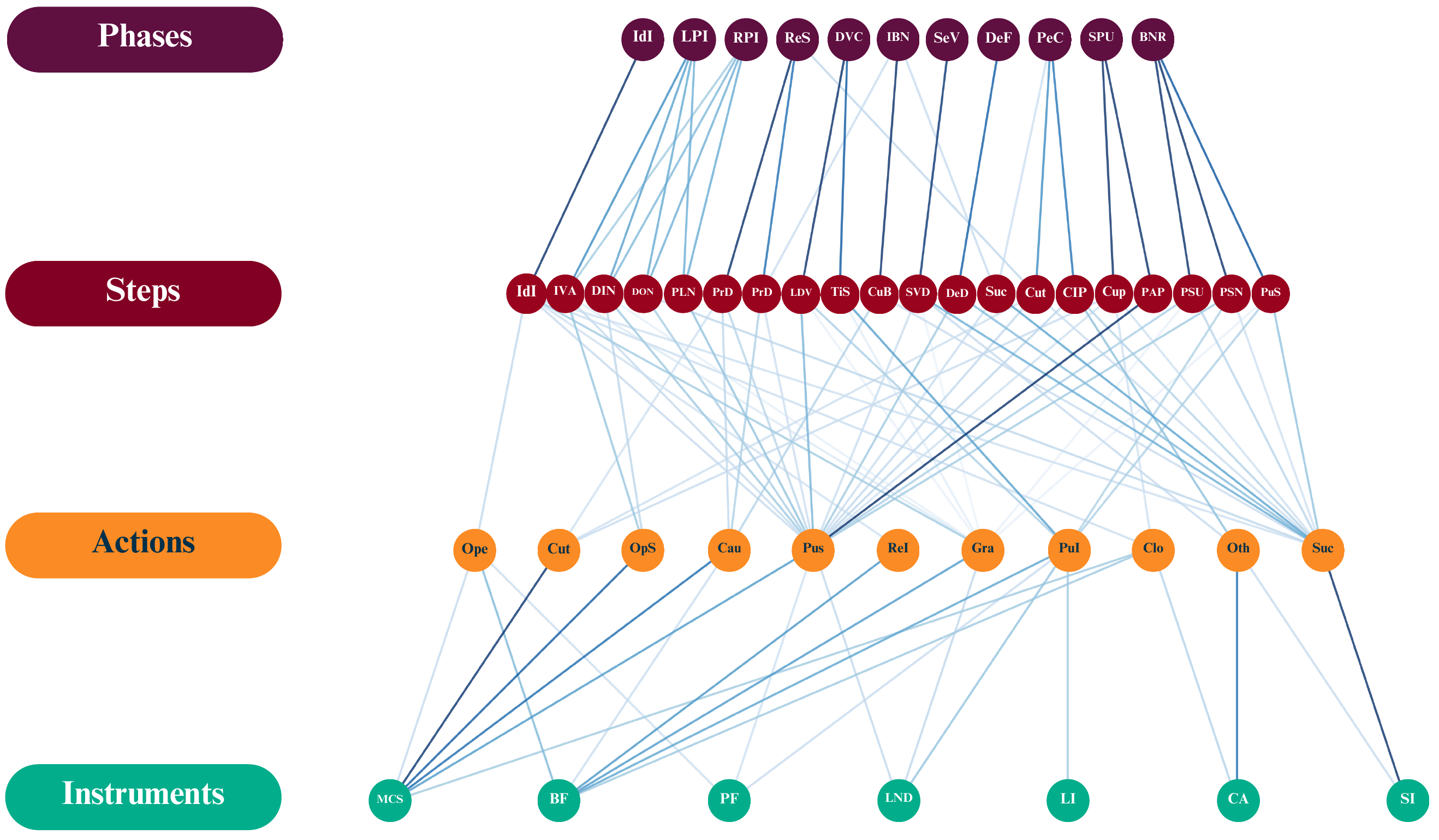}
    \caption{\textbf{Hierarchical Structure of GraSP Tasks.} This figure illustrates the hierarchical organization of tasks in GraSP based on temporal and spatial granularity. Nodes represent category labels within each task level, while edges denote connections between categories across levels. The intensity of the edge color reflects the co-occurrence of the linked categories based on annotations from GraSP. To enhance visualization, we reorganized the categories and removed weakly connected edges, as well as highly persistent action categories.}
    \label{fig:hierarchy}
\end{figure}

In our initial work~(\cite{valderrama2022tapir}), we laid the foundation for studying surgical scenes across multiple temporal and spatial granularities by modeling surgical procedures as a hierarchy of long-term video-level tasks, such as phase and step recognition, and short-term spatio-temporal tasks including instrument and atomic visual action localization. We base our modeling on the Atomic Visual Actions formulation in \cite{AVA_2018} and the activity theory of \cite{barker1954midwest}, which define activity as a hierarchy starting from its finest expression as atomic interactions performed by acting agents, to the broader development of complex objectives. Our work adapts these concepts into surgical workflow analysis with a stratification of semantic tasks where surgical phase recognition represents the most general level of understanding. These phases can be further decomposed into finer-grained steps, where each step involves performing specific atomic actions that require using particular surgical instruments. We introduced surgical atomic actions as a spatially localized approach to surgical action recognition and the finest temporal modeling of surgical activities. We portray this hierarchical structure in Figure~\ref{fig:hierarchy}. By breaking down the procedure into these distinct yet interconnected levels, we emphasize the importance of understanding the individual components and their collective dynamics, capturing the essence of both the macro and micro-elements of surgeries.

In this work, we extend our original framework by integrating an instrument instance segmentation task to achieve the highest level of spatial granularity. Additionally, we augment our dataset with five new fully annotated videos, curate all our annotations, and define specific data splits and metrics for benchmarking. Our extension in data accounts for more than a 55\% increase in data size and annotation over the initial PSI-AVA dataset~(\cite{valderrama2022tapir}). We call this advanced framework the Holistic and Multi-\textbf{Gra}nular \textbf{S}urgical Scene Understanding of \textbf{P}rostatectomies (GraSP) dataset. These improvements make our benchmark the first publicly available in the robotic surgery domain to include phase and step annotations and an instrument instance segmentation task using real human in-patient surgical data. To our knowledge, GraSP is the first EV framework to consider multiple long and short-term tasks in their most fine-grained form. 

Besides the extensions to our benchmark, we propose the Transformers for Actions, Phases, Steps, and Instrument Segmentation (TAPIS) model. We introduce TAPIS as a versatile, fully transformer-based architecture designed to tackle each task in the GraSP benchmark. Its flexibility allows it to function as both a multi-task framework, leveraging complementary tasks with varying granularities, or as a modular system capable of handling disjoint tasks independently. This adaptability enables TAPIS to seamlessly adapt to diverse endoscopic vision datasets and accommodate a wide range of video recognition tasks. 
TAPIS builds upon our previous methodologies for holistic surgical scene understanding~(\cite{valderrama2022tapir}) and segmentation~(\cite{ayobi2023matis}) by incorporating an instrument segmentation baseline that provides region proposals and shape-wise region features that enhance instrument and action recognition. We also introduce an improved \textit{region classification head} that exploits transformers' cross-attention to fully integrate temporal and spatial embeddings. Through comparisons with alternative baselines and SOTA models on GraSP and publicly available benchmarks, we demonstrate that TAPIS consistently achieves state-of-the-art performance across diverse tasks. Our efforts in dataset extension, methodological innovation, and benchmark validation, contribute to surgical data science by paving the way towards a holistic understanding of surgical procedures.

To summarize, the main contributions of our work are:
\begin{itemize}
    \item We propose the GraSP dataset for holistic surgical scene understanding, a curated and augmented version of the PSI-AVA benchmark with defined benchmark splits.
    \item We introduce a new instrument instance segmentation task into our benchmark.
    \item We present a generalized, transformer-based architecture that strongly outperforms alternative baselines across all tasks of the GraSP framework.
\end{itemize}

For the sake of fairness and reproducibility and to promote further research in holistic and multi-granular surgical scene understanding, we make public the entire GraSP dataset along with our source codes and pretrained models under the MIT License in \url{https://github.com/BCV-Uniandes/GraSP}. 

\label{intro}

\section{Related Work}\renewcommand{\arraystretch}{1.2}
\begin{table*}[t]
\caption{\textbf{Our GraSP benchmark compared with current surgical workflow analysis and endoscopic vision datasets.} Each row represents a current Endoscopic Vision dataset, and each column is a possible dataset attribute corresponding to either the inclusion of multiple levels of semantic comprehension, annotations of a particular surgical scene understanding the task, video duration, public availability, or data source type. We show whether an attribute is present (\checkmark), absent (–), has a different formulation (*), or its information is unavailable (?). We also indicate if the source of the provided data is from traditional Laparoscopic Surgery (LS), \textit{ex vivo} Surgical Training Procedures (ST), Robot-Assisted Surgery on animal subjects (RASA), or Robot-Assisted Surgery on human subjects (RASH). Our GraSP dataset, shown in bold, is the first publicly available benchmark to provide all the presented attributes. Note: We combine the Instrument Detection and Instrument Segmentation attributes to identify datasets with instance-level annotations. Datasets with both attributes have Instrument Instance Segmentation annotations, datasets with only Instrument Detection have bounding box annotations, and datasets with only Instrument Segmentation annotations have Semantic Segmentation annotations.}
\resizebox{\textwidth}{!}{
\begin{tabular}{lccccccccccc}
\hline
Dataset & \begin{tabular}[c]{@{}c@{}}Multi-level\\ Understanding\end{tabular} & \begin{tabular}[c]{@{}c@{}}Phase\\ Recognition\end{tabular} & \begin{tabular}[c]{@{}c@{}}Step\\ Recognition\end{tabular} & \begin{tabular}[c]{@{}c@{}}Instrument\\ Recognition\end{tabular} & \begin{tabular}[c]{@{}c@{}}Instrument\\ Detection\end{tabular} & \begin{tabular}[c]{@{}c@{}}Instrument\\ Segmentation\end{tabular} & \begin{tabular}[c]{@{}c@{}}Action\\ Recognition\end{tabular} & \begin{tabular}[c]{@{}c@{}}Action \\ Localization\end{tabular} & \begin{tabular}[c]{@{}c@{}}Video\\ hours\end{tabular} & \begin{tabular}[c]{@{}c@{}}Publicly\\ Available\end{tabular} & \begin{tabular}[c]{@{}c@{}}Data\\ Domain\end{tabular} \\ \hline
Cholec80~(\cite{endonet-cholec80}) & \checkmark & \checkmark & \xmark & \checkmark & \xmark & \xmark & \xmark & \xmark & 51.25 & \checkmark & LS\\
\rowcolor[HTML]{EFEFEF}
M2CAI16~(\cite{stauder2016lapchole}) & \xmark & \checkmark & \xmark & \xmark & \xmark & \xmark & \xmark & \xmark & 23.24 & \checkmark & LS \\
JIGSAWS~(\cite{jigsaws}) & \xmark & \xmark & \xmark & \xmark & \xmark & \xmark & \checkmark & * & 2.62 & \checkmark & ST \\
\rowcolor[HTML]{EFEFEF}
Nephrec9~(\cite{jigsaws}) & \xmark & \xmark & \checkmark & \xmark & \xmark & \xmark &\xmark  & \xmark & 10.52 & \checkmark & RASH \\
EndoVis 2015~(\cite{endovis2015})  & \xmark & \xmark & \xmark & \checkmark & \xmark & \checkmark & \xmark & \xmark & 0.10 & \checkmark & LS, ST \\
\rowcolor[HTML]{EFEFEF}
EndoVis 2017~(\cite{endovis2017})  & \xmark & \xmark & \xmark & \checkmark & \checkmark & \checkmark & \xmark & \xmark & 0.83 & \checkmark & RASA \\
EndoVis 2018~(\cite{endovis2018, isinet, islam2020interactions})  & \xmark & \xmark & \xmark & \checkmark & \checkmark & \checkmark & \checkmark & \checkmark & 1.58 & \checkmark & RASA \\
\rowcolor[HTML]{EFEFEF}
LSRAO~(\cite{sharghi2020lrsao}) & \xmark & \xmark & \xmark & \xmark & \xmark &\xmark & \checkmark & \xmark & ? & \xmark & RASH \\
CholecT40, T45 \& T50~(\cite{cholecT40,nwoye2021rendezvous}) & \checkmark & \checkmark & \xmark & \checkmark & \xmark & \xmark & \checkmark & \xmark & 28.03 & \checkmark & LS \\
\rowcolor[HTML]{EFEFEF}
HeiCo~(\cite{heico,rob2021robustmis}) & \checkmark & \checkmark & \xmark & \checkmark & \checkmark & \checkmark & \xmark & \xmark & 96.12 & \checkmark & LS\\
MISAW~(\cite{misaw}) & \checkmark & \checkmark & \checkmark & * & \xmark & \xmark & \checkmark & \xmark & 1.52 & \checkmark & ST \\
\rowcolor[HTML]{EFEFEF}
ESAD~(\cite{bawa2021saras_esad}) & \xmark & \xmark & \xmark & \xmark & \xmark & \xmark & \checkmark & \checkmark & 9.33 & \checkmark & RASH \\
SAR-RARP50~(\cite{amsterdam2022rarp45,psychogyios2024sarrarp50}) & \checkmark & \xmark & \xmark & \checkmark & \xmark & \checkmark & \checkmark & \xmark & 3.2 & \checkmark & RASH \\
\rowcolor[HTML]{EFEFEF}
AutoLaparo~(\cite{wang2022autolaparo}) & \checkmark & \checkmark & \xmark & \checkmark & \checkmark & \checkmark & \xmark & \xmark & 23.13 & \checkmark & RASH \\
RARR~(\cite{kolbinger2023contextaware}) & \xmark & \checkmark & \xmark & \xmark & \xmark & \xmark & \xmark & \xmark & 393.3 & \xmark & RASH \\ 
\rowcolor[HTML]{EFEFEF}
HeiChole~(\cite{wagner2023heichole}) & \checkmark & \checkmark & \xmark & \checkmark & \xmark & \xmark & \checkmark & \xmark & 22 & \checkmark & LS \\ \hline \hline
PSI-AVA~(\cite{valderrama2022tapir}) & \checkmark & \checkmark & \checkmark & \checkmark & \checkmark & \xmark & \checkmark & \checkmark & 20.45 & \checkmark & RASH \\ 
\rowcolor[HTML]{EFEFEF}
\textbf{GraSP (this-work)} & \boldcheckmark & \boldcheckmark & \boldcheckmark & \boldcheckmark & \boldcheckmark & \boldcheckmark & \boldcheckmark & \boldcheckmark& 32.37 & \boldcheckmark & RASH \\ \hline
\end{tabular}
}
\label{tab:datasets}
\end{table*}
\renewcommand{\arraystretch}{1}

\subsection{Surgical Workflow Analysis Benchmarks}

Table~\ref{tab:datasets} summarizes key SWA benchmarks introduced over the years along with their respective semantic tasks. Early efforts primarily focused on temporal analysis with independent video or frame-level tasks. The JIGSAWS dataset~(\cite{jigsaws_dataset}) used videos from simulated robot surgeries annotated for surgical skill assessment and gesture recognition. The Cholec80 dataset~(\cite{endonet-cholec80}) first introduced \textit{in vivo} human data of laparoscopic cholecystectomies with annotations for surgical phase and tool recognition. Simultaneously, M2CAI16~(\cite{stauder2016lapchole}) used similar data with surgical phase labels only. The Nephrec9 dataset~(\cite{nephrec9}) first introduced human Robot-Assisted Partial Nephrectomy data with finer labels for surgical step recognition rather than phases. However, none of these datasets provided any spatial annotations.

Furthermore, the HeiCo~(\cite{heico}) and AutoLaparo~(\cite{wang2022autolaparo}) datasets feature various laparoscopic procedures with temporal and spatial annotations for phase recognition and surgical instrument segmentation, but lack more granular temporal tasks. 
The CholecT50 dataset~(\cite{cholecT40,nwoye2021rendezvous,nwoye2023data}) used part of Cholec80's data with additional test videos and annotations for presence recognition of instruments, actions, and anatomical targets using surgical action triplets. Similarly, the HeiChole dataset~(\cite{wagner2023heichole}) encompasses laparoscopic cholecystectomies and presents annotations for phases, actions, and instrument recognition. While CholecT50 and HeiCHole allow for finer-grained SWA, they still disregard spatially localized information or step recognition.

Recent frameworks have partially addressed previous gaps in multi-granularity modeling and the availability of RAS data. The MISAW dataset~(\cite{misaw}) focuses on artificial micro-surgical anastomosis, offering temporal annotations across three granularity levels: phases, steps, and specific activities, though it lacks spatial information. In contrast, the RARP-45 dataset~(\cite{amsterdam2022rarp45}) comprises videos from human Robot-Assisted Radical Prostatectomies with temporal annotations for gesture recognition, and spatial annotation for instrument segmentation~(\cite{psychogyios2024sarrarp50}). The RARR dataset~(\cite{kolbinger2023contextaware}) introduces annotations for phase recognition and anatomical target segmentation in human \textit{in vivo} robot-assisted rectal resection. However, RARP45 and RARR only account for a single temporal task.

Conversely, our dataset uniquely combines annotations for high-level surgical phases and their decomposition into steps, making it the only SWA dataset to offer both levels of temporal semantic understanding using human \textit{in vivo} data. Additionally, it incorporates detailed annotations for atomic actions and instrument segmentation, thus addressing previous gaps in temporal and spatial multi-granularity and presenting a more holistic approach to SWA.

\subsection{Surgical Instrument Segmentation Benchmarks}

Most benchmarks for surgical instrument segmentation originated from the Endoscopic Vision Grand Challenge\footnote{\url{https://endovis.grand-challenge.org/}}.
First, the 2015 Instrument Segmentation and Tracking Challenge (Endovis 2015)~(\cite{endovis2015}), offered laparoscopic colorectal surgeries and \textit{ex vivo} robot-assisted procedures with semantic segmentation of instrument sub-parts.
Later, the 2017 Robotic Instrument Segmentation Challenge (Endovis 2017)~(\cite{endovis2017}) introduced instrument type recognition with instance-wise annotations in robot-assisted porcine procedures. 
Similarly, The 2018 Robotic Scene Segmentation Challenge (Endovis 2018)~(\cite{endovis2018}) used similar data with semantic segmentation annotations for instrument parts, anatomical structures, and additional surgical objects. 
Nevertheless, these benchmarks lack higher-level temporal tasks, do not include human \textit{in vivo} RAS data, and, in the cases of EndoVis 2015 and 2018, they overlook the instance-level nature of surgical instrument segmentation. 

In our previous research~(\cite{isinet}), we adapted Endovis 2018's annotations for instance-level segmentation of instrument types and demonstrated the superior effectiveness of instance-based approaches for surgical instrument segmentation.  
Further, the 2019 ROBUST-MIS challenge~(\cite{rob2021robustmis}) used HeiCo's~(\cite{heico}) data to establish the first instance-based evaluation for instrument-type segmentation, which was subsequently employed by AutoLaparo~(\cite{wang2022autolaparo}) on laparoscopic hysterectomies.
More recently, the SAR-RARP50~(\cite{psychogyios2024sarrarp50}) challenge expanded RARP-45's~(\cite{amsterdam2022rarp45}) data and introduced the first segmentation dataset with human RAS data for semantic segmentation of instrument parts and surgical objects. Consequently, no prior dataset offers instance-level instrument type segmentation on human RAS data, making our benchmark the first to provide this critical task.

\subsection{Surgical Action Recognition}

In the surgical context, action recognition represents a finer temporal task than phases or steps~(\cite{khatibi,rupprecht}), and several formulations have been developed to model this task. 
The JIGSAWS dataset~(\cite{jigsaws_dataset}) first defined surgical actions as gestures, which represent short temporal segments of discrete surgical activity units, an approach later adopted by the RARP-45~(\cite{amsterdam2022rarp45}) benchmark. 
Subsequently, the CholecT50 dataset~(\cite{cholecT40,nwoye2021rendezvous}) formulated surgical actions as multiple frame-wise action triplets in the structure of  $<$\textit{instrument, verb, target}$>$, which were also utilized by the MISAW dataset~(\cite{misaw}). 
The HeiChole~(\cite{heico}) datasets approached action recognition as a frame-wise presence recognition problem on a small set of atomic action categories.
However, these benchmarks do not provide the explicit location of actions, which would allow for a more detailed understanding of surgical scenes.   

Furthermore, \cite{islam2020interactions} introduced action localization within surgical scenes in the Endovis 2018 dataset using a graph structure, where nodes represented instrument or tissue bounding boxes, and edges depicted the interactions between instruments and tissues. Subsequently, the ESAD dataset~(\cite{bawa2021saras_esad}) extended the localization of actions to Robot-Assisted Radical Prostatectomy data, using bounding boxes around visible instruments and labeling their primary actions.
Nonetheless, these datasets assign each instrument a single action category and offer classes that could be further analyzed to identify even finer actions. 

Contrary to previous approaches, our work adopts the Atomic Visual Actions~(\cite{AVA_2018}) formulation for a more detailed study of temporal activities. By definition, atomic actions are indivisible, agent-centered, and represent the finest level of action granularity. Accordingly, our benchmark deconstructs surgical activities into various independent atomic actions that each surgical tool performs simultaneously. As a result, GraSP presents a localized and multi-target atomic action detection task, a novel and highly challenging approach that facilitates more in-depth analysis of surgical actions and is a naturally transferable notion to any surgical domain.

\subsection{Surgical Scene Understanding with Vision Transformers}

In recent years, transformer models~(\cite{attention,vit}) have gained prominence for their ability to capture long-range dependencies using attention mechanisms. These models offer a superior modeling capabilities that consistently surpass Convolutional Neural Networks (CNNs) in image and video analysis~(\cite{vit, detr, mvit, def_detr}). This success has impacted the domain of endoscopic surgeries. For instance, LapFormer~(\cite{lapformer}) and Opera~(\cite{opera}) employed CNN-based frame feature extractors with temporal consistency transformers for instrument and phase recognition, respectively. Moreover, Trans-SVNet~(\cite{transvnet}) and SAHC~(\cite{ding2021sahc}) used transformer layers to refine the outputs of a CNN backbone and a Temporal Convolutional Network for phase recognition. Recently, LoVit and SKiT~(\cite{liu2025lovit,liu2023skit}) used a fully transformer-based, two-stage architecture for phase recognition, leveraging a Vision Transformer (ViT)~(\cite{vit}) backbone and a long-term transformer. However, these phase recognition methods rely on global frame-wise encoders to produce generalized frame embeddings for long-range temporal architectures, hence limiting their adaptability for more granular tasks that demand finer spatial and temporal embeddings.

Furthermore, transformers have been utilized to improve surgical action triplet recognition. Rendezvous~(\cite{nwoye2021rendezvous}) employs a CNN backbone with an adapted transformer encoder, while Rendezvous in Time (RiT)~(\cite{nwoye2023rit}) further enhances the encoder by integrating a temporal consistency module. In contrast, MCIT-IG~(\cite{sharma2023surgical}) leveraged a transformer detector~(\cite{def_detr}) coupled with a CNN backbone and a modified transformed decoder to improve triplet identification. Nevertheless, these models also utilize frame-wise encoders and have limited temporal processing capabilities. While these models can process certain levels of activity localization~(\cite{nwoye2021rendezvous}), they do not explicitly output instruments or action regions, which limits their effectiveness for detailed spatial analysis.

Regarding instrument segmentation, TraSeTR~(\cite{trasetr}) utilized a modified MaskFormer~(\cite{cheng2021maskformer}) with a CNN backbone, integrating instrument region embeddings of previous frames to improve temporal modeling. Furthermore, \cite{rohan2023qpd} employed a Mask DINO~(\cite{li2023mask}) model, combining a CNN backbone with a Query Proposal Network to enhance region proposals. Recent approaches, such as~\cite{zhou2023promptable,yue2023surgicalsam}, employ prompting strategies for foundation transformer models to improve generalizability, but these methods differ from the supervision approach of this work. Still, current segmentation architectures use very limited temporal context and cannot adapt to localized temporal tasks like atomic actions.

Therefore, previous methodologies are tailored to specific surgical scene understanding tasks, often relying on frame-wise CNN feature extractors. Consequently, no fully transformer-based architecture has been developed to adapt to different tasks while capturing the holistic and multi-granular nature of surgeries, therefore limiting the exploitation of transformers' powerful modeling capabilities.
Regarding these gaps, our previous works, \cite{valderrama2022tapir} and \cite{ayobi2023matis}, demonstrated outstanding results on diverse tasks by using a fully transformer-based model. In this regard, our TAPIS model introduces a transformer method incorporating a video-wise feature encoder and a region proposal network. This design enables TAPIS to effectively handle multiple surgical workflow analysis tasks at various levels of visual, temporal, and spatial understanding. Thus, it presents the first generalized state-of-the-art transformer model for understanding multi-granular surgical scenes.

\begin{table*}[t]
\caption{\textbf{Category sets for all the presented tasks in GraSP.} We present the defined sets of semantic categories for each proposed task in our GraSP dataset. We showcase the ID number and the defined label name for each phase, step, instrument, or atomic action category.}
\resizebox{\textwidth}{!}{
\begin{tabular}{llllll|ll}
\hline
\multicolumn{2}{c|}{\textbf{Phases Categories}}                                    & \multicolumn{4}{c|}{\textbf{Steps Categories}}                                                                                                                                                                                                                                                                                                                                      & \multicolumn{2}{c}{\textbf{Atomic Action Categories}}                                                     \\ \hline
0  & \multicolumn{1}{l|}{Idle.}                                                     & 0                  & Idle.                                                                                                                                                            & 10                 & Cut the tissue between the prostate and the urethra.                                                                                                                     & 1                   & Cauterize                                                                           \\
1  & \multicolumn{1}{l|}{Left pelvic isolated lymphadenectomy.}                     & 1                  & Identification and dissection of the Iliac vein and artery.                                                                                                      & 11                 & Hold prostate.                                                                                                                                                           & 2                   & Close                                                                               \\
2  & \multicolumn{1}{l|}{Right pelvic isolated lymphadenectomy.}                    & 2                  & Cutting and dissection of the external iliac vein’s lymph node.                                                                                                  & 12                 & Insert prostate in retrieval bag.                                                                                                                                        & 3                   & Cut                                                                                 \\
3  & \multicolumn{1}{l|}{Developing the Space of Retzius.}                          & \multirow{2}{*}{3} & \multirow{2}{*}{\begin{tabular}[c]{@{}l@{}}Obturator nerve and vessel path identification, dissection \\ and cutting of the obturator lymph nodes.\end{tabular}} & 13                 & Pass suture to the urethra.                                                                                                                                              & 4                   & Grasp                                                                               \\
4  & \multicolumn{1}{l|}{Ligation of the deep dorsal venous complex.}               &                    &                                                                                                                                                                 & 14                 & Pass suture to the bladder neck.                                                                                                                                         & 5                   & Hold                                                                                \\
5  & \multicolumn{1}{l|}{Bladder neck identification and transection.}              & 4                  & Insert the lymph nodes in retrieval bags.                                                                                                                        & 15                 & Pull suture.                                                                                                                                                             & 6                   & Open                                                                                \\
6  & \multicolumn{1}{l|}{Seminal vesicle dissection.}                               & 5                  & Prevessical dissection.                                                                                                                                          & 16                 & Tie suture.                                                                                                                                                              & 7                   & Open Something                                                                      \\
7  & \multicolumn{1}{l|}{Development of the plane between the prostate and rectum.} & 6                  & Ligation of the dorsal venous complex.                                                                                                                           & 17                 & Suction.                                                                                                                                                                 & 8                   & Pull                                                                                \\
8  & \multicolumn{1}{l|}{Prostatic pedicle control.}                                & 7                  & Prostate dissection until the levator ani.                                                                                                                       & 18                 & Cut suture or tissue.                                                                                                                                                    & 9                   & Push                                                                                \\
9  & \multicolumn{1}{l|}{Severing of the prostate from the urethra.}                & 8                  & Seminal vesicle dissection.                                                                                                                                      & 19                 & Cut between the prostate and bladder neck.                                                                                                                               & 10                  & Release                                                                             \\
10 & \multicolumn{1}{l|}{Bladder neck reconstruction.}                              & 9                  & Dissection of Denonviliers’ fascia.                                                                                                                              & 20                 & Vascular pedicle control.                                                                                                                                                & 11                  & Still                                                                               \\ \cline{1-6}
\multicolumn{6}{c|}{\textbf{Instrument Categories}}                                                                                                                                                                                                                                                                                                                                                                                                                      & 12                  & Suction                                                                             \\ \cline{1-6}
1  & Bipolar Forceps (BF)                                                          & 4                  & Monopolar Curved Scissors (MCS)                                                                                                                                 & \multirow{3}{*}{7} & \multirow{3}{*}{\begin{tabular}[c]{@{}l@{}}Laparoscopic Instrument (LI):\\ {[}Laparoscopic Retraction Forceps, Laparoscopic\\Suture Scissors, Laparoscopic Needle Holder{]}\end{tabular}} & 13                  & Travel                                                                              \\
2  & Prograsp Forceps (PF)                                                         & 5                  & Suction Instrument (SI)                                                                                                                                         &                    &                                                                                                                                                                         & \multirow{2}{*}{14} & \multirow{2}{*}{\begin{tabular}[c]{@{}l@{}}Other:\\ {[}Staple, Wash{]}\end{tabular}} \\
3  & Large Needle Driver (LND)                                                     & 6                  & Clip Applier (CA)                                                                                                                                               &                    &                                                                                                                                                                         &                     &                                                                                     \\ \hline
\end{tabular}}
\label{tab:categories}
\end{table*}
\label{relatedwork}

\section{The Holistic and Multi-Granular Surgical Scene Understanding of Prostatectomies Dataset}\subsection{Data Collection}
\label{sec:data_colection}
We recorded 13 surgeries of patients undergoing robotic-assisted laparoscopic Radical Prostatectomy at Fundación Santa Fé de Bogotá, with their explicit consent for data usage. We followed all the data collection protocols with the approval of the ethics committees of all institutions directly involved in this work.

Three highly experienced surgeons performed the surgeries using a da Vinci Si 3000 robot manufactured by Intuitive Inc. (Sunnyvale, California). The equipment included the robot assembled with a high-definition video camera, providing independent viewing for each eye. For recording purposes, we extracted a 2D image signal from the left view of the camera through an HDMI cable connected to a KARL STORZ (AIDA mini WD 100) high-definition video capture device, providing real-time imagery at 720 dots per inch (dpi). We anonymized all the data collected according to our ethical protocols and we manually removed all instrument usage flags or signals in our video frames to avoid biases in model training and evaluation. We provide an example of this frame modification procedure in Figure~\ref{fig:post_processing} of the Supplementary Material.

The team placed the Monopolar Energy Scissors (Monopolar Curved Scissors) in port \#1 and the Bipolar Energy Forceps (Bipolar Forceps), capable of being fenestrated or configured as Maryland Forceps, in port \#2. The third port accommodated a Prograsp Retraction Forceps (Prograsp Forceps). For the reconstructive stages of the surgery, surgeons removed the previous instruments located in ports \#1 and \#2 and positioned two robotic Needle Holders (Large Needle Drivers). Additionally, the surgical setup designated two ports for the bedside assistant: one port housed a Laparoscopic Suction cannula (Suction Instrument), and the other contained either a Laparoscopic Intestinal Retraction Forceps, a Laparoscopic Needle Holder, a Laparoscopic Scissors designed explicitly for cutting suture material (Laparoscopic Instruments), or a Clip Applier with Plastic Hemostatic clips of the HEM-O-LOCK type.

\subsection{Long-Term Tasks Annotation}

We categorize phases and steps as long-term reasoning tasks, following the temporal breakdown outlined in~\cite{misaw}. The definitions of phase and step categories and their respective nomenclature adhered to the procedural framework of radical prostatectomy with bilateral extended lymphadenectomy, as validated by the guidelines in \cite{campbell_2020}. Within this framework, defining surgical phases involved establishing the main stages within the surgical process, each occurring singularly and with minimum repetition. In contrast, surgical steps refer to specific acts undertaken during the surgical process, often occurring multiple times. Hence, the aggregation of multiple steps forms a phase and the length of a phase correlates with the amount of steps.

We then established the sequencing and hierarchy of phase and step categories based on standardized stages of Robot-Assisted Radical Prostatectomy, as routinely practiced at Fundación Santa Fé de Bogotá. Subsequently, two urologists specializing in Robotic-Assisted Surgery reviewed and refined the final category sets, which we portray in Table~\ref{tab:categories}, and their hierarchical structure is visually represented in Figure~\ref{fig:dendogram} in the Supplementary Material. 

After establishing the labels and hierarchies of the long-term tasks, we conducted an annotation procedure that involved a team of two urologist residents specializing in transperitoneal radical prostatectomy surgeries. Each resident independently annotated all surgical videos, defining specific timestamps for the start and end times of phases and step intervals down to the hour, minute, and second, resulting in two annotations per video. In the case of the eight previously annotated surgeries, these were annotated by the team once more, also resulting in two sets of annotations per surgery. When discrepancies arose between the annotations, the resident annotators collectively reviewed and reconciled differences to establish a consensus on the final annotations.

\subsubsection{Long-Term Annotation Consistency Analysis}
To evaluate the consistency between our annotators, we analyzed the percentage of frames with differing annotations for each annotator. For phase labels, we observed differing annotations in 21\% of frames overall, with an average of 20\% differing frames per video and a standard deviation of 9.33\%. Step labels exhibited a higher rate, with differing annotations in 37\% of frames overall, averaging 36\% per video and a standard deviation of 11.33\%. These statistics indicate a fair annotator agreement, as the majority of frames were consistently annotated, thus supporting the reliability of our dataset. The increased percentage of differing step annotations reflects their finer granularity and shorter temporal segments, making them more prone to interpretative differences and temporal misalignments. Furthermore, longer videos showed higher rates of differing annotations, likely due to the increased potential for misalignment over extended segments. Despite these challenges, the annotation team's conciliation process ensured the final dataset's accuracy and quality for model training and evaluation.

\subsection{Short-Term Tasks Annotation}
\label{sec:short-term_tasks}

We define Surgical Instrument Segmentation and Atomic Action Detection as short-term, frame-wise, and spatial localization tasks. To ensure balanced category representation while optimizing computational and annotation resources, we only annotated keyframes sampled at 35-second intervals throughout the surgical videos. Initially, the annotations for Instrument Segmentation and Atomic Action Detection were performed independently by separate teams but were later unified into a single dataset, as described in the following subsections. Throughout the process, all annotators had access to 3-second clips centered on each keyframe and full videos for their assigned keyframes to support accurate short-term task annotation. We provide a visual diagram of our short-term annotation process in Figure~\ref{fig:annotation_diagram} of the Supplementary Material.

\subsubsection{Surgical Instrument Segmentation}
\label{sec:segmentation_annotations}

The annotation process for instrument segmentation followed a structured, multi-step approach to ensure high-quality and consistent annotations across the dataset.
First, we defined surgical instrument classes based on the da Vinci Robot user manual~(\cite{daVinci,intuitive}) and the instrument setup described in Section~\ref{sec:data_colection}, with the final categories detailed in Table~\ref{tab:categories}. Second, we trained a team of 17 annotators using a standardized and comprehensive tutorial on instrument delineation and classification, followed by multiple annotation tests on challenging frames (e.g., Figure~\ref{fig:hard_examples} of the Supplementary Material) where annotators had to consistently produce high-quality segmentation masks.

Subsequently, each annotator was randomly assigned a similar number of keyframes to ensure balanced workloads. All annotators were provided with standardized annotation resources, including the open-source platforms Label-Studio (\cite{LabelStudio}) and Toronto Annotation Suite (\cite{torontoannotsuite}), instrument bounding boxes from surgeries annotated in \cite{valderrama2022tapir}, and mask predictions generated by the MATIS model~(\cite{ayobi2023matis}) pre-trained on Endovis 2017~(\cite{endovis2017}) and 2018~(\cite{endovis2018}). Annotators were free to utilize any of these resources to delineate and label instrument instances on each keyframe, choosing to either draw and tag masks from scratch, refine the bounding boxes provided by \cite{valderrama2022tapir} while verifying instrument categories and missing instances, or enhance and label MATIS-generated predictions. Notably, annotators heavily relied on the 3-second clips centered on each keyframe for precision, with most masks being manually drawn with the assistance of each platform's annotation tools.

After the initial round of segmentation annotations, four surgical data science experts conducted a thorough review to identify errors in classification, delineation, or missed instances. The surgical data scientist then meticulously documented these issues for further correction. We found errors in 9\% of final instances, with 8.2\% attributed to delineation issues such as imprecise boundaries, inclusion of blood or tissue within the mask, or failure to exclude openings in instrument tips. Misclassifications accounted for 0.3\% of instances, while 0.6\% involved missing instances. These flagged instances were returned to the annotators for a second round of re-annotation. Following this re-annotation phase, the surgical data science team performed a final validation review of the entire dataset, identifying and personally correcting minor delineation errors in 2\% of the instances. While delineation errors are expected due to the inherent complexity of segmentation tasks, our iterative, multi-staged annotation protocol effectively minimized all error types, ensuring high-quality and reliable annotations.

\begin{table}[t]
\caption{\textbf{Distribution of the number of frames on each defined dataset split}. The Table presents the duration in seconds and the number of frames sampled at 1fps and 30fps of each collected Case in our dataset. Each Case corresponds to an entire surgery performed. We also show our predefined dataset partition into Fold 1 and Fold 2 for two-fold cross-validation and a Test set for final model testing.}
\centering
\resizebox{\linewidth}{!}{
\begin{tabular}{ccccc}
\hline
\textbf{Set} & \textbf{Case No.} & \textbf{Duration (h)} & \textbf{Frames  1fps} & \textbf{Frames 30fps} \\ \hline
\multirow{4}{*}{\textbf{Fold   1}} & 001 & 3.05 & 10972 & 329159  \\
 & 004 & 2.36 & 8490 & 254678 \\
 & 014 & 2.77 & 9973 & 299118 \\
 & 015 & 2.48 & 8969 & 269031 \\ \hline
\multirow{4}{*}{\textbf{Fold   2}} & 002 & 2.62 & 9443 & 283261 \\
 & 003 & 1.63 & 5882 & 176453 \\
 & 007 & 2.23 & 8018 & 240522 \\
 & 021 & 3.30 & 11871 & 356097 \\ \hline
\multirow{5}{*}{\textbf{Test}} & 041 & 2.63 & 9452 & 283564 \\
 & 047 & 1.81 & 6504 & 195131 \\
 & 050 & 3.86 & 13899 & 416973\\
 & 051 & 0.69 & 2496 & 74876 \\
 & 053 & 2.93 & 10552 & 316552 \\ \hline
\textbf{Total} & 13 & 32.37 & 116521 & 3495415 \\ \hline
\end{tabular}}
\label{tab:general_info}
\end{table}

\subsubsection{Atomic Action Detection}
\label{sec:atomic_action_annotation} 

The annotation process for atomic action detection was adapted from the methodology presented in \cite{AVA_2018}. Our surgical experts first defined a detailed set of atomic actions representing the most detailed actions performed by instruments during surgery, treating these actions as localized, discrete, and instrument-centric states that could be undertaken simultaneously by each instrument. The final atomic action categories are shown in Table~\ref{tab:categories}. Additionally, the surgical team trained the surgical data scientists to identify each atomic action accurately. Alternatively, the surgical team specified which actions could be performed simultaneously by the same instrument and which types of instruments could perform each action. We further organized this information into theoretical co-occurrence matrices (Figure~\ref{fig:expert_coocurrences} in the Supplementary Material) that we later used to validate the annotations.

Furthermore, the annotation process was conducted per frame and per instance, meaning that each instrument in every keyframe was labeled independently. The surgical data scientists used the Toronto Annotation Suite (\cite{torontoannotsuite}) to draw bounding boxes around each instrument instance in the keyframes. To determine the atomic actions performed by each instrument, annotators relied on the 3-second video clips centered on each keyframe, identifying the instrument’s activity at the exact keyframe moment (1.5 seconds into the clip). Thus, each bounding box per keyframe was annotated with up to three atomic action categories, as each instrument instance can perform a maximum of three atomic actions simultaneously.

To ensure annotation robustness and reduce biases, each instrument bounding box was independently labeled three times by different surgical data scientists, resulting in three sets of atomic action labels per instance. The final set of labels for each instance was determined through a consensus approach by retaining only the actions present in at least two of the three annotation sets. 

For 2.18\% of instances, the overlap between the three annotation sets was notably low (intersection-over-union of two sets over three below 0.3). These cases were carefully reviewed by the surgical data science team, who collaborated to establish the final combination of atomic action labels. Additionally, 5.4\% of instances did not align with the established theoretical co-occurrence matrixes. These discrepancies mostly arose when specific actions that could not logically occur in isolation were left as standalone labels after the consensus process. To solve these inconsistencies, the surgical data science team thoroughly reviewed the three original sets of atomic action labels for these instances to understand the source of the inconsistency and defined the final set of atomic action labels. Expert surgeons assisted in all the atomic action annotation review processes, ensuring medical consistency on final annotations.

\begin{figure}[t]
    \centering
    \includegraphics[width=\linewidth]{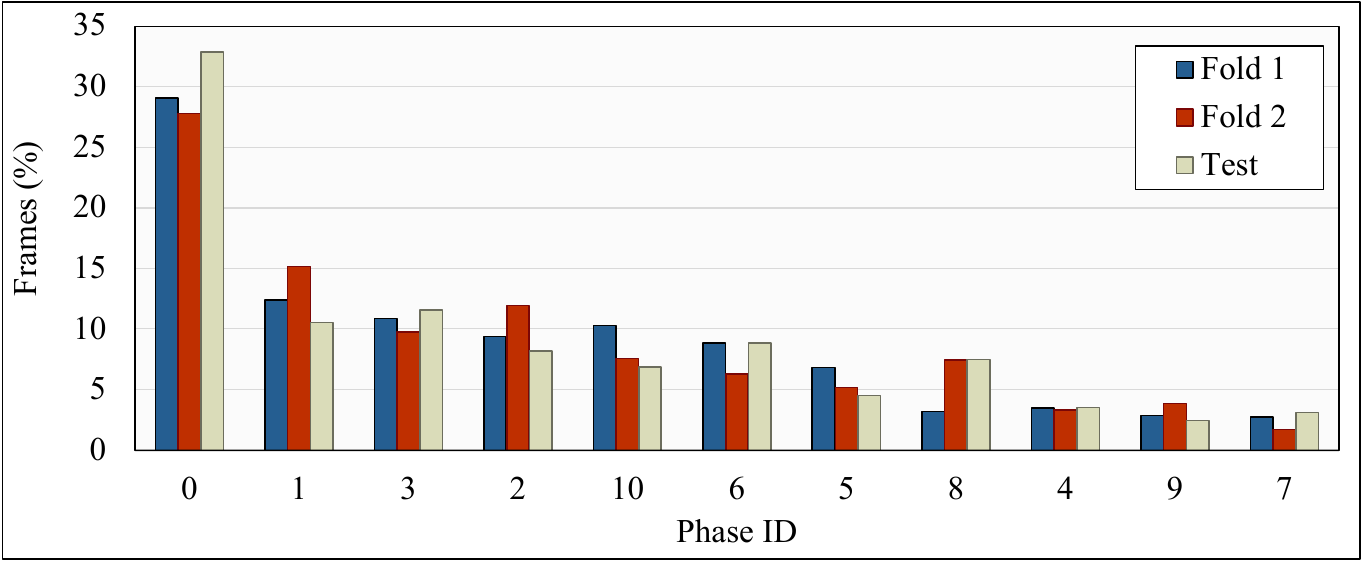}
    \caption{\textbf{Distribution of the percentage of frames per surgical phase category on each data split.} We organize the graphic in descending order and present correspondence between phase IDs and phase labels in Table~\ref{tab:categories}. The figure is best viewed in color.}
    \label{fig:perc_phases}
\end{figure}

\subsubsection{Instrument and Atomic Actions Annotation Alignment}

We aligned the atomic action annotations with the instrument segmentation annotations to produce a unified dataset. We achieved this alignment through a bipartite matching per keyframe, where we maximized the Intersection over Union (IoU) between the atomic action bounding boxes and the bounding boxes derived from the boundaries of the segmentation masks. This matching process also served as an additional validation step as it enabled a direct comparison between the instrument masks and the bounding boxes annotated by different teams to identify discrepancies in the number and size of annotated instances. During this process, we identified 2\% of the final instances missing from the atomic action annotations and 1.2\% from the segmentation annotations, which were mostly highly occluded instances. These instances were re-annotated by the surgical data scientists for both tasks to ensure completeness. Ultimately, we kept the bounding boxes derived from segmentation as the final boxes for atomic action annotations. These boxes adhered to the segmentation masks, making them more precise than those drawn during the atomic action annotation process. This procedure resulted in a cohesive dataset with unified and consistent annotations for instrument instance segmentation and atomic action detection.

\begin{figure}[t]
    \centering
    \includegraphics[width=\linewidth]{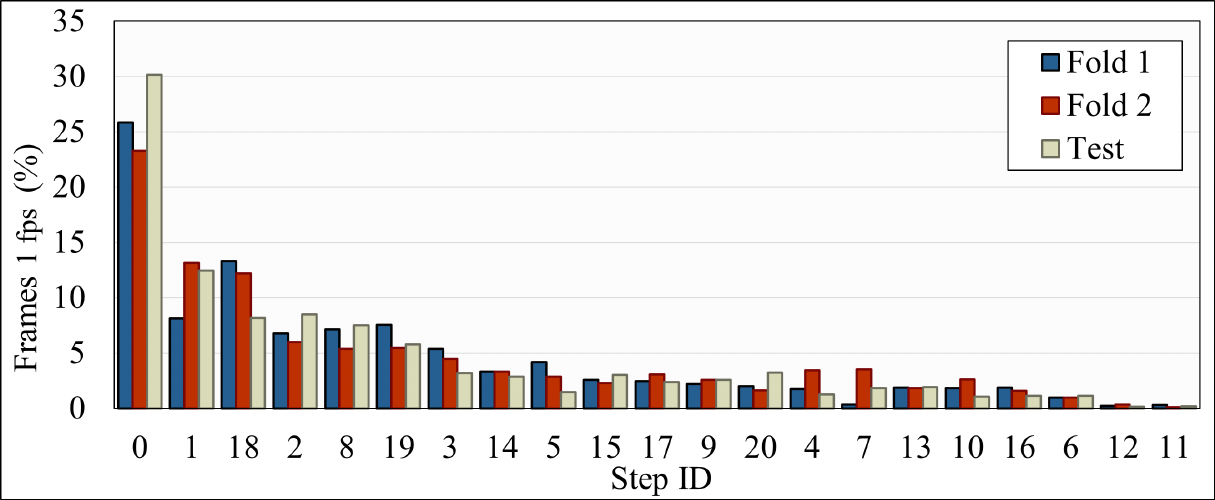}
    \caption{\textbf{Distribution of the percentage of frames per surgical step category on each data split.} We organize the graphic in descending order and present correspondence between step IDs and step labels in Table~\ref{tab:categories}. The figure is best viewed in color.}
    \label{fig:perc_steps}
\end{figure}

\subsection{Dataset Statistics} \label{ssec:dataset_statistics}

Our GraSP dataset consists of 13 Robot-Assisted Radical Prostatectomy videos, all captured at 30 fps. The total duration of these surgical videos is 32.37 hours, with an average video length of 2.49 hours and a standard deviation of 0.77 hours. For benchmarking purposes, we establish a two-fold cross-validation setup for training and validation and a separate test set for final performance evaluation. The original videos of the PSI-AVA dataset with curated annotations make the cross-validation set, and the test set comprises the five new videos included in this work. Table~\ref{tab:general_info} presents the dataset splits, including the duration and the total number of frames available, sampled at 30 fps and 1 fps. While the frames sampled at 30 fps are available for temporal window creation,  the 1 fps frames serve as the primary frames for model evaluation and statistical analysis.

Additionally, Figure~\ref{fig:grasp} and Figure~\ref{fig:general_visuals} in the Supplementary Material offer an overview of our proposed tasks and annotations. Table~\ref{tab:categories} lists the semantic categories we have established. Similarly, the hierarchical dendrogram in Figure~\ref{fig:dendogram} of the Supplementary Material illustrates the relationships between each surgical phase and its corresponding nested steps.

Figures~\ref{fig:perc_phases} and~\ref{fig:perc_steps} illustrate the distribution of the percentage of annotated frames for each phase and step class, respectively. We also present the number of frames per phase and step category in Tables~\ref{tab:frequencies_phases} and \ref{tab:frequencies_steps} of the Supplementary Material, along with transition matrices of the annotated categories and visualizations of the temporal progression of phases and steps in Figures~\ref{fig:mat_trans_phases}, \ref{fig:mat_trans_steps}, \ref{fig:overall_phase}, and \ref{fig:overall_step} of the Supplementary Material. Notably, these distributions exhibit a pronounced long-tail effect, with the \textit{Idle} class over-represented, as most phases and steps frequently transition from or to an \textit{Idle} state due to surgeons repeatedly pausing, resting, or adjusting the surgical setup between stages.

\begin{figure}[t]
    \centering
    \includegraphics[width=\linewidth]{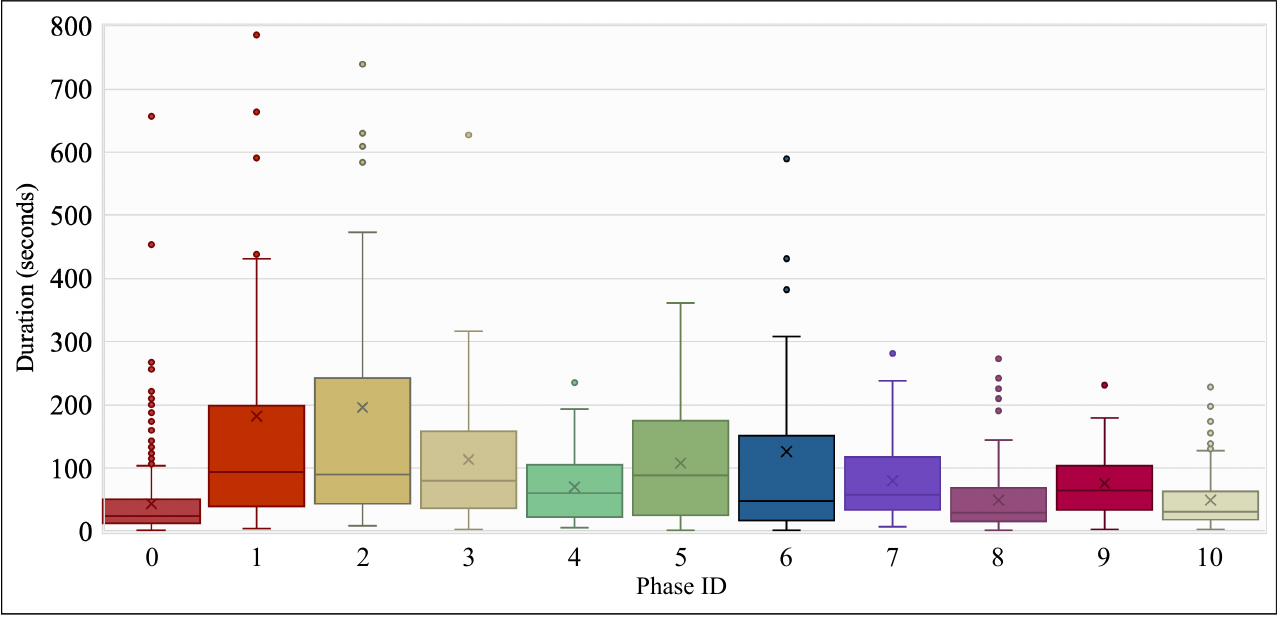}
    \caption{\textbf{Boxplot distribution of the duration of each phase category.} Each boxplot presents the distribution in seconds of the period of all the present temporal windows corresponding to each phase category in our dataset. We clipped this figure to a maximum of 800 seconds for better visualization; the entire figure is presented in Figure~\ref{fig:boxplot_complete_phases} of the Supplementary Material. The figure is best viewed in color.}
    \label{fig:boxplot_phases}
\end{figure}

\begin{figure}[t]
    \centering
    \includegraphics[width=\linewidth]{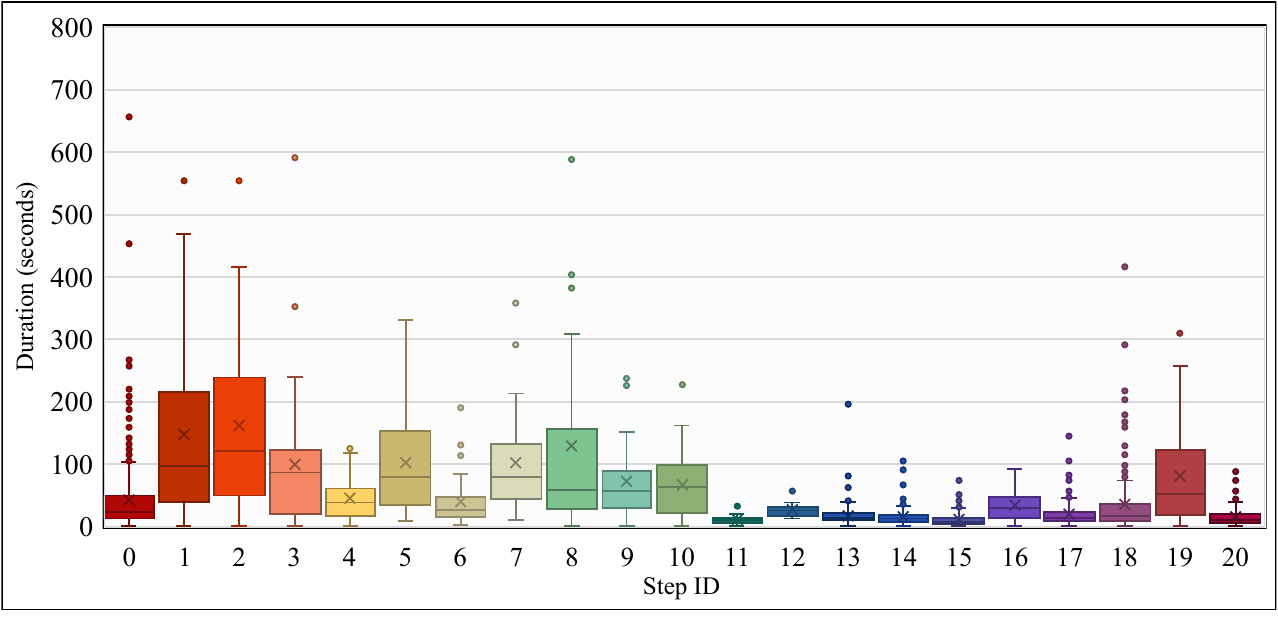}
    \caption{\textbf{Boxplot distribution of the duration of each step category.} Each boxplot presents the distribution in seconds of the period of all the present temporal windows corresponding to each step category in our dataset. We clipped this figure to a maximum of 800 seconds for better visualization; the entire figure is presented in Figure~\ref{fig:boxplot_complete_steps} of the Supplementary Material. The figure is best viewed in color.}
    \label{fig:boxplot_steps}
\end{figure}

In addition, we present the duration distribution, in seconds, of each continuous temporal segment for each phase and step category in Figures~\ref{fig:boxplot_phases} and \ref{fig:boxplot_steps}. These distributions and the visualizations in Figures~\ref{fig:overall_phase} and \ref{fig:overall_step} of the Supplementary Material underscore the granularity difference between phases and steps: phases represent broader surgery segments with longer average durations, with some segments lasting over 20 minutes. Conversely, steps recur in shorter and repetitive segments, with the most extended segments spanning around 10 minutes. Figures~\ref{fig:boxplot_phases} and~\ref{fig:boxplot_steps} also denote different intra-class duration variability across all phase and step categories with multiple outliers. Intuitively, categories involving tissue dissection, vessel cutting, or bleeding control (e.g., Phases 1-3, Steps 1-3,5 \& 6) often present longer durations and variability as they demand more time, may vary among patients, and often require alternation between steps. In contrast, categories related to the suturing process (e.g., Phase 8, Steps 13-16) show shorter durations and variability as these are usually performed within faster and more standardized time windows. These statistical distributions of our long-term annotations demonstrate substantial challenges within our benchmark and closely reflect the realities of surgical scenarios. 

\begin{figure}[t]
    \centering
    \includegraphics[width=\linewidth]{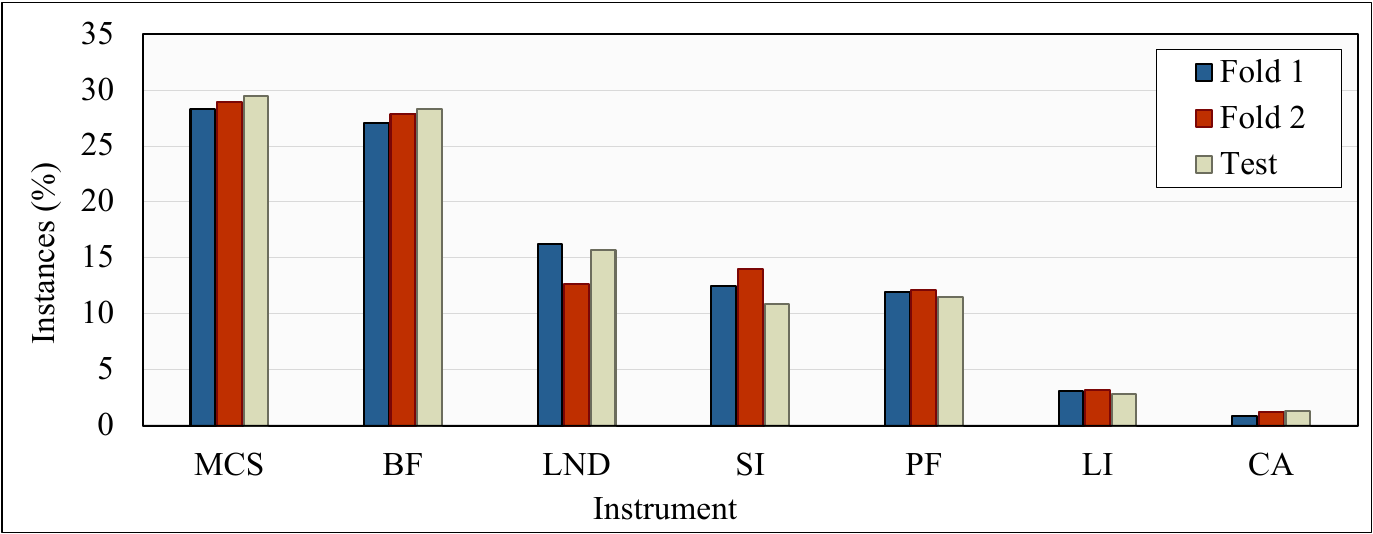}
    \caption{\textbf{Distribution of the percentage of instrument instances per instrument type category on each data split.} We organize the graphic in descending order. The instrument categories are Monopolar Curved Scissors (MCS), Bipolar Forceps (BF), Large Needle Driver (LND), Suction Instrument (SI), Prograsp Forceps (PF), Laparoscopic Instrument (LI) and Clip Applier (CA). The figure is best viewed in color.}
    \label{fig:perc_instruments}
\end{figure}

\begin{figure}[t]
    \centering
    \includegraphics[width=\linewidth]{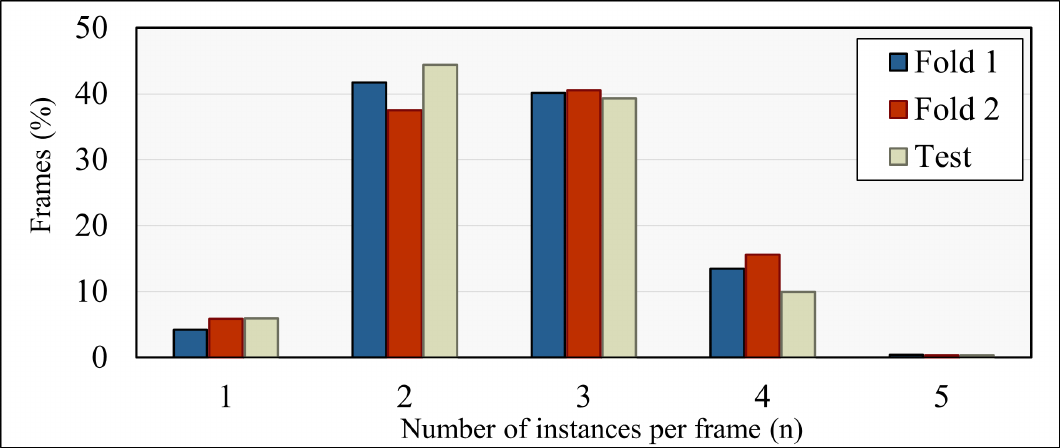}
    \caption{\textbf{Distribution of the percentage of frames presenting (n) instrument instances per frame on each data split.} The percentage of frames at 35-second intervals with (n) instrument instances. The figure is best viewed in color.}
    \label{fig:instruments_per_frame}
\end{figure}

Furthermore, Figure~\ref{fig:perc_instruments} and Table~\ref{tab:frequencies_instruments} (Supplementary Material) display the frequency distribution of instrument instances per class, also demonstrating a natural long-tail effect. This distribution is due to robotic instruments like BF, and MCS being more actively used owing to their higher versatility and action capacity than LND or PF, which are more specialized. In contrast, laparoscopic instruments like CA or LI are additional to the da Vinci Surgical System (dVSS) and are employed for specific purposes during particular stages of the surgery. Moreover, Figure~\ref{fig:instruments_per_frame} and Table~\ref{tab:instances_instruments_frame} (Supplementary Material) illustrate the distribution of the number of instrument instances per video keyframe. We observe that each keyframe contains 1 to 5 instrument instances (2.23 average of instrument instances per frame), with approximately 80\% of frames containing between 2 and 3 instances. These statistics are characteristic of the dVSS, typically involving two robotic instruments operated by the surgeon with occasional stationary robotic tools or additional non-robotic tools handled by an assistant that increase the complexity of surgical scenes.

\begin{figure}[t]
    \centering
    \includegraphics[width=\linewidth]{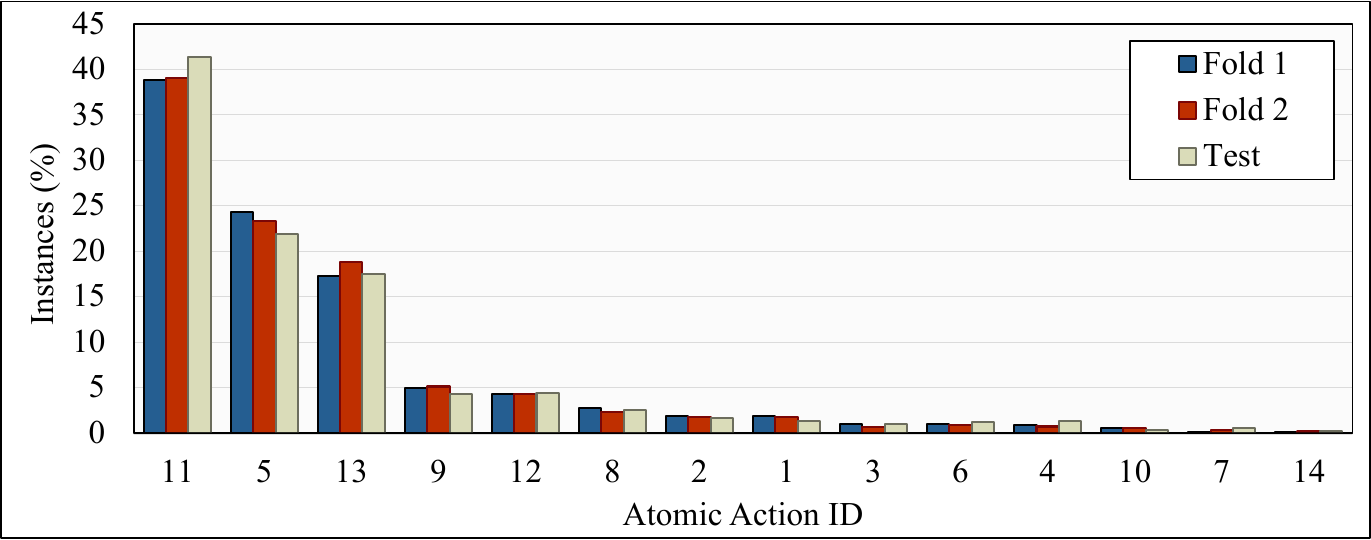}
    \caption{\textbf{Distribution of the percentage of instances per atomic action category on each data split.}  We consider each annotated action within an instrument instance an independent atomic action instance. We organize the graphic in descending order and present the correspondence between atomic action IDs and atomic action labels in Table~\ref{tab:categories}. The figure is best viewed in color.}
    \label{fig:perc_actions}
\end{figure}

Figure~\ref{fig:perc_actions} and Tables~\ref{tab:frequencies_actions} and \ref{tab:actions_coocurrences} (Supplementary Material) present the frequency distribution of the atomic action instances per class. Again, the distribution of atomic actions suggests a long-tail behavior where \textit{still}, \textit{hold}, and \textit{travel} are the most frequent since most instruments are used to hold and maintain tissues and objects in place or to move them to a different location. On the contrary, \textit{Push}, \textit{Pull}, and \textit{Suction} represent more specialized movements, and the remaining actions like \textit{Cauterize}, \textit{Cut}, \textit{Open} and \textit{Other}, are usually done by a specific type of instrument in very short time windows, thus making them the least frequent. 

Finally, the statistical distributions presented in Figures~\ref{fig:perc_phases},~\ref{fig:perc_steps},~\ref{fig:perc_instruments},~\ref{fig:instruments_per_frame} and~\ref{fig:perc_actions} demonstrate that our dataset maintains similar distributions across all the predefined dataset splits, highlighting the reliability of our entire dataset and annotation processes. Additionally, we include further statistical analysis in \ref{sec:appendix:stats} of the Supplementary Material. This section provides information about the size (Figure~\ref{fig:area_inst}) and total number of instrument instances (Table~\ref{tab:instances_instruments_frame}) in each fold of the GraSP dataset, and the co-occurrences matrixes for instruments-phases (Figure~\ref{fig:inst_phase}), instrument-steps (Figure~\ref{fig:inst_step}), and instruments-actions (Figure~\ref{fig:inst_action}). 

Our dataset statistics prove the intrinsic complementarity among the proposed tasks in GraSP. For instance, the most time-consuming phase categories encompass many of the most time-consuming step categories and vice versa. Likewise, multiple actions can only be performed by specific instruments, thus causing a correlation between instrument and action frequencies. Additionally, the co-occurrence matrixes demonstrate an evident correspondence between instruments and phases or steps as different instruments are more present or completely absent in different surgical stages. 

\subsection{Evaluation Metrics}

For phase and step recognition, we keep the mean Average Precision ($mAP$) originally proposed in \cite{valderrama2022tapir} as it is an accepted standard for frame-wise evaluation of action recognition (\cite{caba2015activitynet}). Additionally, we include the video-wise $F1 score$ as suggested in~\cite{funke2023metrics}.~\footnote{\href{https://scikit-learn.org/1.5/modules/model_evaluation.htmls}{sklearn.metrics}} Following standard practice~(\cite{endonet-cholec80,funke2023metrics}) we calculate these metrics on frames sampled at 1fps. 

For the instrument segmentation task, we adopt the conventional instance-based mean Average Precision ($mAP@0.5IoU_{segm}$) from PASCAL VOC (\cite{everingham2015pascal}) to promote research toward instance-based evaluation of this task.\footnote{\url{https://github.com/activitynet/ActivityNet}} Additionally, we maintain the standard and stringent semantic segmentation metrics we introduced in \cite{isinet}, which include the Mean Intersection over Union (mIoU), Intersection over Union (IoU), and Mean Class Intersection over Union (mcIoU).\footnote{\url{https://github.com/BCV-Uniandes/MATIS}}.

For the Atomic Action Detection task, we adhere to the evaluation framework established by AVA (\cite{AVA_2018,li2020avakinetics,caba2015activitynet}). Since surgical atomic actions are instrument-centric and the primary goal is to locate these actions rather than delineating the exact shape of the executing agent, we maintain the original object detection metric ($mAP@0.5IoU_{box}$) from AVA, applied to instrument bounding boxes instead of segmentation masks. Finally, given the multi-target nature of atomic action detection, where each instrument instance can perform up to three actions simultaneously, we evaluate the prediction of each atomic action category independently for each instrument instance instead of assessing the entire expected combination of ground truth actions. This methodology offers a more robust and sensitive way of measuring the atomic action detection performance.
\label{dataset}

\section{Transformers for Actions, Phases, Steps and Instrument Segmentation}\begin{figure*}[ht]
    \centering
    \includegraphics[width=\textwidth]{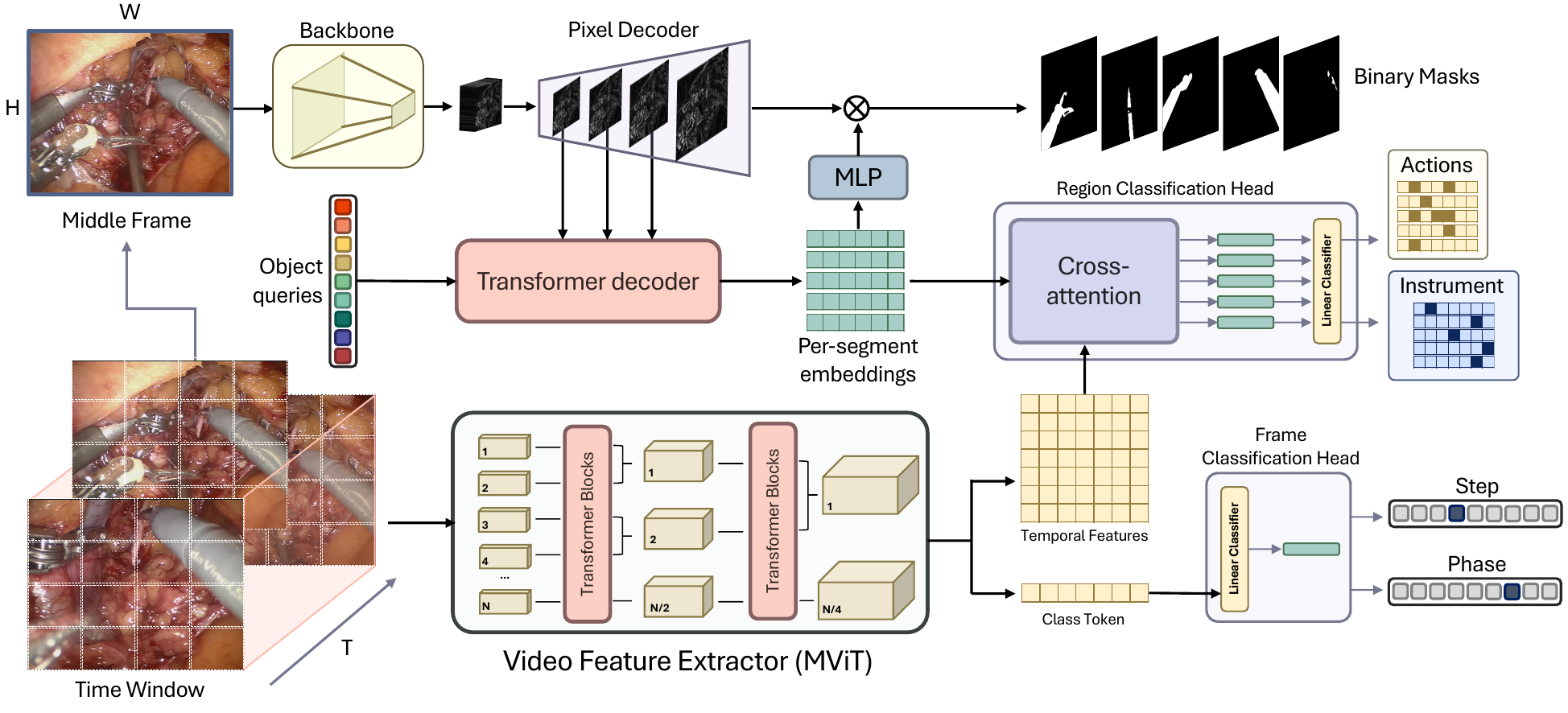}
    \caption{\textbf{TAPIS} utilizes Mask2Former~(\cite{cheng2021mask2former}) as its instrument segmentation baseline and MViT~(\cite{mvit}) as its video feature extractor. On the one hand, the segmentation baseline takes an input frame and computes a set of per-segment embeddings by cross-attending learnable object queries with multi-scale features extracted with a visual backbone and processed with a pixel decoder. Then, the model generates a set of binary masks with the dot multiplication between the linearly transformed per-segment embedding with pixel features from a pixel decoder. On the other, the video feature extractor uses a time window centered on the keyframe used for segmentation and computes a sequence of spatio-temporal embeddings by patchifying the input video and applying multiple stages made by blocks of Multi Head Pooling Attention. These stages progressively reduce the sequence length and increase the number of features. For the long-term tasks, the \textit{frame classification head} linearly classifies the learnable class token of the video feature extractor into either a phase or a step category. For the short-term tasks, a region classification head cross-attends the precalculated segment embeddings from the instrument segmentation with the spatio-temporal features, and then they are linearly classified into instruments or action classes. The figure is best viewed in color.}
    \label{fig:method}
\end{figure*}

We propose the Transformers for Actions, Phases, Steps, and Instrument Segmentation (TAPIS) model, a generalized architecture designed to tackle all the proposed tasks in the GraSP benchmark. Figure~\ref{fig:method} portrays the general architecture of TAPIS. Our method utilizes a localized instrument segmentation baseline applied on independent keyframes that acts as a region proposal network and provides pixel-precise instrument masks and their corresponding segment embeddings. Further, our model uses a global video feature extractor on time windows centered on a keyframe to compute a class embedding and a sequence of spatio-temporal embeddings. A \textit{frame classification head} uses the class embedding to classify the middle frame of the time window into a phase or a step, and a \textit{region classification head} interrelates the global spatio-temporal features with the localized region embeddings for atomic action prediction or instrument region classification. In the following subsections, we explain the details of our proposed architecture.

\subsection{Instrument Segmentation Baseline}
\label{segmentation-baseline}

We follow MATIS's aproach~(\cite{ayobi2023matis}) and adopt Mask2Former~(\cite{cheng2021mask2former}) as our primary instrument segmentation baseline and region proposal method. Generally, Mask2Former employs a transformer decoder that cross-attends a set of $\mathcal{N}$ learnable object queries with image features extracted from a backbone, therefore transforming these queries into $\mathcal{N}$ per-segment embeddings. Additionally, Mask2Former uses a Multi-Scale Deformable Attention pixel decoder~(\cite{def_detr}) with upsampling layers to compute multi-scale per-pixel features to feed into the transformer decoder.
Mask2Former replaces conventional global cross-attention with masked attention, restricting attention to relevant regions using binary masks generated from the pixel embeddings and the transformed object queries. The output segment embeddings are then projected into class probabilities through a linear classifier, while a multi-layer perceptron (MLP) transforms the embeddings into mask embeddings. These mask embeddings are combined with the per-pixel features via dot multiplication to generate the final binary masks. Figure~\ref{fig:method} illustrates this methodology at the top.

We refer the reader to the original Mask2Former~(\cite{cheng2021mask2former}) and MATIS~(\cite{ayobi2023matis}) papers for further details on the Mask2Former architecture and its adaptation for surgical instrument segmentation. We explain how TAPIS utilizes Mask2Former's regions for atomic action detection and instrument reclassification in Section~\ref{tapis}. Finally, we follow \cite{ayobi2023matis} and refer to our instrument segmentation baselines as \textit{TAPIS Frame} in our results tables.

\subsection{Video Feature Extractor}
\label{phases-steps-recognition}

Following our previous work, we employ MViT~(\cite{mvit}) as our video feature extractor to capture intricate details across various space-time scales by leveraging transformers. Once again, we briefly review MViT and refer the reader to the original paper for further details~(\cite{mvit}).

MViT is based on consecutive \textit{stages} of transformer blocks. This architecture first partitions the image into overlapping patches and, through each stage, contracts the spatio-temporal dimensions (sequence length/resolution) while expanding the channel dimensions (features length). To achieve this, MViT uses  Multi-Head Pooling Attention (MHPA) that pools queries, keys, and values at the beginning of each stage to reduce the spatio-temporal resolution before computing attention, while a linear projection at the end of each stage increases the channel dimension. This process creates an internal feature pyramid network, with the initial stages operating at more detailed resolutions with minor features, and the final stages processing shorter sequences with complex spatio-temporal features. Additionally, MViT includes a learnable class token to act as a single embedding of the entire sequence for classification tasks.

We utilize MViT's classification nature for phase and step recognition by classifying an entire time window with the class of the middle frame. For this purpose, we use a \textit{frame classification head} that employs a linear classifier on the output class token of the video features extractor to derive the final Phase or Step class probability distribution. Figure~\ref{method} portrays the video feature extractor and the \textit{frame classification head} at the bottom. The following section extends how we use the remaining embeddings for region classification and action recognition.

\begin{table*}[ht]
\caption{\textbf{Performance of TAPIS in GraSP.} We report the results of TAPIS on the test set of GraSP, and we compare its performance with a CNN-based baseline (SlowFast), the TAPIR model proposed in \cite{valderrama2022tapir} and a Video Swin Transformer-based implementation (TAPIS-VST). We present the number of parameters (Params.) and floating point operations (FLOPs) of each model without counting the region proposal network. The best results are shown in bold.}
\centering
\resizebox{\textwidth}{!}{
\begin{tabular}{ccc|cc|cc|c|ccccc}
\hline
\multirow{2}{*}{Method} & \multirow{2}{*}{FLOPS (G)} & \multirow{2}{*}{Params (M)} &\multicolumn{2}{c|}{Phases} & \multicolumn{2}{c|}{Steps} & Atomic Actions & \multicolumn{5}{c}{Instrument Segmentation} \\
\cline{4-13} 
& && $mAP$ & \textit{F1 score} &  $mAP$ & \textit{F1 score} & $mAP@0.5IoU_{box}$ & $mAP@0.5IoU_{box}$ & $mAP@0.5IoU_{segm}$ & mIoU & IoU & mcIoU \\ \hline
SlowFast  & 81 & 33 & 68.77 & 54.70 & 46.35 & 37.08 & 22.01 & 74.33 & 71.32 & 77.16 & 72.26 & 58.75 \\
TAPIR     & 71 & 36 & 72.59 & 58.62 & 50.24 & 43.82 & 25.57 & 74.43 & \xmark & \xmark & \xmark & \xmark                \\
TAPIS-VST & 66 & 96 & 76.24 & 60.80 & 50.74 & 41.94  & 33.10 & \textbf{90.29} & \textbf{89.58} & 86.36 & \textbf{83.51} & \textbf{77.54} \\
TAPIS     & 71 & 44 & \textbf{76.72} & \textbf{63.42} & \textbf{52.01} & \textbf{45.78} & \textbf{39.26} & 89.85 & 89.10 & \textbf{86.61} & 83.38 & 77.42 \\ \hline
\end{tabular}}
\label{tab:test_results}
\end{table*}

\subsection{Region Classification Head} 
\label{tapis}

We introduce a novel \textit{region classification head} for instrument classification and atomic action detection. Our \textit{region classification head} builds upon the box classification head proposed in \cite{valderrama2022tapir}. In this work, we leverage transformers' cross-attention to correlate region and temporal features. Specifically, we use the per-segment embeddings of the proposed regions as queries in a cross-attention layer. This layer performs Multi-Head Attention over the entire sequence of spatio-temporal features from the video feature extractor, which serve as the keys and values. Consequently, this layer can fully exploit the spatio-temporal resolution to enrich the region features with relevant temporal context. Finally, we use a linear classification layer to project the enriched region features into either instrument or atomic action class probability distributions.

Our region classification head can be agnostic to the specific region proposal method and thus we experimented with multiple Mask2Former-like models~(\cite{def_detr,zhang2023dino,cheng2021mask2former}). Nevertheless, we use our instrument segmentation baseline as the main region proposal network to obtain pixel-wise instrument instance region proposals and their per-segment embeddings. Figure~\ref{fig:method} shows this head in the upper-right part.

\subsection{Implementation Details}

All versions of TAPIS involving the video feature extractor were trained on 16-frame time windows using a cosine learning rate scheduler~(\cite{loshchilov2016cosine}) and an SGD optimizer, and we conducted all experiments on 4 NVIDIA Quadro RTX 800 GPUs. We adopt the following three-stage training approach, portrayed in Figure~\ref{fig:training_dragram} of the Supplementary Material:

\noindent \textbf{1. Multi-Task Long-Term Recognition:} We trained the video feature extractor to perform simultaneous phase and step recognition using separate classification heads for each task and a combined cross-entropy loss. We initialized MViT with its pretrained weights from Kinetics 400~(\cite{kinetics}) and trained it for 30 epochs on temporal windows centered on all GraSP frames sampled at 1fps. We use a window stride of 1s, a batch size of 24, a base learning rate of $1e-1$, an end learning rate of $1e-3$, and 5 warm-up epochs.

\noindent \textbf{2. Instrument Segmentation Baseline (TAPIS Frame):} 
We independently trained Mask2Former for frame-wise instrument segmentation on the annotated keyframes using its pretrained weights on MS-COCO~(\cite{lin2014coco}) with Swin Large~(\cite{swin}) backbone and a batch size of 8. The training spanned 100 epochs using an ADAMW optimizer~(\cite{loshchilov2017adamw}) with a base learning rate of $1 e-5$ decayed by a factor of 0.1 at epochs 50 and 75. We selected the top 5 scoring regions with confidence scores greater than 0.1 for inference and region proposal.

\noindent \textbf{3. Short-Term Task Fine-tunning:} Using the pretrained weights from phase and step recognition, we fine-tuned TAPIS for instrument reclassification and atomic action detection. We froze the instrument segmentation baseline and used its precalculated instrument regions for both tasks. Both tasks were trained on all the keyframes annotated for short-term tasks as the window's central frame, with batch size 24. For instrument classification, we trained TAPIS for 20 epochs with a window stride of 0.5 seconds, a batch size of 24, a base learning rate of $1.25e-2$, and an end learning rate of $1e-4$. For atomic action detection, training spanned 50 epochs using a window stride of 0.033s, a binary cross-entropy loss, a base learning rate of $1.25e-1$, and an end learning rate of $1e-2$. \\

\noindent\textbf{Benchmarking:} We performed all optimization and design experiments in the defined cross-validation folds. For testing, we retrained our best-performing models on the combined training set (fold one and two) and evaluated the test set to fully utilize the available training data.
\label{method}

\section{Experiments and Results}Following the experimental setup in~\cite{valderrama2022tapir}, we compare TAPIS with two alternative models: SlowFast~(\cite{slowfast}) as a CNN-based baseline using ISINet~(\cite{isinet}) for region proposals, and an alternative implementation called TAPIS-VST, which replaces TAPIS's video feature extractor with the Video Swin Transformer (VST)~(\cite{vst}). The results in Table \ref{tab:test_results} demonstrate that both TAPIS and TAPIS-VST outperform the CNN-based baseline. This finding highlights the advantages of fully transformer-based approaches, as their enhanced modeling capabilities, powered by long-range attention mechanisms, improve the understanding of complex surgical scenes. Additionally, TAPIS surpasses TAPIS-VST in most tasks while achieving comparable instrument segmentation performance with significantly fewer parameters. We attribute TAPIS-VST's slight advantage in instrument region classification to its finer-grained spatial processing and a larger number of parameters.

Table~\ref{tab:test_results} compares TAPIS with the TAPIR model~(\cite{valderrama2022tapir}), which we trained using its original implementation details and box proposal network on GraSP's training set. TAPIS significantly outperforms TAPIR across all tasks, achieving a 53\% relative improvement in the challenging atomic action detection task. This performance difference validates the impact of our enhanced instrument segmentation annotations and our technical and methodological advancements. Notably, TAPIR exhibits a slight decrease in performance for short-term tasks compared to the initially reported metrics\footnote{\url{https://github.com/BCV-Uniandes/TAPIR}}, which we attribute to the curation of our dataset that eliminated biases and included annotations for previously unannotated and challenging tiny instrument segments. 

In the following sections, we validate TAPIS by comparing it with publicly available state-of-the-art (SOTA) models designed for specific single tasks similar to those in GraSP. For this purpose, we evaluate TAPIS, its alternative baselines, and these SOTA models on our cross-validation set and on alternative benchmarks featuring tasks analogous to GraSP. To ensure a fair comparison, we train each SOTA model independently for its respective task and adapt all temporal models for offline processing to align with TAPIS's framework. These comparisons also assess the consistency of our GraSP dataset by analyzing how the performance ranking of various models on external benchmarks aligns with their performance on GraSP and the expected performance ranking from the current SOTA.

\begin{table}[t]
\caption{\textbf{Comparative Results of TAPIS in Instrument and Action Presence Recognition in GraSP.} We report the results of our model on frame-wise prediction of instrument presence and atomic action presence, regardless of location or instance count. The best results are shown in bold.}
\centering
\resizebox{\linewidth}{!}{
\begin{tabular}{ccc|cc} \hline
\multirow{2}{*}{Model} & \multirow{2}{*}{Regions} & \multirow{2}{*}{Time} & \multicolumn{2}{c}{Presence $mAP$ (\%)} \\ \cline{4-5} 
&&&  Instruments & Actions   \\  \hline
Rendezvous~(\cite{nwoye2021rendezvous}) & \ding{55}  & \ding{55}  & 80.41 & 20.65 \\ 
RiT~(\cite{nwoye2023rit}) & \ding{55}  & \checkmark  & 78.20 & 21.76 \\\hline
\multirow{4}{*}{TAPIS (Ours)}           & \ding{55}  & \ding{55}  & 53.19 & 24.77 \\
                                        & \ding{55}  & \checkmark & 84.76 & 41.26 \\
                                        & \checkmark & \ding{55}  & 93.54 & 31.04 \\
                      & \checkmark & \checkmark & \textbf{94.33} & \textbf{43.85} \\ \hline
\end{tabular}
}
\label{tab:instrument_action_recognition_results}
\end{table}

\begin{table}[t]
\caption{\textbf{Comparative results of TAPIS in Instrument and Interactions Detection on the EndoVis 2018 dataset.} We test our TAPIS model in the benchmark proposed in \cite{islam2020interactions} and compare our performance with previously proposed methods. The best results are shown in bold.
}
\centering
\resizebox{\linewidth}{!}{
\begin{tabular}{c|cc} \hline
\multirow{2}{*}{Model} & \multicolumn{2}{c}{$mAP@0.5IoU_{box}$ (\%)} \\ \cline{2-3} 
&  Instruments & Interactions   \\  \hline
MSLRGR-GISFSG~(\cite{islam2022interactions})  &  42.59      & 39.11          \\
\textbf{TAPIS (Ours)} & \textbf{63.55} & \textbf{41.97}   \\ \hline
\end{tabular}}
\label{tab:interactions_results}
\end{table}

\subsection{Validation of TAPIS in Public Benchmarks}
\label{validation_experiments}

\subsubsection{Instruments and Actions Presence Recognition}
\label{sec:presence_recognition}

We adapt our benchmark and methods for simpler, frame-level tasks such as instrument and action binary presence recognition, following the conventional formulations in~\cite{endonet-cholec80,nwoye2021rendezvous,heico,wagner2023heichole}. We modify TAPIS for this task by using our best instrument segmentation baseline and consolidating predictions from all proposed regions into a multi-label binary classification vector.

We compare TAPIS with two groups of baseline methods for this task. First, we adapt Rendezvous (\cite{nwoye2021rendezvous}) and RiT (\cite{nwoye2023rit}) on our benchmark by combining all instrument categories, along with the "null verb" and "null target" placeholders and using all action categories combined with the "instrument" and "null target" placeholders. Consequently, we train Rendezvous and RiT to recognize the presence of instruments and actions in our keyframes. Second, we test alternative implementations of TAPIS without region proposal features, by directly predicting instrument and action presence from the video feature extractor; and without temporal processing by running TAPIS on a single frame. We train all models without pretraining on any surgical datasets to ensure a fair comparison, and we assess performance using the $mAP$.

The results in Table~\ref{tab:instrument_action_recognition_results} demonstrate that TAPIS significantly outperforms its implementations without region proposals or temporal processing and achieves superior performance compared to the Rendezvous and RiT models for both instrument and action presence recognition. These results underscore the value of GraSP's spatial annotations, as including localized information about instruments and actions enhances the model's ability to understand and process the surgical scene. Additionally, TAPIS's temporal component further improves recognition by incorporating valuable temporal context that enables the model to better capture dynamic interactions and temporal progressions within surgical workflows.

\begin{table}[t]
\caption{\textbf{Comparative results of TAPIS on the gesture recognition task of the RARP-45 dataset}. We compare the performance of TAPIS with SOTA models for phase recognition, and its alternative baseline architectures on gesture recognition in the dataset proposed by \cite{amsterdam2022rarp45}. The best results are shown in bold.}
\centering
\resizebox{\linewidth}{!}{
\begin{tabular}{ccc|cc}\hline
\multirow{2}{*}{Method} & \multirow{2}{*}{FLOPs (G)} & \multirow{2}{*}{Param (M)} & \multicolumn{2}{c}{Gesture Recognition (\%)} \\ \cline{4-5} 
                   &        &       & \textit{mAP}   & \textit{F1 score}\\ \hline
TeCNO                 & 4.50   & 23.64 & 43.08          &  45.94           \\
Trans-SVNet           & 11.96  & 23.67 & 42.15          &  42.85           \\
SAHC                  & 11.95  & 25.73 & 43.79          &  41.80           \\
SlowFast              & 25.29  & 33.67 & 49.00          &  46.79           \\
TAPIS-VST             & 65.76  & 87.66 & 52.65          &  48.25           \\
\textbf{TAPIS (Ours)} & 71.80  & 36.31 & \textbf{57.25} &  \textbf{54.17}           \\ \hline
\end{tabular}}
\label{tab:RARP-45_comparison}
\end{table}

\begin{table}[t]
\caption{\textbf{Comparative results of TAPIS in phases and steps recognition on the MISAW dataset}. We compare the performance of TAPIS with SOTA models for phase recognition, and its alternative baseline architectures on Phase and Step recognition in the benchmark proposed by \cite{misaw}. The best results are shown in bold.}
\centering
\resizebox{\linewidth}{!}{
\begin{tabular}{ccc|cccc}
\hline
\multirow{2}{*}{Method} & \multirow{2}{*}{FLOPS (G)} & \multirow{2}{*}{Params. (M)} & \multicolumn{2}{c}{Phases  (\%)} & \multicolumn{2}{c}{Steps   (\%)}\\ \cline{4-7} 
& & & \textit{mAP}                 & \textit{F1 score}          & \textit{mAP}                 & \textit{F1 score}       \\ \hline
TeCNO       & 4.50  & 23.64 & 95.58 & 62.83 & 68.89 & 62.99  \\
Tras-SVNet  & 11.96 & 23.67 & 90.38 & 88.17 & 72.44 & 63.07  \\
SAHC        & 11.95 & 25.73 & 95.10 & \textbf{92.43} & \textbf{78.79} & \textbf{67.88}  \\
SlowFast    & 25.29 & 33.67 & 94.04 & 81.98 & 70.50 & 62.48 \\
TAPIS-VST   & 65.76 & 87.66 & 96.16 & 89.71 & 73.46 & 63.36 \\
\textbf{TAPIS} & 70.80 & 36.31 & \textbf{97.14} & 90.12 & 77.52 & 66.51 \\ \hline
\end{tabular}}
\label{tab:misaw_validation}
\end{table}

\begin{table*}[t]
\caption{\textbf{Comparative results of TAPIS in phase and step recognition on GraSP's cross-validation set}. We evaluate the TAPIS trained for phase and step recognition jointly (multi-task) and independently, and compare it with alternative baseline and SOTA models for surgical phase recognition trained independently. The best results are shown in bold.}
\centering
\resizebox{\linewidth}{!}{
\begin{tabular}{cccc|cccc|cc}\hline
\multirow{2}{*}{Model} & \multirow{2}{*}{Multi-Task}  & \multirow{2}{*}{FLOPs (G)} & \multirow{2}{*}{Params. (M)} & \multicolumn{2}{c}{Phase Recognition (\%)} & \multicolumn{2}{c|}{Step Recognition (\%)}  & \multicolumn{2}{c}{Long-Term Average (\%)}            \\  \cline{5-10}
& & & & $mAP$ & \textit{F1 score} & $mAP$ & \textit{F1 score} & $mAP$ & \textit{F1 score} \\ \hline
TeCNO~(\cite{czempiel2020tecno})  & \ding{55} &  4.50    & 23.64     & 64.64 $\pm2.88$ & 61.51 $\pm2.73$ & 27.39 $\pm1.78$ & 22.70 $\pm2.21$ & 46.01 $\pm21.59$ & 42.11 $\pm22.50$ \\
Tran-SVNet~(\cite{transvnet})   & \ding{55} & 11.96     & 23.67    & 66.89 $\pm1.79$ & 58.20 $\pm3.80$ & 33.96 $\pm2.02$ & 30.31 $\pm2.22$ & 50.42 $\pm19.08$ & 44.25 $\pm16.30$ \\
SAHC~(\cite{ding2021sahc})  & \ding{55} & 11.95     & 25.73     & 70.80 $\pm1.76$ & \textbf{66.35} $\pm1.60$ & 37.55 $\pm1.08$ & 35.85 $\pm1.24$ & 54.17 $\pm19.23$ & 50.10 $\pm18.80$ \\
SlowFast  & \ding{55} & 25.29     & 33.67    & 68.44 $\pm0.10$ & 58.59 $\pm3.06$ & 43.72 $\pm1.79$ & 40.29 $\pm3.81$ & 56.15 $\pm14.23$ & 49.44 $\pm10.94$ \\ \hline
\multirow{2}{*}{\textbf{TAPIS-VST}}  & \ding{55} & 65.76     & 87.66     & 70.46 $\pm0.09$ & 62.25 $\pm0.33$ & 45.51 $\pm1.48$ & 39.45 $\pm0.87$ & 57.98 $\pm14.43$ & 50.85 $\pm13.17$ \\
  & \checkmark & 65.76     & 87.68      &  70.60 $\pm1.59$ &  61.64 $\pm3.32$ &  47.17 $\pm2.65$ &  42.25 $\pm3.72$ & 58.89 $\pm13.64$ & 51.95 $\pm11.56$ \\ \hline
\multirow{2}{*}{\textbf{TAPIS}}   & \ding{55} & 70.80     & 36.31     & \textbf{72.31} $\pm2.04$ & 64.04 $\pm2.13$ & 49.72 $\pm2.35$ & 45.57 $\pm2.03$ & 61.01 $\pm13.17$ & 54.81 $\pm10.80$ \\
  & \checkmark & 70.80     & 36.32      &  71.36 $\pm1.33$ &  64.72 $\pm1.92$ &  \textbf{50.74} $\pm2.53$ &  \textbf{46.91} $\pm3.36$ & \textbf{61.05} $\pm12.02$ & \textbf{55.82} $\pm10.52$ \\
\hline
\end{tabular}}
\label{tab:sota_comparison}
\end{table*}

\subsubsection{Interactions Recognition in Endovis 2018}
\label{sec:interactions_endovis2018}

We evaluate TAPIS on the EndoVis 2018 interaction dataset (\cite{islam2020interactions}), following its established splits. We exclude the \textit{stapler} class, which only appears in a single sequence, and omit the \textit{kidney} class, as TAPIS does not incorporate tissue information. For this experiment, we replace Mask2Former with DINO~(\cite{zhang2023dino}) as the region proposal network, trained for instrument detection using the annotations from~\cite{islam2020interactions}. Additionally, we train TAPIS without pretraining the video feature extractor on surgical datasets. TAPIS is then jointly trained for instrument and interaction detection by combining their respective losses, weighted at 0.95 for action loss and 0.05 for instrument loss.

We compare TAPIS with MSLRGR-GISFSG~(\cite{islam2022interactions}) by evaluating their instrument and action predictions using the $mAP@0.5IoU_{box}$ metric. Since MSLRGR-GISFSG relies on ground-truth bounding boxes for inference, we assume perfect bounding box locations to evaluate its performance, assigning instrument classes based on predicted masks and assessing interactions using each bounding box's predicted interaction with the kidney.

As shown in Table~\ref{tab:interactions_results}, TAPIS significantly outperforms MSLRGR-GISFSG in both instrument localization and interaction detection tasks, proving the robustness of our transformer-based approach for surgical localization tasks. Despite the dataset's graph structure, TAPIS effectively integrates global features from the video feature extractor with local detections from DINO, demonstrating its capability to model complex instrument-scene relationships across diverse datasets.

\subsubsection{Gesture Recognition in RARP45}
\label{sec:gestures_rarp45}

We validate TAPIS on the RARP-45 benchmark for gesture recognition. Given the absence of a public test set in the original RARP-45 dataset, we create data splits as detailed in \ref{sec:appendix:results} of the Supplementary Material. Following our proposed experimental framework, we train our video feature extractor on RARP45 and compare it with SlowFast, TAPIS-VST, and multiple SOTA models designed for surgical phase recognition~(\cite{czempiel2020tecno,transvnet,ding2021sahc}). Table~\ref{tab:RARP-45_comparison} shows that TAPIS outperforms alternative baselines, with all transformer-based methods achieving the best results. Similarly, our model strongly surpasses SOTA models for phase recognition due to the finer-grained nature of the gesture recognition task for which these architectures have not been designed. This behavior shows TAPIS's adaptability to tasks analogous to those proposed in GraSP yet differently formulated.

\subsubsection{Phase and Step Recognition in MISAW}
\label{sec:phases_n_steps_misaw}

To further validate our phase and step recognition benchmark, we evaluate TAPIS on the MISAW dataset~(\cite{misaw}) using MISAW's predefined dataset splits and follow the comparative experimental setup described in Section~\ref{sec:gestures_rarp45}. The results in Table~\ref{tab:misaw_validation} show that TAPIS outperforms most SOTA models and its alternative baselines, reinforcing the advantages of transformers for surgical video analysis and demonstrating the consistency of our approach.

Nevertheless, TAPIS slightly underperforms SAHC, and we attribute this to MISAW's simpler formulation of phase and step recognition tasks, which are characterized by fewer categories, shorter simulated videos, and longer, uninterrupted temporal segments. In contrast, GraSP includes multiple phase and step categories with highly variable and interrupted segment durations on long \textit{in-vivo} videos. Thus, models like SAHC, which process extended temporal contexts of global frame-wise features and are tailored for datasets with similar phase recognition setups (e.g., Cholec80~(\cite{endonet-cholec80}), M2CAI16~(\cite{stauder2016lapchole})), may be better suited to MISAW's data. Nonetheless, despite being optimized for GraSP's more complex setup, TAPIS achieves competitive performance and demonstrates its generalization ability across diverse benchmarks.

\subsection{Design and Ablation Experiments}
\subsubsection{Long-Term Tasks Experiments}

Table~\ref{tab:sota_comparison} presents the performance of TAPIS and TAPIS-VST trained for phase and step recognition, independently and jointly. To directly compare models, we calculate the average Long-Term performance as the average per-metric across all tasks and folds. Even when trained independently, TAPIS outperforms alternative models on both tasks and demonstrates better long-term performance, while only SAHC achieves comparable performance in phase recognition. Additionally, TAPIS shows a significant advantage over SOTA models in step recognition, thus highlighting limitations in phase recognition models, which lack versatility for more granular tasks. In contrast, TAPIS's single-stage, video-wise encoder captures finer spatio-temporal details that allow better performance in finer temporal tasks. Similarly, the relative performance ranking of the SOTA models aligns with current phase recognition benchmarks, confirming our benchmark's validity.
Finally, TAPIS yields better results when trained jointly for phase and step recognition, with the most notable improvements observed in the challenging step recognition task. This demonstrates TAPIS's multi-task learning capabilities, allowing it to leverage task correlations and perform both tasks within a single training run.

\begin{table*}[ht]
\caption{\textbf{Comparative results of TAPIS in Instrument Segmentation in the cross-validation set}. We showcase the instance and semantic segmentation performance of our instrument segmentation baseline (\textit{TAPIS Frame}) and our complete model (\textit{TAPIS Full}), and we compare it to our CNN baseline and previous state-of-the-art models for instrument segmentation. The parameters and floating point operations reported for the \textit{TAPIS Full} models correspond to the video feature extractor and the region classification head while the values reported for \textit{TAPIS Frame} correspond to Mask2Former. The best results are shown in bold.}
\resizebox{\textwidth}{!}{
\begin{tabular}{cccc|ccccc}
\hline
\multirow{2}{*}{Model} & \multirow{2}{*}{Backbone} & \multirow{2}{*}{FLOPs (G)} & \multirow{2}{*}{Params. (M)} & \multicolumn{2}{c}{Instance Segmentation Metrics (\%)} & \multicolumn{3}{c}{Semantic Segmentation Metrics (\%)}              \\ \cline{5-9} 
& & & & $mAP@0.5IoU_{box}$& $mAP@0.5IoU_{segm}$ & mIoU & IoU & mcIoU \\ \hline
TernausNet~(\cite{shvets2018ternausnet}) & UNet 16 & - & - & \xmark & \xmark & 41.74 $ \pm5.07$ & 24.46 $ \pm6.04$ & 16.87 $ \pm3.70$ \\
MF-TAPNet ~(\cite{yueming2019mftapnet}) & TAPNet & - & - & \xmark & \xmark & 66.63 $ \pm1.24$ & 29.23 $ \pm1.43$ & 24.98 $ \pm0.59$ \\
ISINet   ~(\cite{isinet}) & R50 & 201 & 44 & 79.85 $ \pm2.17$ & 78.29 $ \pm2.82$ & 78.44 $ \pm1.13$ & 70.85 $ \pm0.00$ & 56.67 $ \pm1.46$ \\
QPD Mask DINO~(\cite{rohan2023qpd}) & R50 & 228 & 54 & 88.46 $ \pm1.14$ & 87.39 $ \pm1.75$ & 83.89 $ \pm1.23$ & 82.56 $ \pm1.14$ & 74.36 $ \pm1.04$ \\ \hline
TAPIS Frame & \multirow{2}{*}{R50} & 226 & 44 & 89.65 $ \pm1.26$ & 88.65 $ \pm1.43$ & 84.81 $\pm1.62$ & 81.34 $ \pm1.44$ & 73.48 $ \pm0.88$ \\
TAPIS Full & & 71 & 44 & 88.66 $ \pm0.79$ & 87.20 $ \pm1.12$ & 84.76 $ \pm1.63$ & 81.64 $ \pm1.45$ & 74.43 $ \pm0.79$ \\ \hline
TAPIS Frame & \multirow{2}{*}{SwinL} & 868 & 216 & \textbf{92.65} $ \pm1.57$ & \textbf{91.71} $ \pm1.72$ & 86.91 $ \pm1.59$ & 83.92 $ \pm0.68$ & 77.59 $ \pm0.08$ \\
TAPIS Full & & 71 & 44 & 91.72 $ \pm0.87$ & 90.34 $ \pm1.11$ & \textbf{87.05} $ \pm1.63$ & \textbf{84.45} $ \pm0.72$ & \textbf{78.82} $\pm 0.88$ \\ \hline
\end{tabular}}
\label{tab:instrument-segmentation}
\end{table*}

\begin{table}[t]
\caption{\textbf{Comparative results in Instrument Detection in the cross-validation set.} We test multiple object detection architectures using the bounding boxes generated by GraSP's segmentation annotations. We also report the number of parameters (Params.) and floating point operations (FLOPs) of each model. The best results are shown in bold.}
\centering
\resizebox{\linewidth}{!}{
\begin{tabular}{lccc|c}
\hline
Model & Backbone & FLOPs (G)  & Params (M)   &  $mAP@0.5IoU_{box}$ \\ \hline
Faster R-CNN & R50 & 180 &   42     & 80.88 $\pm 1.16$ \\
Deformable DETR & R50 & 173 & 40 &  87.33 $\pm 2.91$ \\ 
DINO & SwinL & 310 & 218 &  \textbf{90.93} $\pm 1,15$ \\\hline
\end{tabular}}
\label{tab:detection}
\end{table}

\begin{figure*}[ht]
    \centering
    \includegraphics[width=\textwidth]{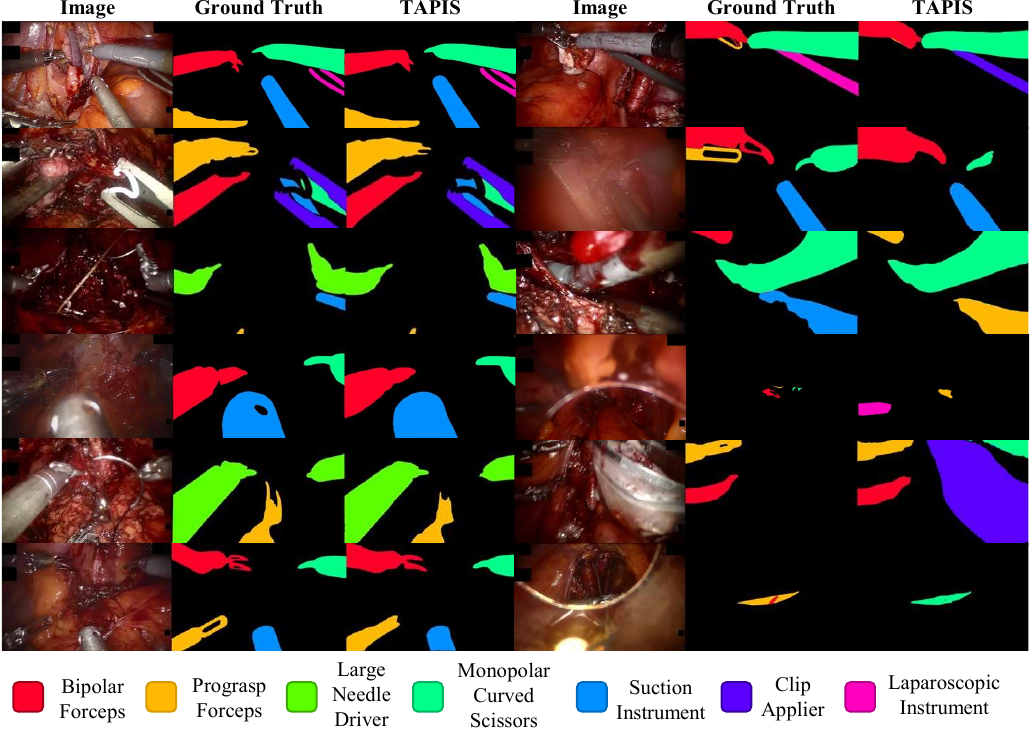}
    \caption{\textbf{Qualitative Results of TAPIS in Instrument Segmentation.} We present twelve examples of our best model's segmentation results. The first three rows on the left show nearly perfect results, the next three rows on the left show average results, the first three rows on the right show slight segmentation errors and the last three rows show significant segmentation errors.}
    \label{fig:visu_segm_principal}
\end{figure*}

We present per-class performance results for phase and step recognition in Tables~\ref{tab:per_class_phase_performance} and \ref{tab:per_class_step_performance} of the Supplementary Material. For phases, classes 1–6 and 10 consistently achieve higher AP scores, while classes 7 and 8, characterized by lower representation in the dataset and shorter mean durations, exhibit lower scores. Similarly, step categories with lower durations, such as steps 12–18, or those with sparse representation, like steps 6, 7, and 11, show lower AP scores, while steps 0, 1, and 19 perform well due to higher representation. Notwithstanding, some highly represented classes like steps 2, 3, and 5 still underperform, likely due to high variability in their durations. 

\subsubsection{Instrument Segmentation Experiments}

We trained and optimized multiple object detectors on our cross-validation set to assess the quality of our enhanced bounding box annotations and establish reference points for instance segmentation. The results in Table~\ref{tab:detection} compare Faster R-CNN~(\cite{ren2015faster}) as a CNN-based detector with Deformable DETR~(\cite{def_detr}) and DINO~(\cite{zhang2023dino}) with its largest backbone (Swin Large~(\cite{swin})) as transformer-based detectors. Transformer detectors consistently surpass CNN-based detectors, with DINO achieving the best performance. These results are consistent with our initial results~(\cite{valderrama2022tapir}) and with the general state-of-the-art in object detection, thus showing the reliability of our data and its applicability for instrument detection tasks.

We validate our segmentation benchmark by comparing various configurations of TAPIS  with publicly available methods for instrument segmentation (\cite{shvets2018ternausnet,yueming2019mftapnet,isinet,rohan2023qpd}).  
As shown in Table~\ref{tab:instrument-segmentation}, we assess the performance of our sole instrument segmentation baseline (referred to as \textit{TAPIS Frame}) using different backbones, including ResNet50 for a fair comparison with alternative methods and Swin Large (SwinL). Transformer-based methods consistently outperform CNN-based models, improving performance as the backbone complexity increases. Notably, all TAPIS Frame configurations surpass their detection counterparts in $mAP@0.5IoU_{box}$, with even the smallest TAPIS Frame setup achieving performance comparable to the largest detection models while requiring significantly fewer parameters and FLOPs. These results demonstrate how our segmentation annotations support training pixel-level models to capture finer visual details that enhance instrument localization and recognition. Moreover, the relative performance of the compared methods aligns with trends in the current state of the art, with TAPIS achieving the best performance.

Table~\ref{tab:instrument-segmentation} also presents the results of the entire TAPIS model for instrument segmentation (called \textit{TAPIS Full}), which integrates \textit{region classification head} with \textit{TAPIS Frame}'s region proposals. Including the \textit{region classification head} enhances the stringent semantic segmentation metrics but decreases the $mAP@0.5IoU_{segm}$. This trade-off arises from the head's ability to reclassify more prominent regions over time while occasionally misclassifying smaller, momentary regions. Throughout our experimentation, we noted minimal variance in instrument classification performance related to the specific design of the \textit{region classification head} design.  

We detail the per-class instrument segmentation results in Tables~\ref{tab:class-instance-segmentation} and \ref{tab:class-semantic-segmentation} of the Supplementary Material. Performance strongly correlates with instrument frequency in the dataset and their visual characteristics. Instruments like the BF, MCS, and LND demonstrate high accuracy due to their distinctive shafts and tips and their frequent, unobstructed foreground visibility. In contrast, the PF exhibits lower performance as it is frequently occluded and often appears as a smaller or distant object within the frame. At the same time, the LI faces challenges from high intra-class variability as it is a generic category for external laparoscopic instruments. These variations show the complexity of surgical instrument segmentation and the importance of leveraging both visual and temporal context to achieve accurate results.

Figure~\ref{fig:visu_segm_principal} shows examples of qualitative results of TAPIS for instrument segmentation, and we present more examples in Figures~\ref{fig:visu_segm_excellent}, \ref{fig:visu_segm_good} and \ref{fig:visu_segm_bad} of the Supplementary Material. The results on the left part of Figure~\ref{fig:visu_segm_principal} demonstrate TAPIS's ability to accurately segment complex images with up to five instrument instances with intricate, varied, and occasionally overlapping figures. Nevertheless, the results on the first three rows on the right show minor defects on instruments occluded by tissue and fluids. The last three rows on the right show inaccurate predictions due to a high degree of overlapping between instances and a complex views of specific instruments. However, we highlight that TAPIS achieves a very high segmentation recall as it can propose a segmentation mask to cover most instances, failing primarily in tiny segments.

\subsubsection{Atomic Action Detection}

We first evaluate the impact of using different region proposal networks for the action detection tasks. The results in Table~\ref{tab:action-regions} demonstrate that atomic action detection performance increases with instrument detection or segmentation performance, as higher-quality region proposals increase instrument location recall and provide richer semantic region features for action recognition. Additionally, using \textit{TAPIS Frame}'s segmentation features with a ResNet50 backbone achieves comparable performance to DINO's detection features with a SwinL backbone, demonstrating the value of segmentation approaches in providing more detailed features with instrument shape cues that improve atomic action comprehension. 

\begin{table}
\caption{\textbf{Comparative results of TAPIS in Atomic Action Detection}. We present the results of employing region features from multiple detection and segmentation models. We also report the number of parameters and floating point operations corresponding to the region proposal networks. The best results are shown in bold.}
\resizebox{\linewidth}{!}{
\begin{tabular}{cccc|c}
\hline
\multicolumn{4}{c|}{Region Proposal Method}                       & \multirow{2}{*}{$mAP@0.5IoU_{box}$ (\%) }        \\ \cline{1-4}
Model                        & Backbone & FLOPs (G) & Params. (M) &                                            \\ \hline
Deformable DETR              & R50      & 173       & 40          & 28.54 $\pm 0.68$                           \\
DINO                         & SwinL    & 310       & 218         & 31.37 $\pm 0.50$                           \\ \hline
\multirow{2}{*}{TAPIS Frame} & R50      & 226       & 44          & 31.24 $\pm 1.84$                           \\
                             & SwinL    & 868       & 216         & \textbf{35.46} $\pm 2.40$ \\ \hline
\end{tabular}}
\label{tab:action-regions}
\end{table}

We assess the effectiveness of our cross-attention-based \textit{region classification head} against the original designs of TAPIR~(\cite{valderrama2022tapir}) and MATIS~(\cite{ayobi2023matis}), with results summarized in Table~\ref{tab:cross-attention}. TAPIR's classification head pools spatio-temporal features and concatenates them with region embeddings (\textit{Time Pooling} in Table~\ref{tab:cross-attention}), while MATIS introduces a Time MLP with additional linear layers around TAPIR's pooling operation (\textit{Time MLP} in Table~\ref{tab:cross-attention}). Our findings reveal that performance scales with increasing parameters, as Time MLP outperforms Time Pooling, consistent with previous results in~\cite{ayobi2023matis}. More importantly, our cross-attention layer significantly outperforms both designs despite having slightly fewer parameters and FLOPs than Time MLP. This enhancement demonstrates the superiority of our approach, as the cross-attention mechanism effectively leverages the full spatio-temporal context from the video feature extractor, avoiding the information loss associated with pooling operations and leading to more robust classification performance.

\begin{table}[t]
\caption{\textbf{Ablation results on the design of the region classification head in the cross-validation set.}  We compare the performance in atomic action detection using the Time Pooling design of \cite{valderrama2022tapir}, the Time MLP from \cite{ayobi2023matis}, and our new Cross-Attention layer design. The best results are shown in bold.}
\resizebox{\linewidth}{!}{
\begin{tabular}{l|ccc}
\hline
\multirow{2}{*}{Region Proposal Model} & \multicolumn{3}{c}{Atomic Actions $mAP@0.5IoU_{box}$} \\ \cline{2-4} 
& Time Pooling &  Time MLP & Cross-Attention \\ \hline
Deformable DETR  & 25.03 $\pm 1.06$  & 26.97
$\pm 0.09$ & \textbf{28.54} $\pm 0.68$  \\
DINO & 26.64 $\pm 2.39$ & 28.64 $\pm 0.34$ & \textbf{31.37} $\pm 0.50$\\
TAPIS Frame (R50)  & 29.61 $\pm 0.72$ & 30.67 $\pm 0.08$ & \textbf{31.24} $\pm 1.84$  \\
TAPIS Frame (SwinL) & 32.48 $\pm 1.23$ & 33.43 $\pm 1.38$ & \textbf{35.46} $\pm 2.40$  \\ \hline
FLOPs (G) & 0.003 & 0.633 & 0.538 \\
Params. (M) & 0.289 & 8.882 & 8.086  \\  \hline
\end{tabular}}
\label{tab:cross-attention}
\end{table} 

We provide the per-class results of the atomic action detection task in Table~\ref{tab:class-action-detection} of the Supplementary Material. The performance once again correlates with the frequency of each action, where actions like \textit{travel}, \textit{hold}, and \textit{still} demonstrate the highest performance metrics. However, performance is also influenced by the complexity and visual variability of actions. Specific actions such as \textit{cauterize} and \textit{suction} record high \textit{AP} values due to their association with distinct instruments and contextual cues from the video feature extractor, such as smoke or blood movement, that enhance their recognition. Conversely, actions like \textit{pull} and \textit{push} often get interchanged, leading to lower performance. Similarly, abstract actions such as \textit{release} or \textit{open something} can present in multiple ways, making it challenging to classify them consistently. These results underscore the complexity of surgical atomic action detection and the importance of capturing nuanced motion patterns to differentiate similar activities.

\begin{table}[]
\caption{\textbf{Ablation results on atomic action fine-tunning}. We evaluate the effect of Long-Term tasks pretraining and using pre-classified region embeddings (Class Regions) for atomic action detection.}
\resizebox{\linewidth}{!}{
\begin{tabular}{ccc|c}
\hline
Model & Class Regions & Long-Term Pretrain & $mAP@0.5IoU_{box}$ \\ \hline
\multirow{2}{*}{Regions MLP} & \ding{55} & \ding{55}   & 20.96 $\pm0.53$   \\ 
                            & \checkmark & \ding{55}  & 22.28 $\pm0.14$  \\ \hline
\multirow{3}{*}{TAPIS}      & \ding{55}  & \checkmark & 33.52 $\pm1.05$  \\ 
                            & \checkmark & \ding{55}  & 34.06 $\pm1.95$   \\ 
                            & \checkmark & \checkmark & \textbf{35.46} $\pm0.34$  \\ \hline
\end{tabular}}
\label{tab:action_generic_ablations}
\end{table}

Table~\ref{tab:action_generic_ablations} evaluates the impact of instrument type segmentation and phase and step recognition signals on atomic action detection. To analyze the role of instrument recognition, we trained TAPIS using unclassified instrument regions obtained from our best segmentation baseline trained to generate generic instrument masks without a classification loss. To assess the impact of independent region features, we trained an MLP directly on region features to predict atomic actions without considering temporal processing (\textit{Regions MLP} in Table~\ref{tab:action_generic_ablations}). Results indicate that region features alone achieve notable performance as they capture rich instrument shape and pose information, facilitating action recognition. Similarly, both the Regions MLP and TAPIS demonstrated improved performance when classified regions were used, proving the importance of instrument type information for atomic action detection. We also compare TAPIS's performance without pretraining the video feature extractor on phase and step recognition and only using MViT's pretraining in kinetics400. The results confirm that long-term task pretraining enhances atomic action detection, highlighting the complementary nature of GraSP's tasks and TAPIS's capability to leverage cross-task correlations for improved performance on individual tasks.

\subsection{Discussion}

TAPIS's strong performance in GraSP and its generalization on alternative frameworks demonstrates its versatility and suitability for multi-granular surgical modeling. The limited adaptability of SOTA models to varying granularities underscores the need for versatile architectures that generalize across task formulations and maintain competitive performance. TAPIS also proved versatility regarding the choice of video feature extractor or region proposal model, delivering competitive results across long- and short-term tasks with different configurations.

Moreover, our results confirm the hypothesized interrelationships among tasks in surgical workflow analysis. First, training long-term tasks simultaneously enhanced their per-task performance, enabling a single model to tackle both tasks while being trained once. In turn, this long-term task pretraining boosted short-term task performance. Similarly,  improved instrument recall, using shape-level features and incorporating instrument class in region embeddings, positively impacted atomic action detection. These findings highlight TAPIS's ability to leverage multi-task signals to enhance individual task performance. Additionally, our results underscore the importance of studying endoscopic vision and surgical workflow analysis tasks as interconnected components rather than isolated objectives. 

Finally, TAPIS's offline nature limits its application in real-time scenarios, confining its immediate utility to retrospective tasks such as data analysis, annotation, or surgical training. However, rather than prioritizing immediate assistive functionality, TAPIS and GraSP introduce groundbreaking approaches to holistic surgical scene understanding by exploiting the inherent complementarity among surgical workflow analysis tasks. This shift transcends the conventional focus on isolated tasks, establishing a foundational framework for future robotic systems capable of complex surgical assistance and cognition through a multi-level understanding of workflows and a fine-grained recognition of surgical agents and actions.
\label{results}

\section{Conclusions}
\label{conlusions}

We introduced a novel dataset and framework for holistic surgical scene understanding with a hierarchy of complementary long-term and short-term tasks with varying levels of temporal and spatial detail. By curating and expanding our previous benchmark and including surgical instrument instance segmentation, we introduced the first Endoscopic Vision dataset with four levels of visual understanding within their highest annotation granularity. Through detailed dataset statistics, we highlighted the challenges presented by GraSP and its representation of real-world complexities in robot-assisted surgical procedures. We also established training, validation, and testing splits to ensure consistent benchmarking.

We developed the TAPIS model to leverage GraSP’s multi-granular annotations. TAPIS surpassed alternative baselines and SOTA models in multiple benchmarks, proving its versatility to varying task granularities and formulations. Our instrument segmentation annotations significantly improved instrument recognition and provided shape-level cues that enhanced atomic action detection. Our experiments also revealed the complementarity among tasks as TAPIS enhanced per-task performance by leveraging multi-task signals.

TAPIS’s transformer-based design consistently outperformed alternative baselines, proving the potential of transformer-based approaches and establishing new SOTA results in our benchmark. Moreover, its consistent performance across public benchmarks validates the reliability of our methodology and dataset. Hence, this work represents a significant step forward in Endoscopic Vision, promoting a multi-granular understanding of surgical procedures and laying the groundwork for holistic surgical scene understanding and image-guided surgical robot intelligence. \\

\noindent\textbf{Acknowledgments:} The authors would like to thank Natalia Valderrama and Paola Ruiz for their contributions to the early stages of this work. We would also like to thank Nicolás Cantillo and Gabriela Monroy for their contribution to data annotation and Dr. Felipe Gomez for his contributions to data collection. Isabela Hernández, Nicolás Ayobi, Alejandra Perez, and Santiago Rodríguez acknowledge the support of the 2021, 2022, and 2023 UniAndes-Google DeepMind Scholarships. This work was partially supported by Azure sponsorship credits granted by Microsoft’s AI for Good Research Lab.

\bibliographystyle{model2-names.bst}\biboptions{authoryear}
\bibliography{refs}

\newpage

\section*{Supplementary Material}











\setcounter{table}{0}
\setcounter{figure}{0}
\appendix

\section{Original Frame Pre-Processing}
\label{sec:appendix::dendogram}

\begin{figure}[ht]
    \centering
    \includegraphics[height=0.45\textheight,width=1\linewidth]{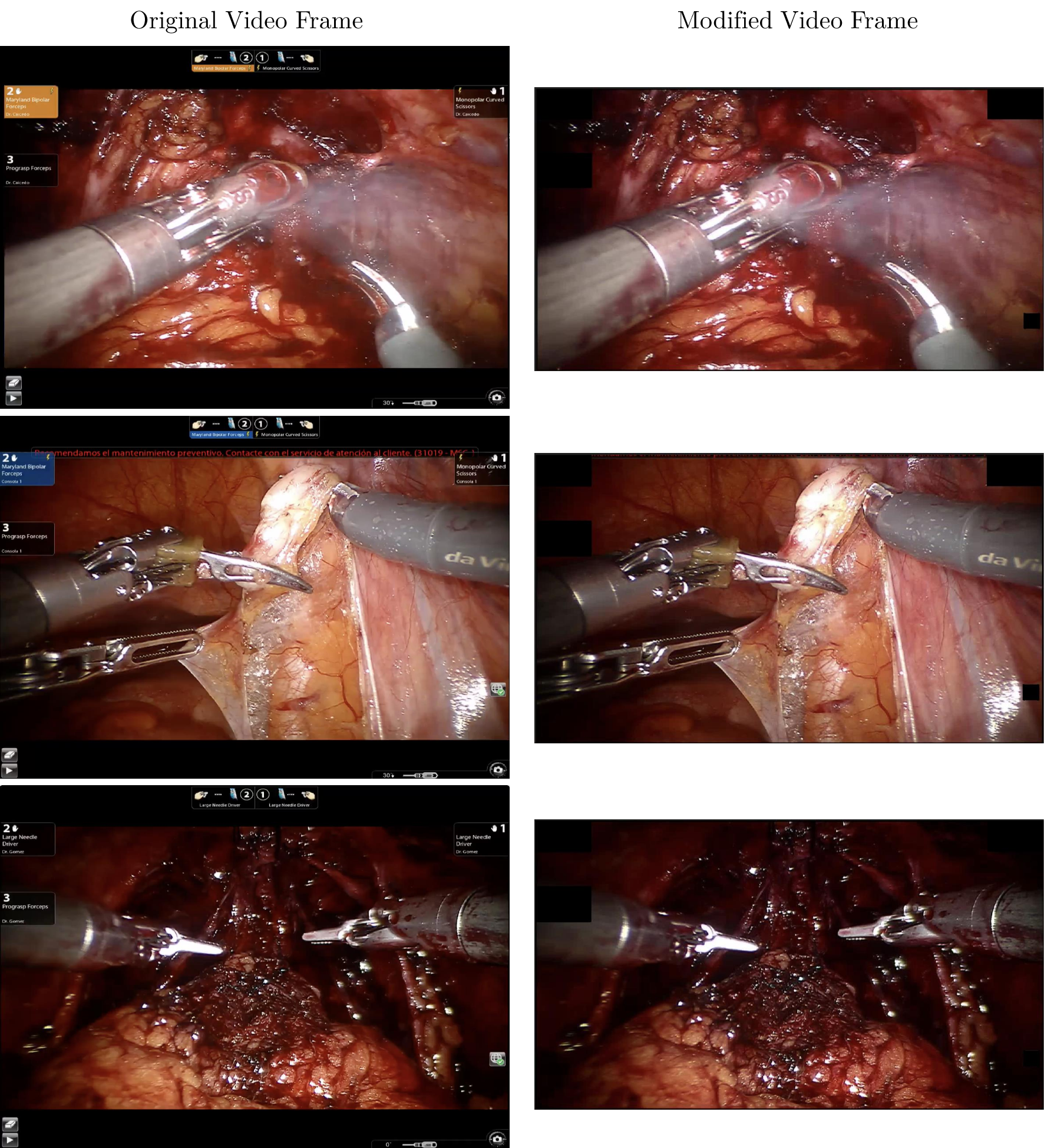}
    \caption{\textbf{Pre-processing examples of GraSP frames.} We removed the top and bottom black borders, along with the instrument name and usage indicators provided by the capturing system.}
    \label{fig:post_processing}
\end{figure}

We modified the original frames captured by the da Vinci Surgical System. Initially, these frames included black borders at the top and bottom, along with indicators displaying the names of the instruments mounted in the robot and whether they were in use. To remove irrelevant visual information, we cropped all the sampled images to eliminate the black borders and the top and bottom indicators, reducing the frames' dimensions from 1024x1280 to 800x1280. Additionally, we applied black masks over the areas containing instrument names and usage indicators to eliminate potential model biases. The outcomes of these modifications are illustrated in Figure~\ref{fig:post_processing}. Our experiments were conducted using the sampled frames that incorporate the abovementioned adjustments. However, we will release both versions of the dataset for possible studies involving the original or the modified frames.

\section{GraSP General Overview}

\begin{figure*}[htbp]
    \centering
    \includegraphics[height=0.9\textheight,width=0.7\linewidth]{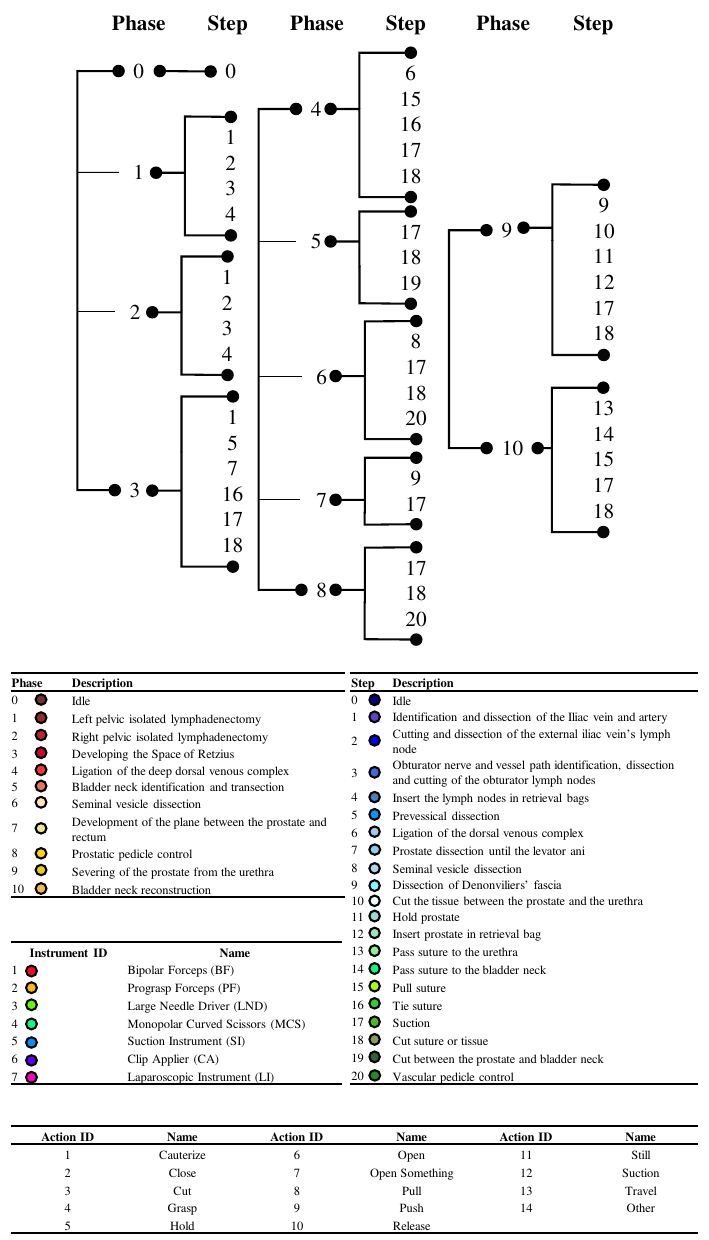}
    \caption{GraSP classes per task. (Top) phases and steps are presented according to their relations along the dataset. (Bottom) list of the class labels for the phase, step, atomic action recognition tasks, and instrument instance segmentation tasks.}
    \label{fig:dendogram}
\end{figure*}

We present Figure~\ref{fig:dendogram} to facilitate the interpretation of phase, step, and action categories throughout their corresponding IDs, used in figures and tables along the main body of this work.
In addition, the dendrogram aims to show the correspondence between phases and steps. These relationships reflect the usual proceedings inside operating rooms, where every surgical procedure has a ``textbook" division into phases, which have themselves a ``textbook" division into steps. They are also the results of the surgical team's efforts to condense their experience into defining a set of processes that span a certain amount of time. Finally, it is essential to highlight the correspondence between the \textit{idle} phase and \textit{idle} step, as it shows that GraSP respects the logical relations between phases and steps. The consistency in the order of appearances of the phases and steps across all videos demonstrates that the GraSP dataset accurately represents surgical scenes. \\

\begin{figure*}[htbp]
    \centering
    \includegraphics[height=0.6\textheight,width=0.9\linewidth]{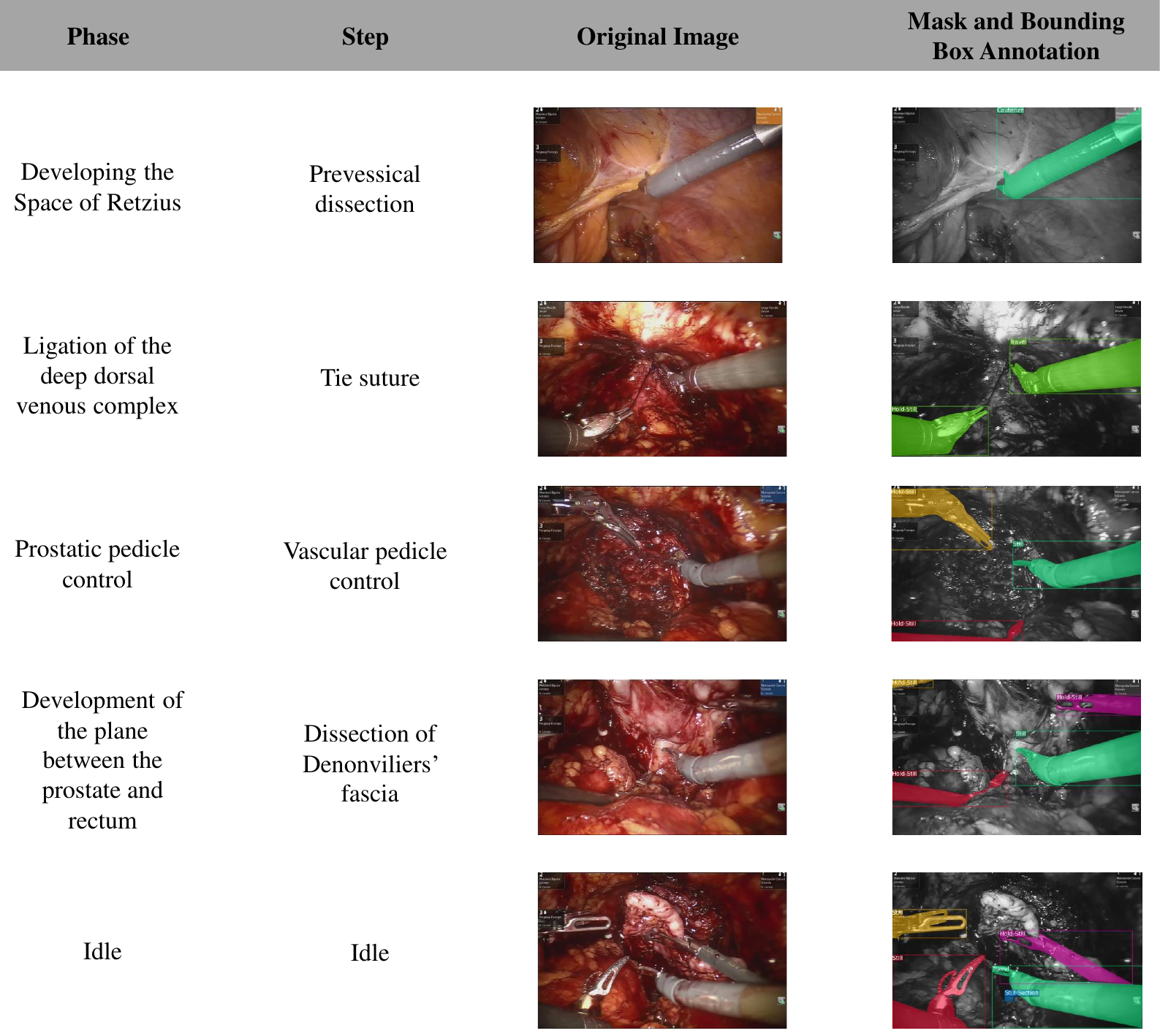}
    \caption{Examples of the GraSP dataset, representing original radical prostatectomy image and their corresponding phase, step, instruments segmentation mask, and atomic action annotations.}
    \label{fig:general_visuals}
\end{figure*}

Furthermore, Figure~\ref{fig:general_visuals} illustrates various instances from the GraSP dataset. These examples highlight a substantial diversity in the steps and phases and the relations between long-term and short-term tasks. For instance, in the ligation of the deep dorsal venous complex, completing that phase requires a specific task, such as the tie suture step. The annotations for instrument mask segmentation encompass two Large Needle Drivers, executing hold-still and travel sets of atomic actions. The successful accomplishment of each annotated phase and step crucially depends on using specific instruments and the corresponding actions they perform. \\

\begin{figure*}[htbp]
    \centering
    \includegraphics[height=0.3\textheight,width=0.9\linewidth]{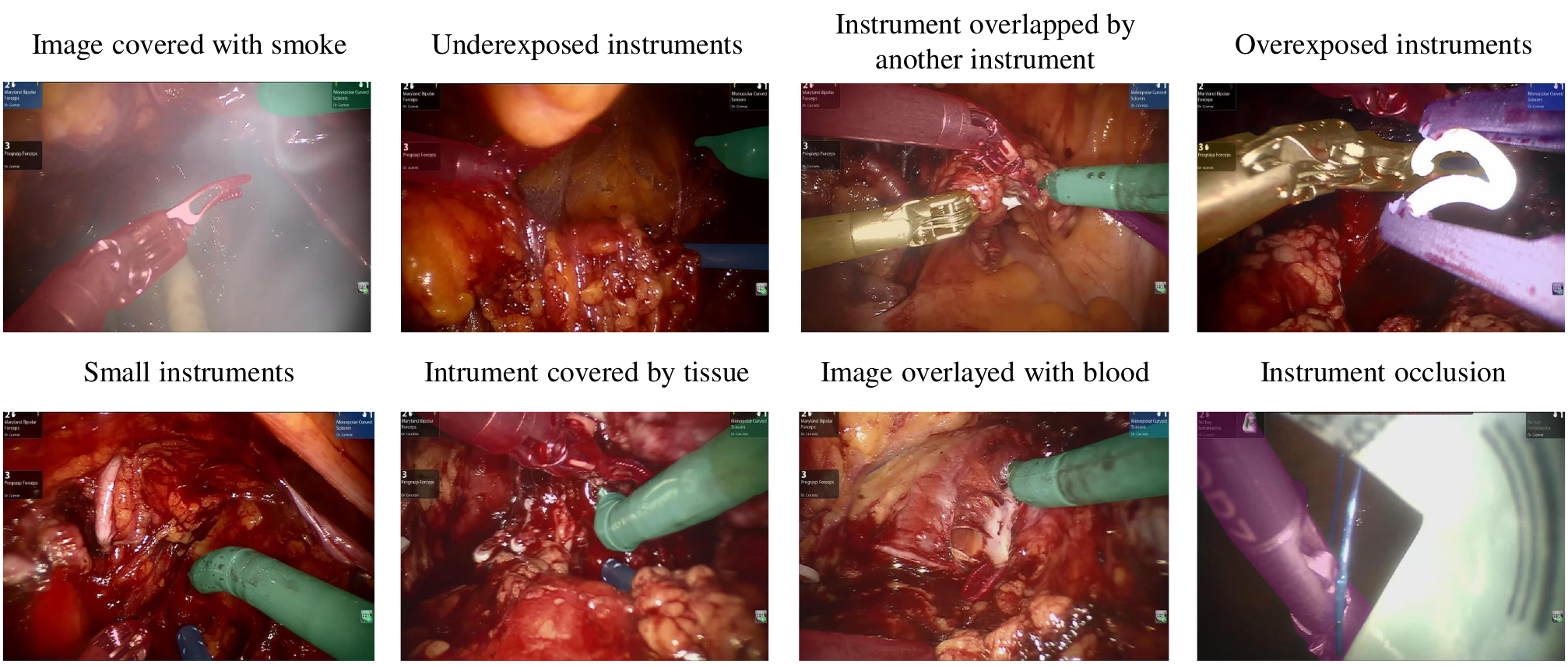}
    \caption{Examples of challenging frames in the GraSP dataset overlaid with multi-instance segmentations performed by experts.}
    \label{fig:hard_examples}
\end{figure*}

Finally, we display complex frame examples in Figure~\ref{fig:hard_examples}. The GraSP dataset contains frames introducing diverse challenges, such as small instances of instruments, overlapping instruments, frames obscured by smoke post-cauterization, instances of instruments in varying exposure conditions, instrument occlusion by non-surgical objects, instruments concealed by tissue, and frames overlaid with blood. These examples reflect challenges encountered in surgical procedures and highlight the need to create methods that effectively overcome these complexities.

\begin{figure*}[htbp]
    \centering
    \includegraphics[width=\textwidth]{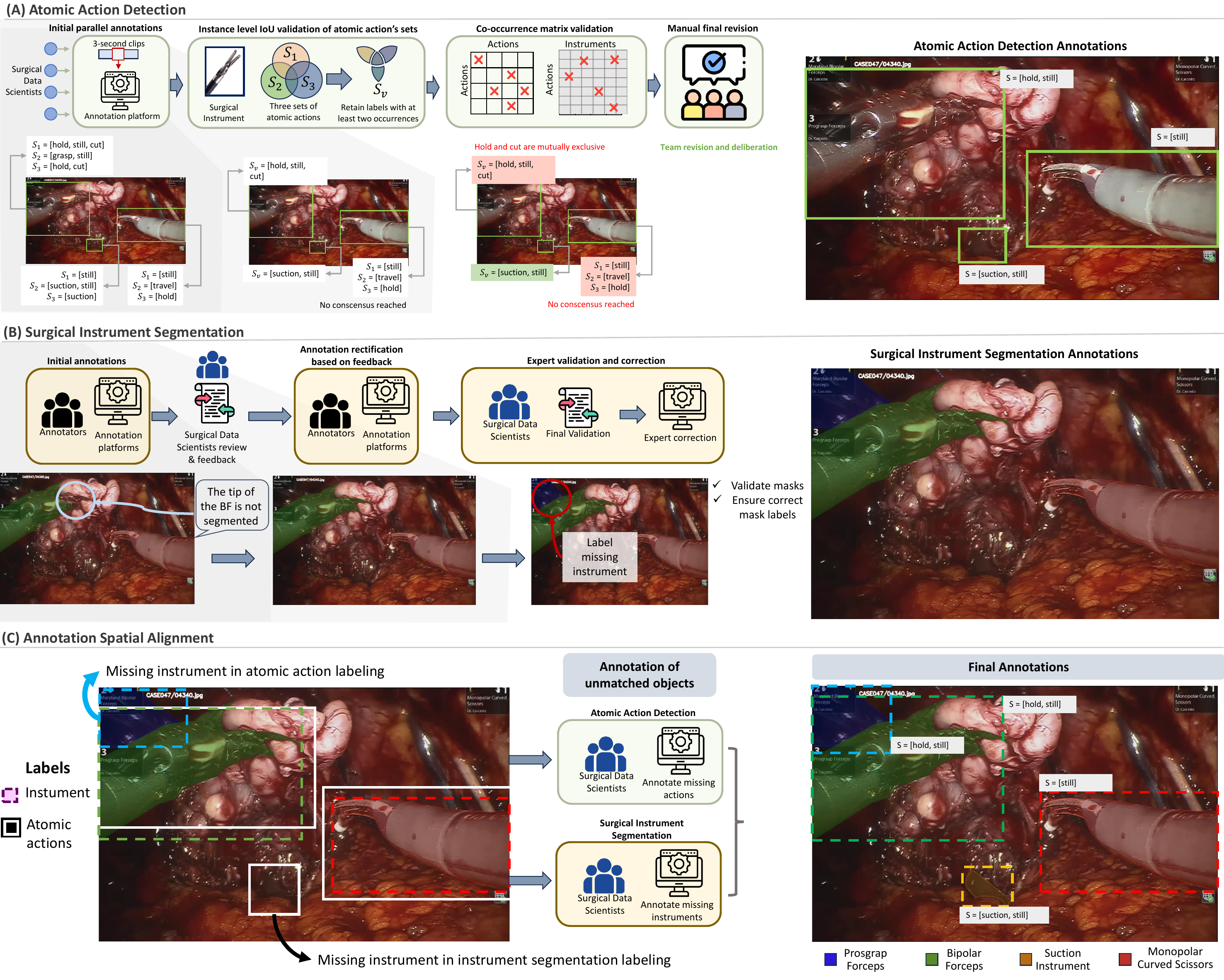}
    \caption{\textbf{Short-Term Tasks Annotation Diagram.} Our annotation process for instrument segmentation and atomic action detection (short-term tasks) was done per frame on keyframes sampled at 35s intervals. First, four surgical data science experts annotated all instrument instances for surgical atomic action recognition on each keyframe independently (top-part of the diagram). In parallel, surgical instruments were segmented by a team of 17 annotators trained and supervised by the surgical data science experts (middle part of the diagram). Third, both annotation types were aligned through bipartite matching, resulting in a single dataset with consistent instance segments and bounding boxes (bottom part of the diagram). We provide further details on the annotation process in Subsection 3.3 of the main manuscript.}
    \label{fig:annotation_diagram}
\end{figure*}


\begin{table*}[]
\caption{\textbf{Number of frames sampled at 1fps for each phase in the GraSP dataset.} Phase categories are Idle (0), Left pelvic isolated lymphadenectomy (1), Right pelvic isolated lymphadenectomy (2), Developing the Space of Retzius (3), Ligation of the deep dorsal venous complex (4), Bladder neck identification and transection (5), Seminal vesicle dissection (6), Development of the plane between the prostate and rectum (7), Prostatic pedicle control (8), Severing of the prostate from the urethra (9), and Bladder neck reconstruction (10).}
\centering
\begin{tabular}{c|ccccccccccc}
\hline
\multirow{2}{*}{\textbf{Set}} & \multicolumn{11}{c}{\textbf{Phase Category ID}}\\ \cline{2-12} 
& 0     & 1    & 2    & 3    & 4    & 5    & 6    & 7    & 8    & 9    & 10   \\ \hline
\textbf{Fold 1}               & 9938  & 4752 & 3719 & 4367 & 1355 & 3349 & 2899 & 680  & 1752 & 1198 & 4936 \\
\textbf{Fold 2}               & 8192  & 5815 & 3704 & 4506 & 1234 & 2274 & 2159 & 655  & 2529 & 1431 & 2715 \\
\textbf{Test}                 & 12934 & 4838 & 3433 & 5317 & 1478 & 2581 & 3608 & 1041 & 3657 & 902  & 3108 \\ \hline
\end{tabular}
\label{tab:frequencies_phases}
\end{table*}

\begin{table*}[]
\centering
\caption{\textbf{Number of frames sampled at 1fps for each step in the GraSP dataset.} Step categories are Idle (0), Identification and dissection of the Iliac vein and artery (1), Cutting and dissection of the external iliac vein’s lymph node (2), Obturator nerve and vessel path identification, dissection, and cutting of the obturator lymph nodes (3), Insert the lymph nodes in retrieval bags (4), Prevessical dissection (5), Ligation of the dorsal venous complex (6), Prostate dissection until the levator ani (7), Seminal vesicle dissection (8), Dissection of Denonviliers’ fascia (9), Cut the tissue between the prostate and the urethra (10), Hold prostate (11), Insert prostate in retrieval bag (12), Pass suture to the urethra (13), Pass suture to the bladder neck (14), Pull suture (15), Tie suture (16), Suction (17), Cut suture or tissue (18), Cut between the prostate and bladder neck (19), and Vascular pedicle control (20).}
\resizebox{\textwidth}{!}{
\begin{tabular}{c|ccccccccccllllllllllc}
\hline
\multirow{2}{*}{\textbf{Set}} & \multicolumn{21}{c}{\textbf{Step Category ID}}  \\ \cline{2-22} 
& 0     & 1    & 2    & 3    & 4    & 5    & 6   & 7    & 8    & 9    & 10  & 11  & 12  & 13  & 14   & 15   & 16  & 17   & 18   & 19   & 20   \\ \hline
\textbf{Fold 1}               & 9917  & 3130 & 2605 & 2060 & 676  & 1612 & 377 & 135  & 2739 & 849  & 698 & 118 & 80  & 718 & 1272 & 992  & 712 & 938  & 5120 & 2895 & 765  \\
\textbf{Fold 2}               & 8192  & 4628 & 2109 & 1578 & 1204 & 1009 & 345 & 1233 & 1893 & 908  & 919 & 35  & 128 & 651 & 1163 & 801  & 559 & 1083 & 4292 & 1916 & 568  \\
\textbf{Test}                 & 12934 & 5340 & 3646 & 1366 & 545  & 630  & 481 & 783  & 3223 & 1112 & 456 & 84  & 62  & 818 & 1228 & 1305 & 491 & 1009 & 3506 & 2486 & 1392 \\ \hline
\end{tabular}
}
\label{tab:frequencies_steps}
\end{table*}



\section{Additional GraSP Statistics}
\label{sec:appendix:stats}

\subsection{Frequency Distributions}


We extend the statistical analysis of frequency distributions across all tasks within GraSP. Tables~\ref{tab:frequencies_instruments},~\ref{tab:frequencies_actions},~\ref{tab:frequencies_phases}, and~\ref{tab:frequencies_steps} demonstrate that we maintained the consistency in annotation and distribution throughout the two-fold cross-validation setup and in the test set. These tables show the total number of instances per task in each fold, enabling the identification of over-represented and under-represented instances in each task. This analysis provides insights into their correlation with real-life surgical scenarios. 

\begin{table}[]
\caption{\textbf{Number of segment instances for each instrument category in the GraSP dataset.} Instrument categories are Bipolar Forceps (BF), Prograsp Forceps (PF), Large Needle Driver (LND), Monopolar Curved Scissors (MCS), Suction Instrument (SI), Clip Applier (CA), and Laparoscopic Graspers/Instruments (LG).}
\centering
\begin{tabular}{c|ccccccc}
\hline
\multirow{2}{*}{\textbf{Set}} & \multicolumn{7}{c}{\textbf{Instrument Category}} \\ \cline{2-8} 
                              & BF   & PF  & LND & MCS & SI  & CA & LG  \\ \hline
\textbf{Fold 1}               & 876  & 386 & 524 & 915 & 405 & 28 & 101 \\
\textbf{Fold 2}               & 818  & 355 & 372 & 850 & 411 & 36 & 93  \\
\textbf{Test}                 & 809  & 330 & 449 & 844 & 310 & 38 & 81  \\ \hline
\end{tabular}
\label{tab:frequencies_instruments}
\end{table}

\begin{figure}[t]
    \centering
    \includegraphics[width=\linewidth]{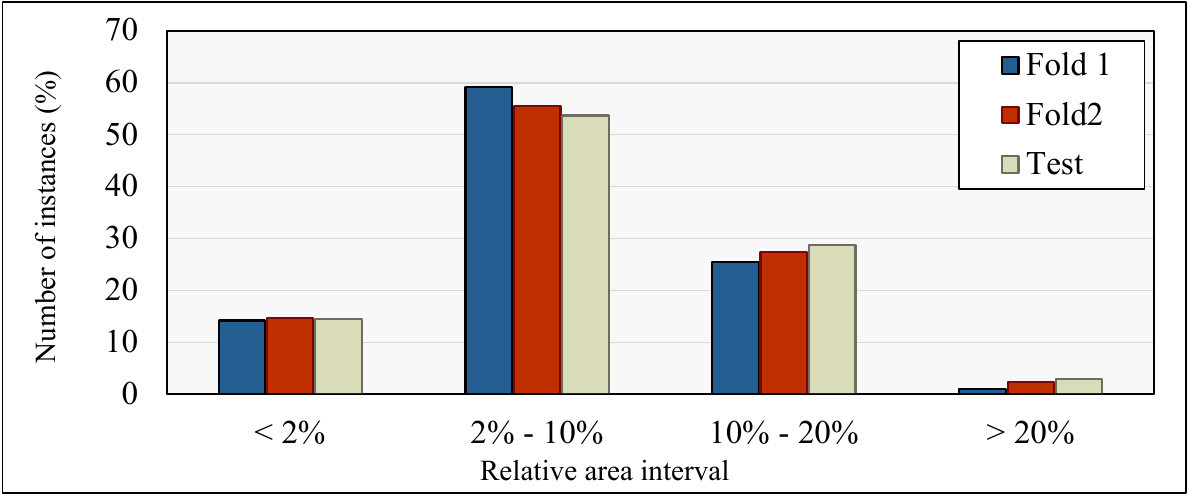}
    \caption{\textbf{Distribution of the percentage of annotated segments per segment relative area interval on each data split.} The relative area of an annotated segment corresponds to the fraction of the image it occupies. The figure is best viewed in color.}
    \label{fig:area_inst}
\end{figure}

Figure~\ref{fig:area_inst} displays the dataset's relative area distribution of instrument instances. Our instrument instances' relative sizes range from 0.02\% to 40.4\%, with an average relative size of 7.5\% and most instruments occupying between 2 and 10\% of the video frames. This variation in relative size stems from factors such as the proximity of the instruments to the endoscope or the portion of the instrument that is visible. In this regard, a prominent challenge of our instrument segmentation benchmark is the significant number of small segments, with more than 10\% of instrument instances corresponding to segments with less than a 2\% relative area. Smaller instances typically result from instruments being distant from the camera, partial occlusion, or most of the instrument body being out of the camera's field of view. One notable improvement in our GraSP dataset over the PSI-AVA benchmark is the inclusion of more instrument instances with minimal areas. This enhancement is a direct result of our meticulous data curation process.


Table~\ref{tab:instances_instruments_frame} presents the number of instrument instances per frame distribution. Notably, we observe that 80\% of frames contain 2 or 3 instruments, and 95\% of frames contain between 2 and 4 instrument instances. These values reflect the standard surgical practice of using multiple instruments to perform complex operations. Nonetheless, using too many instruments can clutter the surgical scene and limit working space. Consequently, across all surgeries analyzed, we find that less than 1\% of frames show 5 instruments. Additionally, the frequency of the instrument can be found in Table~\ref{tab:frequencies_instruments}, displaying the importance of the \textit{Bipolar Forceps} and the \textit{Monopolar Curved Scissors} in the performance of the Radical Prostatectomy surgery. These instruments can execute many actions (from holding, pulling, and pushing to cutting tissues) required throughout the surgical procedure. 


Table~\ref{tab:actions_coocurrences} presents the frequency distribution of action combinations in the GraSP dataset. We observe that the most common actions are \textit{(hold, still)}, \textit{(still)} and \textit{(travel)}. This situation is expected as \textit{still} and \textit{travel} are actions fundamental to describing the activities of all instruments and are constantly and widely performed in any surgery instance. The other action shown here is \textit{hold}. This situation coincides with the action distribution shown in Table~\ref{tab:frequencies_actions} where \textit{hold} is the second most common action. This action cannot be performed alone and, thus, is always combined with another action to describe the activity of an instrument. These action combinations reflect the manipulation of tissues and objects within the surgical scene. We also observe that some sets of actions have a low frequency. These actions correspond to uncommon moves that are nonetheless valid activities that are part of usual procedures and can be seen in other surgeries. 

Tables~\ref{tab:frequencies_phases} and~\ref{tab:frequencies_steps} represent the number of annotations for both tasks. The most frequent phase and step corresponds to idle, and this is because it represents the beginning and end points of the surgery and also transition times between other phases and steps. 

\begin{table}[t]
\caption{\textbf{Number of annotated frames containing each of the possible amounts of instrument instances present.}}
\centering
\begin{tabular}{c|ccccc}
\hline
\multirow{2}{*}{\textbf{Set}} & \multicolumn{5}{c}{\textbf{Number of Instrument Instances}} \\ \cline{2-6} 
& 1        & 2        & 3        & 4        & 5      \\ \hline
\textbf{Fold 1}               & 51       & 511      & 492      & 165      & 5      \\
\textbf{Fold 2}               & 65       & 412      & 446      & 172      & 4      \\
\textbf{Test}                 & 67       & 500      & 442      & 112      & 4      \\ \hline
\end{tabular}
\label{tab:instances_instruments_frame}
\end{table}

\begin{figure}[]
    \centering
    \includegraphics[width=\linewidth]{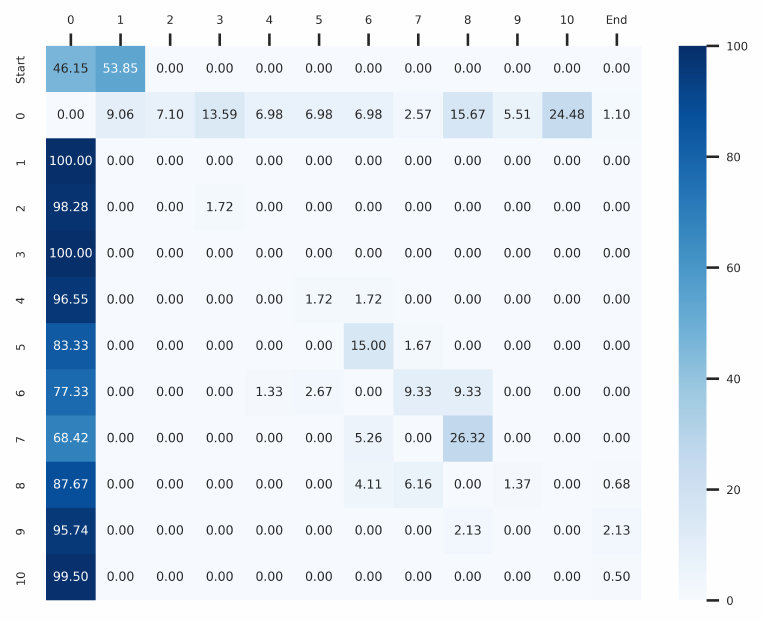}
    \caption{\textbf{Transition probabilities of phases (\%) across the 13 cases.} Each value represents the probability of going from the previous step (rows) to the next step (columns) (e.g., in 50.00\% of cases, phase 7 transitions to phase 0, but in 40.91\% of cases, it transitions to phase 8).}
    \label{fig:mat_trans_phases}
\end{figure}

\begin{figure}[]
    \centering
    \includegraphics[width=\linewidth]{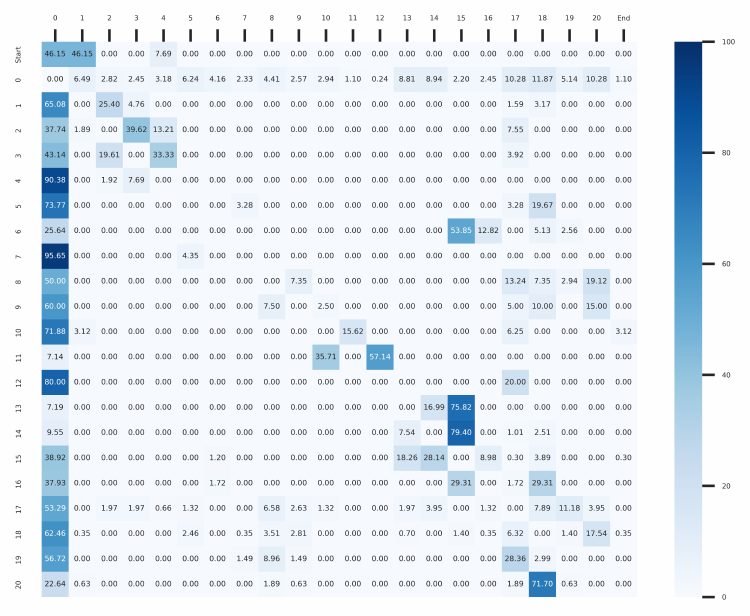}
    \caption{\textbf{Transition probabilities of steps (\%) across the 13 cases.} Each value represents the probability of going from the previous step (rows) to the next step (columns) (e.g., in 15.23\% of cases, step 14 transitions to step 0, but in 71.52\% of cases, it transitions to step 15).}
    \label{fig:mat_trans_steps}
\end{figure}

\begin{figure*}[t]
    \centering
    \includegraphics[width=\textwidth]{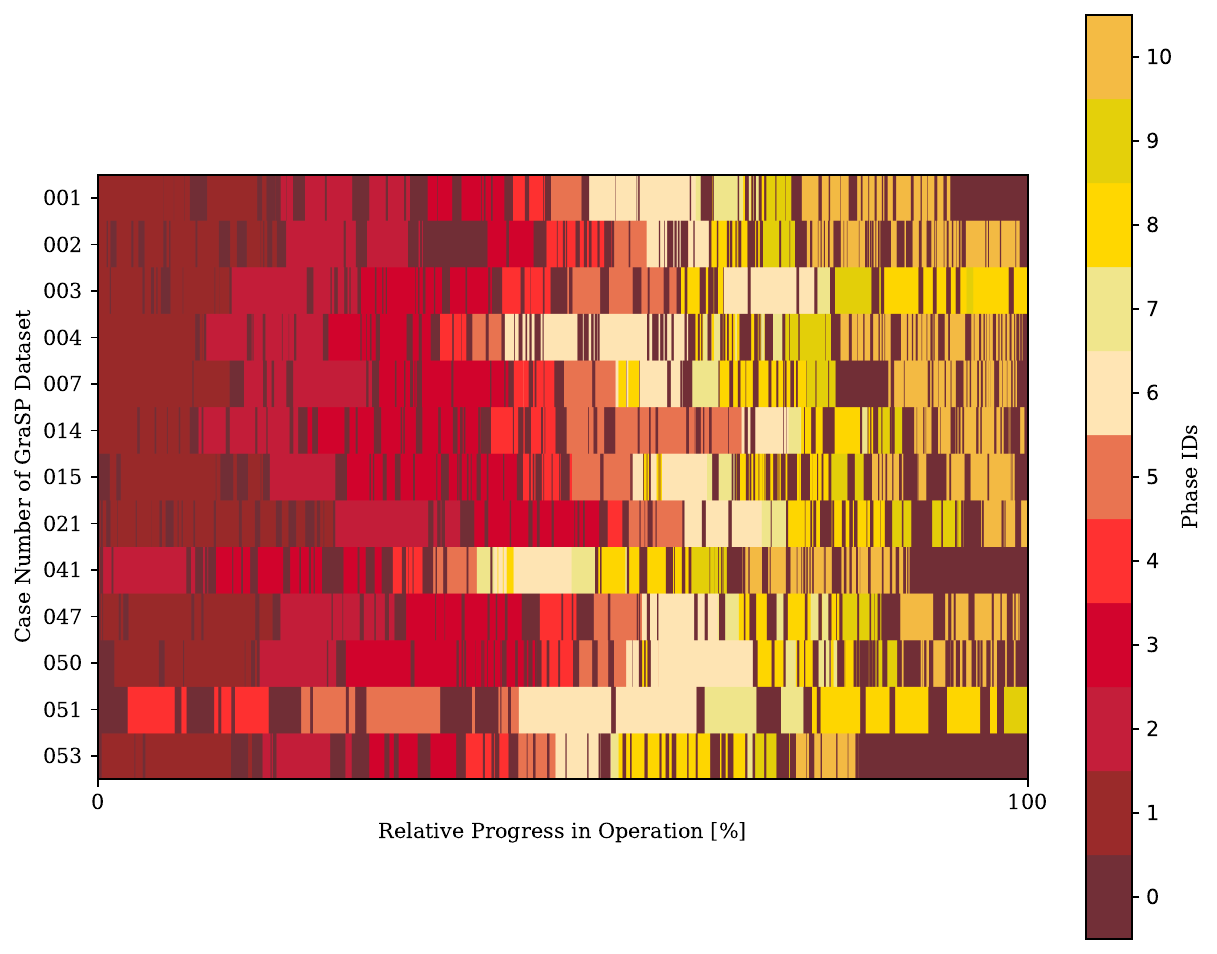}
    \caption{\textbf{Visual transition of phases in the GraSP dataset.} Each row in the y-axis represents one of the 13 surgeries in GraSP and the x-axis represents the relative temporal progression of the surgery from 0\% (start) to 100\% (end).}
    \label{fig:overall_phase}
\end{figure*}

\begin{figure}[]
    \centering
    \includegraphics[width=\linewidth]{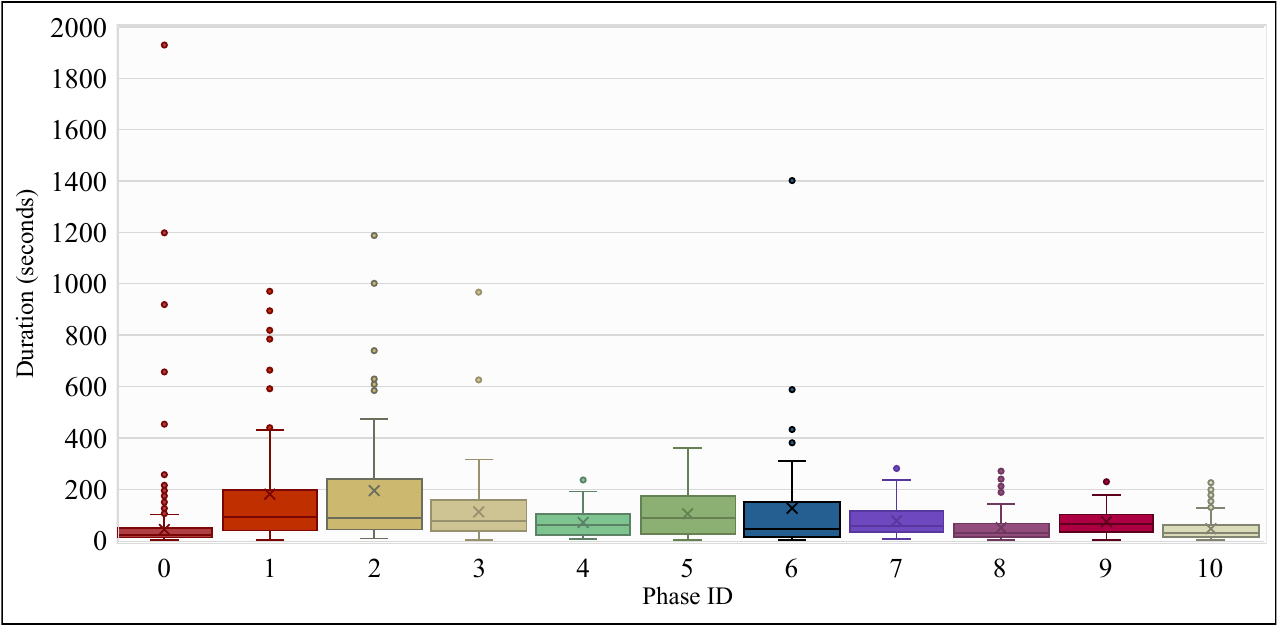}
    \caption{\textbf{Boxplot distribution of the duration of each phase category.} Each boxplot presents the distribution of the time span in seconds of all the present temporal windows corresponding to each phase category in our dataset. The figure contains the entire duration of each of the classes. The figure is best viewed in color.}
    \label{fig:boxplot_complete_phases}
\end{figure}

\begin{figure*}[t]
    \centering
    \includegraphics[width=\textwidth]{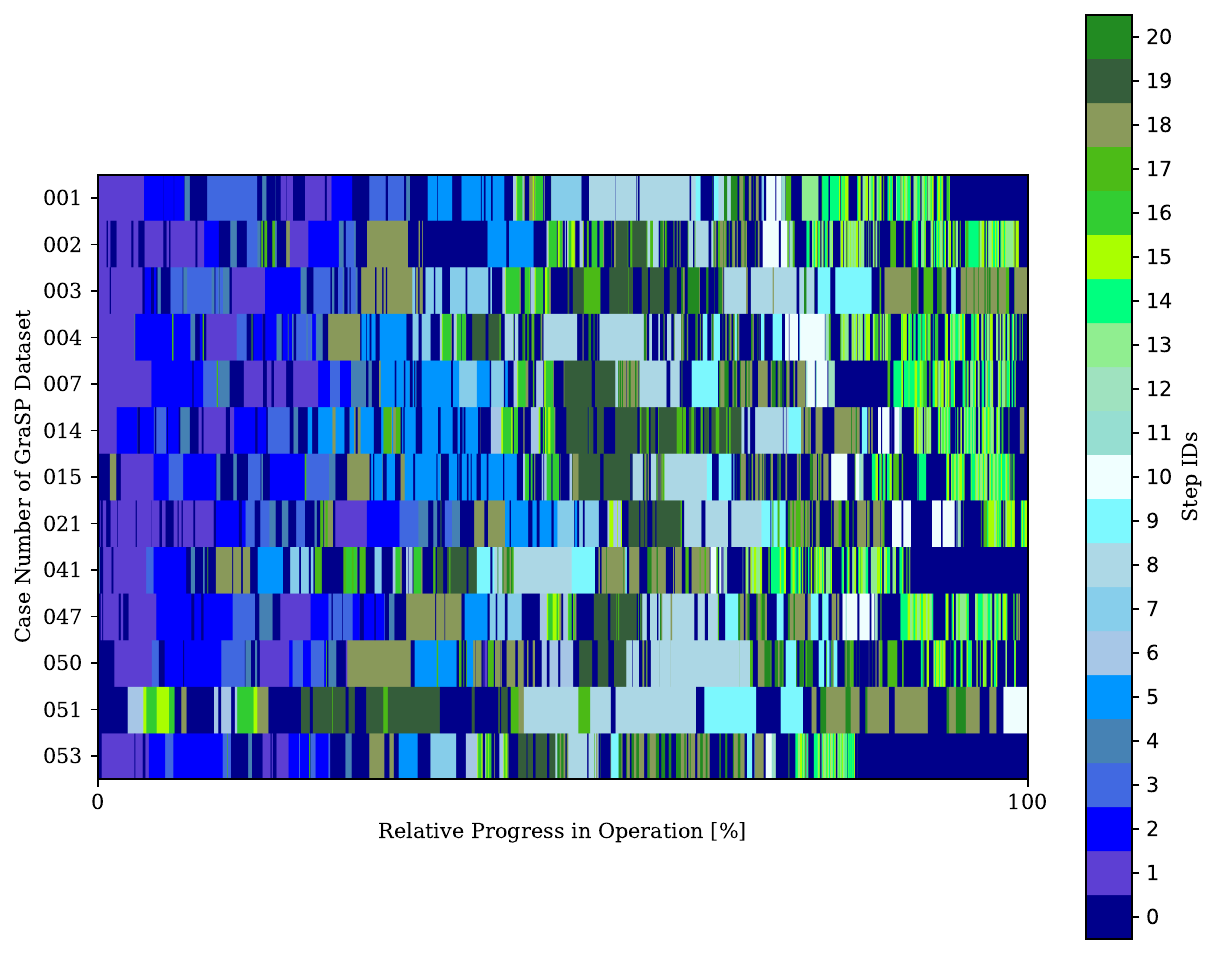}
    \caption{\textbf{Visual transition of steps in the GraSP dataset.} Each row in the y-axis represents one of the 13 surgeries in GraSP and the x-axis represents the relative temporal progression of the surgery from 0\% (start) to 100\% (end).}
    \label{fig:overall_step}
\end{figure*}

\begin{figure}[]
    \centering
    \includegraphics[width=\linewidth]{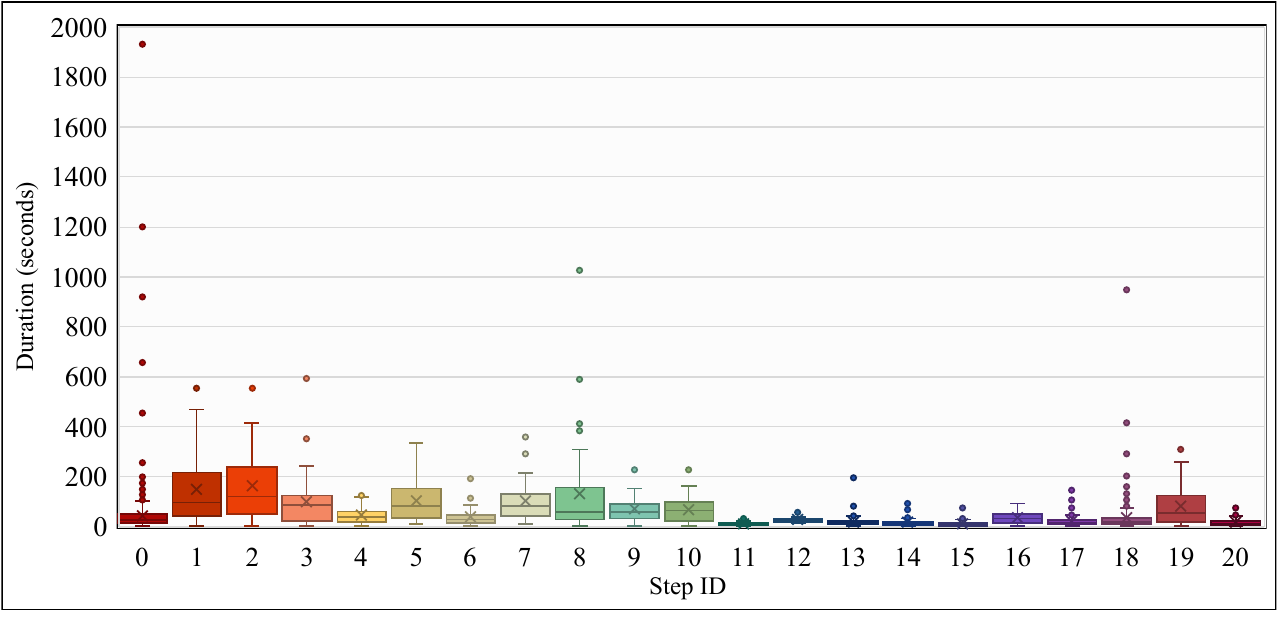}
    \caption{\textbf{Boxplot distribution of the duration of each step category.} Each boxplot presents the distribution of the time span in seconds of all the present temporal windows corresponding to each step category in our dataset. The figure contains the entire duration of each of the classes. The figure is best viewed in color.}
    \label{fig:boxplot_complete_steps}
\end{figure}

\begin{table*}[t]
\centering
\caption{\textbf{Number of instances for each class of the atomic action recognition task.} Action categories are Cauterize (Cau), Close (Clo), Cut (Cut), Grasp (Gra), Hold (Hol), Open (Open), Open Something (O.Sm), Pull (Pull), Push (Pus), Release (Rel), Still (Sti), Suction (Suc), Travel (Tra) and Other (Oth).}
\begin{tabular}{c|cccccccccccccc}
\hline
\multirow{2}{*}{\textbf{Set}} & \multicolumn{14}{c}{\textbf{Action Category}} \\ \cline{2-15} 
& Cau & Clo & Cut & Gra & Hol  & Ope & O.Sm & Pul & Pus & Rel & Sti  & Suc & Tra & Oth \\ \hline
\textbf{Fold 1}               & 93  & 94  & 50  & 43  & 1175 & 48  & 3    & 132 & 242 & 25  & 1883 & 211 & 838 & 5   \\
\textbf{Fold 2}               & 77  & 75  & 30  & 32  & 1013 & 41  & 15   & 99  & 222 & 25  & 1691 & 188 & 813 & 7   \\
\textbf{Test}                 & 59  & 71  & 42  & 56  & 923  & 51  & 25   & 108 & 182 & 15  & 1738 & 186 & 737 & 10  \\ \hline
\end{tabular}
\label{tab:frequencies_actions}
\end{table*}

\begin{table*}[t]
\caption{\textbf{Atomic actions combinations frequencies in the GraSP dataset.} We present the number of instances of each possible action combination in GraSP.}
\centering
\resizebox{18cm}{!}{
\begin{tabular}{llllll}
\hline
\textbf{Name} & \textbf{Count} & \textbf{Name} & \textbf{Count} & \textbf{Name} & \textbf{Count} \\ \hline
(Hold, Still,) & 2532 & (Pull,) & 24 & (Close, Cut, Push) & 2 \\

(Still,) & 2311 & (Cauterize, Cut,) & 24 & (Still, Other,) & 2  \\

(Travel,) & 1882 & (Release, Travel,) & 22 & (Hold, Push, Other) & 1 \\

(Push,) & 465 & (Open, Release,) & 22 & (Close, Grasp, Pull) & 1 \\

(Still, Suction,) & 425 & (Open, Release, Travel) & 19 & (Hold, Pull, Other) & 1 \\

(Hold, Pull,) & 266 & (Cauterize, Travel,) & 17 & (Cauterize, Open, Travel) & 1 \\

(Hold, Travel,) & 203 & (Cauterize, Push,) & 14 & (Cauterize, Close, Still) & 1 \\

(Suction, Travel,) & 120 & (Open Something, Pull,) & 13 & (Close, Push) & 1 \\

(Close, Grasp,) & 92 & (Close, Other) & 10 & (Hold, Open Something, Pull) & 1 \\

(Cauterize,) & 89 & (Close, Still,) & 10 & (Push, Other,) & 1 \\

(Hold, Push,) & 70 & (Cauterize, Cut, Push) & 9 & (Cauterize, Open Something, Push) & 1 \\

(Open, Travel,) & 65 & (Cauterize, Hold, Still) & 8 & (Open, Push, Release) & 1 \\

(Close, Travel,) & 43 & (Cauterize, Close,) & 7 & (Open, Open Something,) & 1 \\

(Close, Cut,) & 42 & (Cut, Travel,) & 8 & (Cut, Pull,) & 1 \\

(Push, Suction,) & 40 & (Close, Grasp, Push) & 6 & (Cauterize, Open, Release) & 1 \\

(Grasp, Pull,) & 32 & (Cauterize, Cut, Travel) & 6 & (Cut, Open Something, Push) & 1 \\

(Cauterize, Cut, Close) & 25 & (Open, Push,) & 5 & (Cauterize, Open,) & 1 \\

(Open Something, Push,) & 25 & (Cut, Push,) & 4 & (Open Something, Hold) & 1 \\

(Cauterize, Hold,) & 25 & (Hold, Other,) & 4 &  \\

(Open, Still,) & 24 & (Travel, Other,) & 3 & \textbf{Total} & 9031 \\
 \hline
\end{tabular}}
\label{tab:actions_coocurrences}
\end{table*}


\subsection{Surgical Transitions}

Figures~\ref{fig:overall_phase} and~\ref{fig:overall_step} show the transitions between phases and steps. These figures show how phase 0 and step 0 are recurrent throughout all surgeries and represent the transition between one step and another and one phase and another. Lastly, both tables and figures exhibit similarity in the order and duration of each phase and step during the temporal progression of the surgeries, hence showing the intrinsic procedural and ordered nature of surgical procedures and proving the consistency in annotation processes for both steps and phase tasks.

The transition matrices in Figures~\ref{fig:mat_trans_phases} and~\ref{fig:mat_trans_steps} present the transition probabilities of phases and steps across the GraSP dataset. Each value represents the probability of going from the previous step/phase (rows) to the next step/phase (columns). They demonstrate the logical connections between the phases and the steps. For instance, Figure~\ref{fig:mat_trans_phases} shows that \textit{Developing the space of Retzius} always follows the \textit{Right pelvic isolated lymphadenectomy}, as defined by the surgical team. There can be a pause between the two phases, corresponding to an \textit{Idle} phase. The Figures show that most phases and steps transition from or to an Idle and that steps present more variability in their transition probabilities as step categories can show repeatedly throughout the procedure. Additionally, Figures~\ref{fig:boxplot_complete_phases} and \ref{fig:boxplot_complete_steps} show the complete boxplots of the distribution of the duration of each phase and step categories. Besides the previous analysis in the Dataset Statistics Section, we observe that phase categories present more outliers, tending to have more extensive durations, once again proving the temporal granularity difference between phases and steps, where phases tend to represent broader temporal segments.

\newpage


\begin{figure}[ht]
\centering
\includegraphics[width=\linewidth]{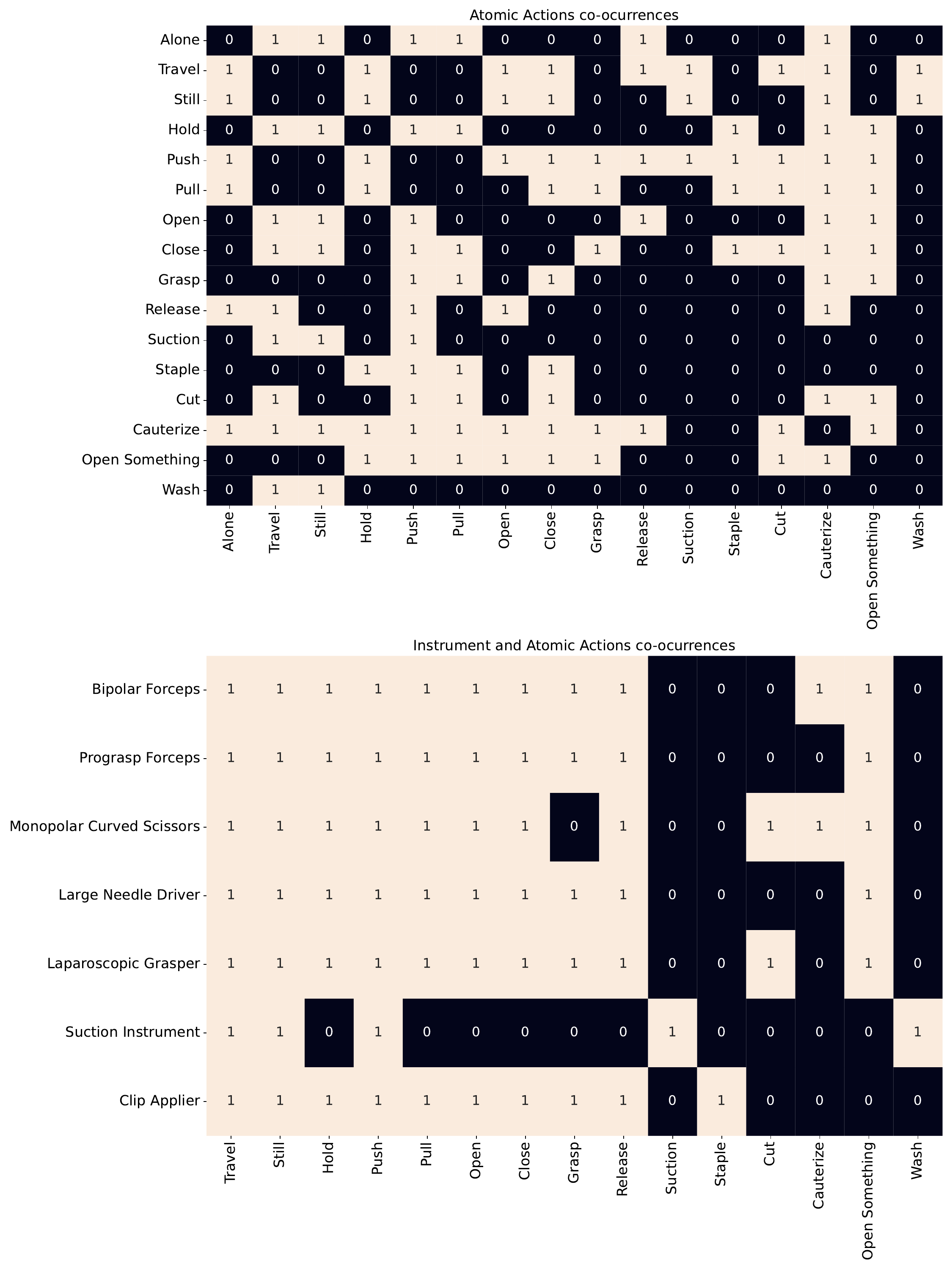}
\caption{\textbf{Expert-guided theoretical co-occurrences matrices made for annotation's validation.} We present the possible co-occurrences between individual atomic actions and between instrument and individual atomic actions in GraSP. Matrix positions with a 1 denote possible co-occurrences of two actions within the same instrument instance or an action with a specific instrument, while a 0 denotes impossible co-occurrence. The \textit{Alone} rows and columns represent whether an action can be performed solely without other simultaneous actions. In our final dataset, we set the Staple and Wash categories as "Other" categories due to the low frequencies of those actions.}
\label{fig:expert_coocurrences}
\end{figure}

\subsection{Co-Occurence Matrices}

Figure~\ref{fig:expert_coocurrences} portrays the theoretical matrixes established by the experienced surgical team. These matrices demonstrate possible combinations of actions that can occur simultaneously in the same instrument instance, and also the possible actions that each instrument can perform. We design the matrices to have a predefined \textit{a priori} rule of possible annotations based on the extensive knowledge of our surgeons. We applied this matrix after the first annotation rounds to validate and correct our annotations.

\begin{figure}[t]
\centering
\includegraphics[width=\linewidth]{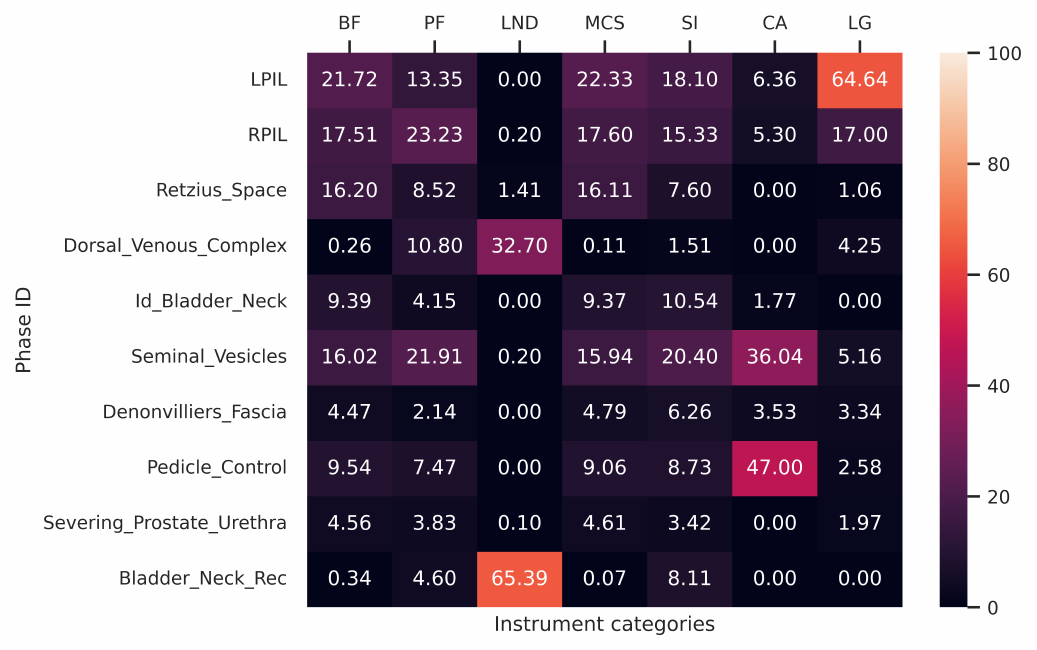}
\caption{\textbf{Co-occurrence distribution matrix of instruments in function of the surgical phase in GraSP.}}
\label{fig:inst_phase}
\end{figure}

\begin{figure}[t]
\centering
\includegraphics[width=\linewidth]{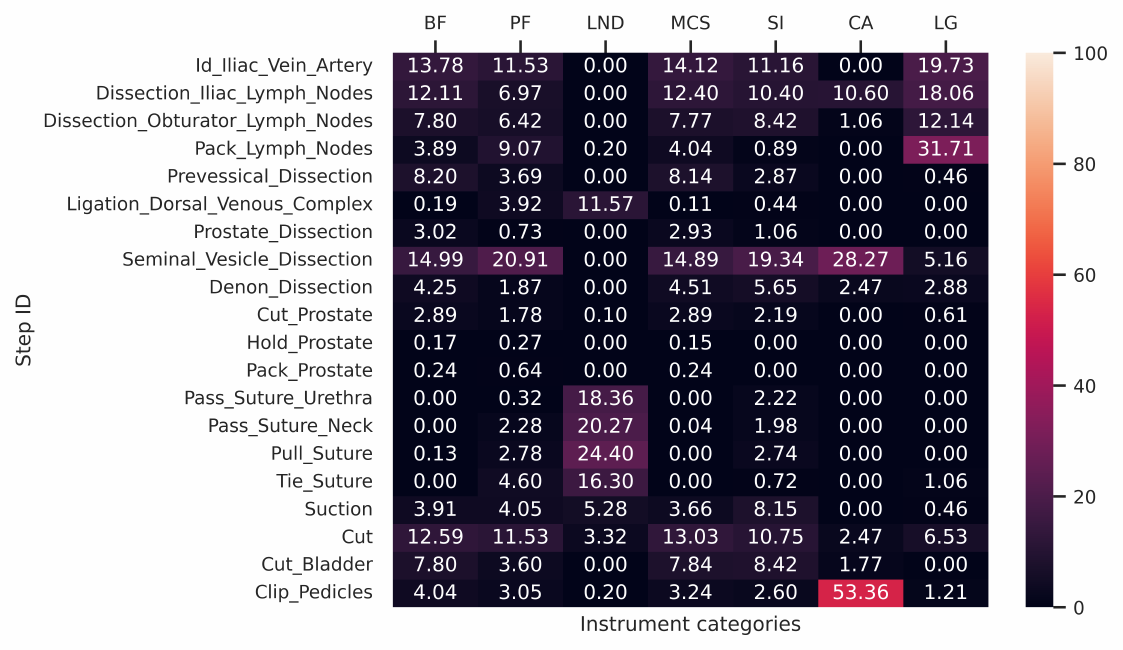}
\caption{\textbf{Co-occurrence distribution matrix of instruments in function of the surgical step in GraSP.}}
\label{fig:inst_step}
\end{figure}

\begin{figure}[t]
\centering
\includegraphics[width = \linewidth]{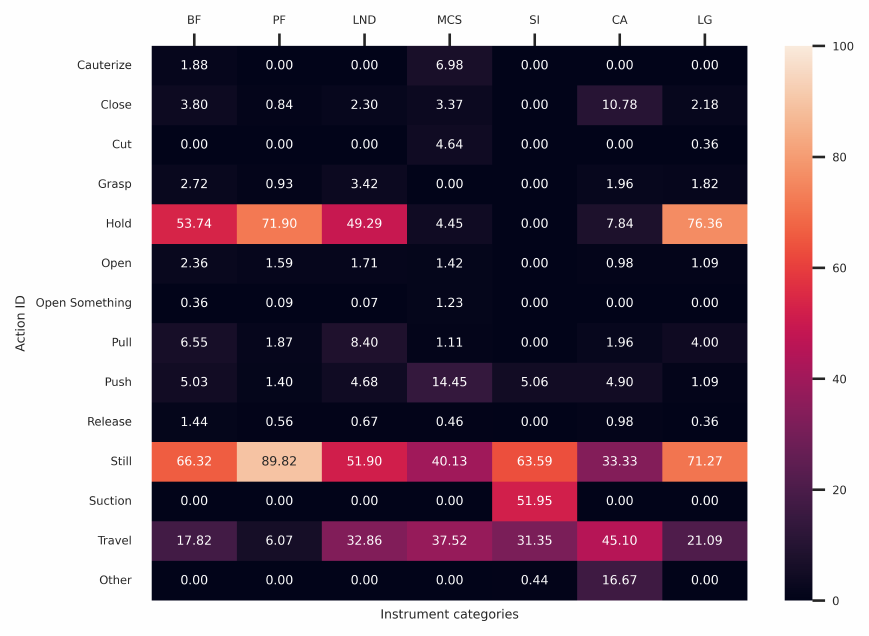}
\caption{\textbf{Co-occurrence distribution matrix of instruments in the function of atomic actions in GraSP.}}
\label{fig:inst_action}
\end{figure}

Figures~\ref{fig:inst_phase} and \ref{fig:inst_step} display the inter-task co-occurrence matrices for instruments-phases and instruments-steps. These matrices demonstrate the varying frequency of distinct instruments during different stages of the surgery. Intuitively, the Large Needle Driver (LND) has the highest co-occurrence with the reconstructive phases of the surgeries like the \textit{Ligation of the dorsal venous complex} and \textit{Bladder neck reconstruction} phases, which involve the use of the LND to maneuver the surgical needles and threats to suture, ligate and reconstruct tissue. Hence, the LND is also the most frequent in suturing steps like \textit{Pass suture to urethra or bladder neck} and \textit{pull or tie suture}. Similarly, the Bipolar Forceps (BF) and Monopolar Curved Scissors are present in most phases and steps but especially towards phases like \textit{Left or Right pelvic isolated lymphadenectomy}, \textit{Bladder neck identification and transection}, and in steps like \textit{Identification and dissection of the Iliac vein and artery} and \textit{Seminal vesicle dissection}, which are mostly related to tissue manipulation and cutting. Moreover, the Clip Applier (CA) has a very high co-occurrence with the \textit{Prostatic pedicle control} phase and the \textit{Vascular pedicle control} step, which require tissue clipping for bleeding control. Also, the Laparoscopic Graspers/Instruments (LG or LI) show mostly in phases and steps requiring the holding of tissue or retrieval bags. Thus, these matrices again demonstrate the strong correlations between different semantic levels of surgical scene understanding.

The matrix in Figure~\ref{fig:inst_action} presents the co-occurrences between the instruments and atomic actions in the GraSP dataset. In other words, this plot shows the typical functions of every instrument in the RARP instances of GraSP. The columns of the plot do not add to 100\%, as instruments can perform up to 3 actions simultaneously. The plot reveals that the Bipolar Forceps usually holds tissues or objects, either in a \textit{still} position or in movement between two locations (\textit{travel}). The Prograsp Forceps differs from it because of its inability to cauterize. Additionally, it is usually used to hold still bigger tissues in the background passively, which explains why it has the highest realization of \textit{still} and the second highest of \textit{hold}. The Large Needle Driver usually stitches tissues with a needle and thread. As such, it is normal for it to be either still or traveling and holding an object. Because the whole purpose of this instrument is to perform the stitching, it travels more than other instruments with pincers that are sometimes used to hold tissues (such as the Bipolar Forceps, the Prograsp Forceps, or the Laparoscopic Graspers/Instruments). The Monopolar Curved Scissors is the only instrument to be equipped with a tip capable of cutting (apart from a particular type of LI that is observed in one case in the fold 1) and thus has the highest percentage of realization of the action \textit{cut}. It is also one of the two instruments capable of performing \textit{cauterize} (the other being the Bipolar Forceps). This action serves to control and prevent bleeding. The Suction Instrument is the only instrument able to perform \textit{suction} and does it in more than half of the frames in which it is present. The actions it performs in the \textit{other} action are instances of \textit{wash}. The Clip Applier is often performing \textit{close} and \textit{release}. These actions can be seen when it takes hold or releases the tissue it has to staple. This instrument is not often seen in the dataset and appears for a short amount of time. It usually travels to the tissue, staples it, and leaves the frame. While it is there, no other procedure is performed. This specificity in its use explains why it is the instrument with the highest percentage of \textit{travel}. Additionally, the action it performs in the \textit{other} action are instances of \textit{staple}. Finally, the Laparoscopic Instrument has a similar role to the Prograsp Forceps, presenting a similar distribution. Nonetheless, it is slightly more active and has a higher frequency of the \textit{travel} action.


\begin{figure*}[ht!]
\centering
\includegraphics[width = \textwidth]{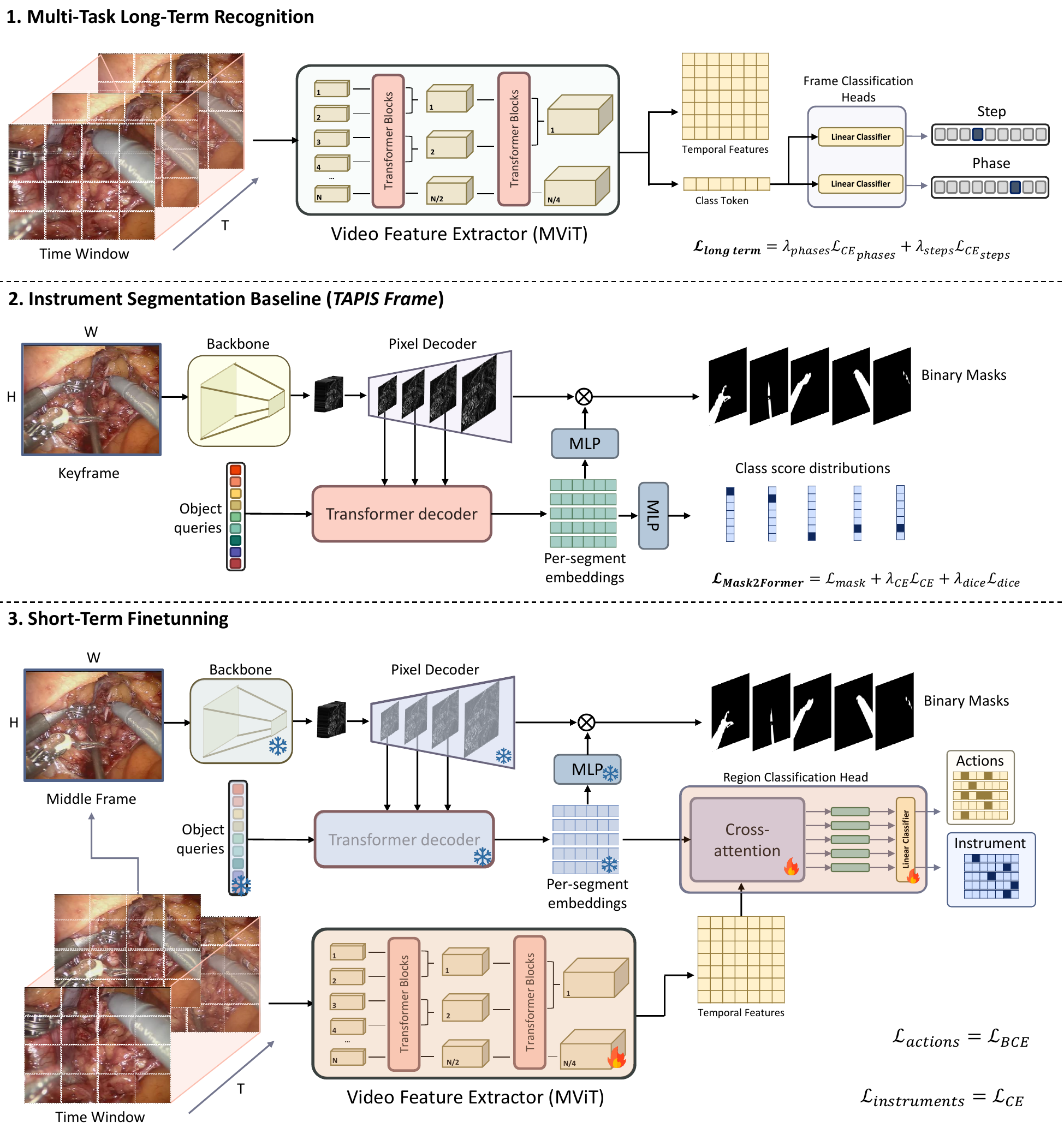}
\caption{\textbf{TAPIS training diagram.} We adopt a three-stage training process. First, we train our video feature extractor to address phase and step recognition jointly. We use a combined long-term loss ($\mathcal{L}_{long term}$), consisting of adding the cross-entropy loss ($\mathcal{L}_{CE}$) of each task multiplied by a per task coefficient ($\lambda$) We observed that the values of $\lambda_{phases}$ and $\lambda_{steps}$ had a very significant impact on the performance of each task and could significantly enhance the performance of one task while strongly reduce the other. However, we found $\lambda_{phases}=1$ and $\lambda_{steps}=0.5$ to achieve the best balance. Second, we independently train Mask2Former (our instrument segmentation baseline) for instrument type instance segmentation on separate keyframes using its original segmentation loss ($\mathcal{L}_{Mask2Former}$). Third, we used the pretrained weights of stages 1 and 2 and we finetuned the entire TAPIS architecture for instrument reclassification or atomic action detection. We froze the segmentation baseline and stored its precalculated region proposals and features to avoid training overhead. Even though we observed that calculating region proposals on the fly during training yielded better performance, it significantly increased training FLOPS and speed, so we opted to employ precalculated regions.}.
\label{fig:training_dragram}
\end{figure*}



\section{TAPIS performance in GraSP}
\label{sec:appendix:results}

\subsection{Dataset Splits for RARP45}
Given the absence of a public test set in the original RARP-45 dataset, we create our splits, using 27 videos for training and 9 videos for evaluation. To ensure the integrity of the splits, we maintain a consistent class distribution with the original training set. Specifically, for the test split, we select videos 2, 3, 4, 6, 7, 9, 14, 17, 23, 28, 32, and 33, while we use the remaining videos for the training split. 

\subsubsection*{Instrument segmentation}

\begin{figure*}[ht]
    \centering
    \includegraphics[width=\linewidth]{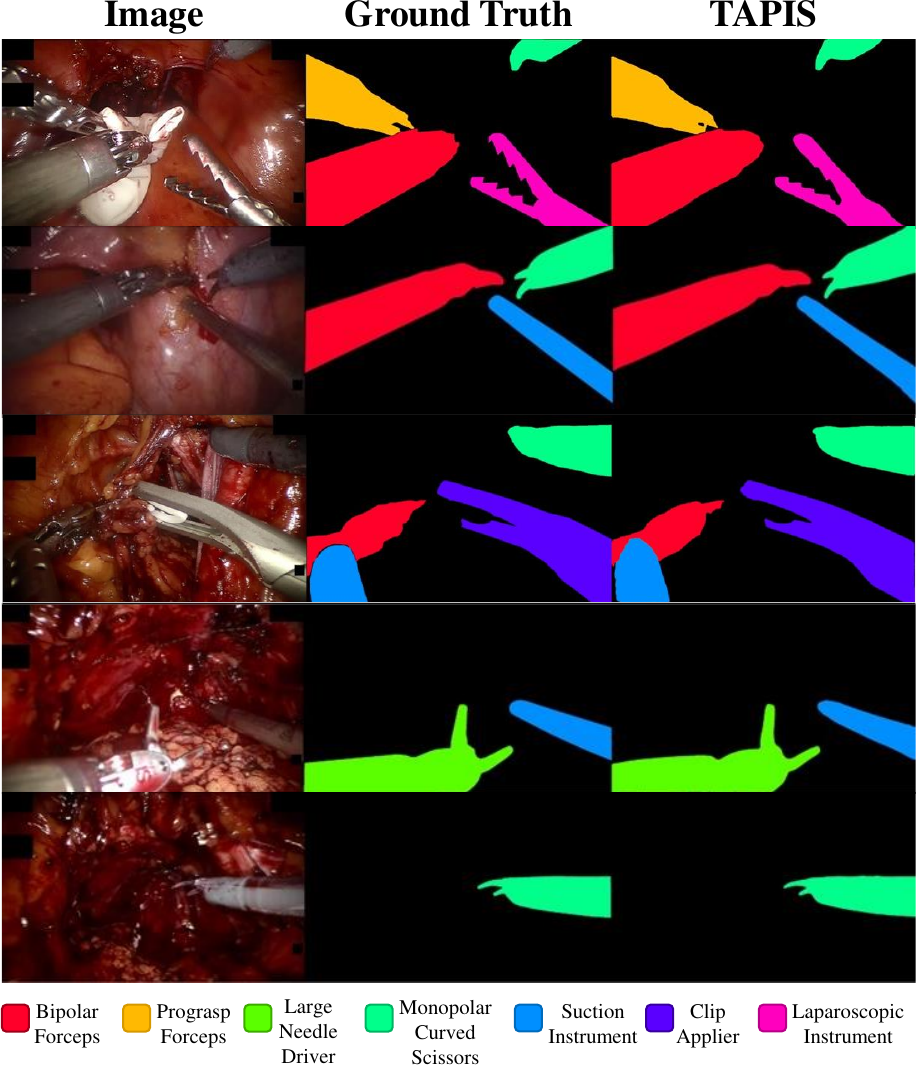}
    \caption{\textbf{Qualitative instance segmentation performance} comparison between TAPIS and corresponding ground truth. We show examples where TAPIS performs remarkably. The figure is best viewed in color.}
    \label{fig:visu_segm_excellent}
\end{figure*}

\begin{figure*}[ht]
    \centering
    \includegraphics[width=\linewidth]{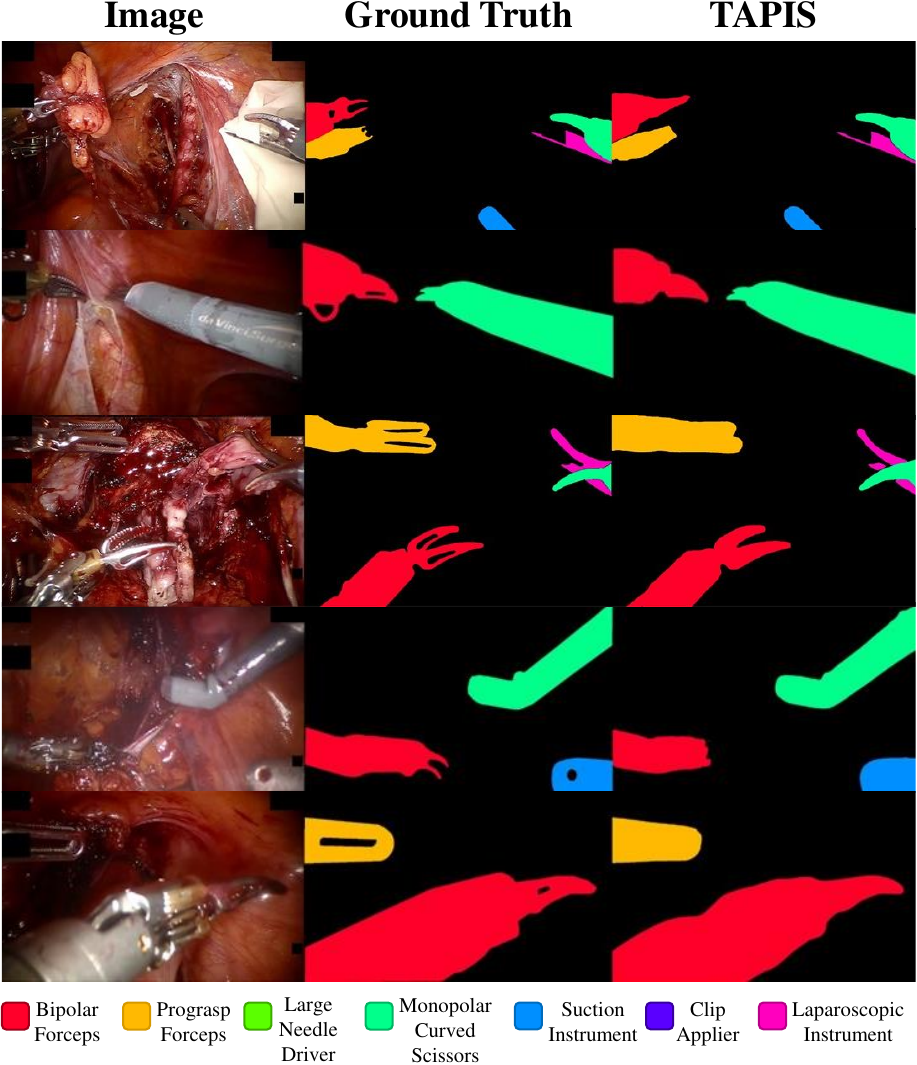}
    \caption{\textbf{Qualitative instance segmentation performance} comparison between TAPIS and corresponding ground truth. We show examples where TAPIS performs satisfactorily. The figure is best viewed in color.}
    \label{fig:visu_segm_good}
\end{figure*}

\begin{figure*}[ht]
    \centering
    \includegraphics[width=\linewidth]{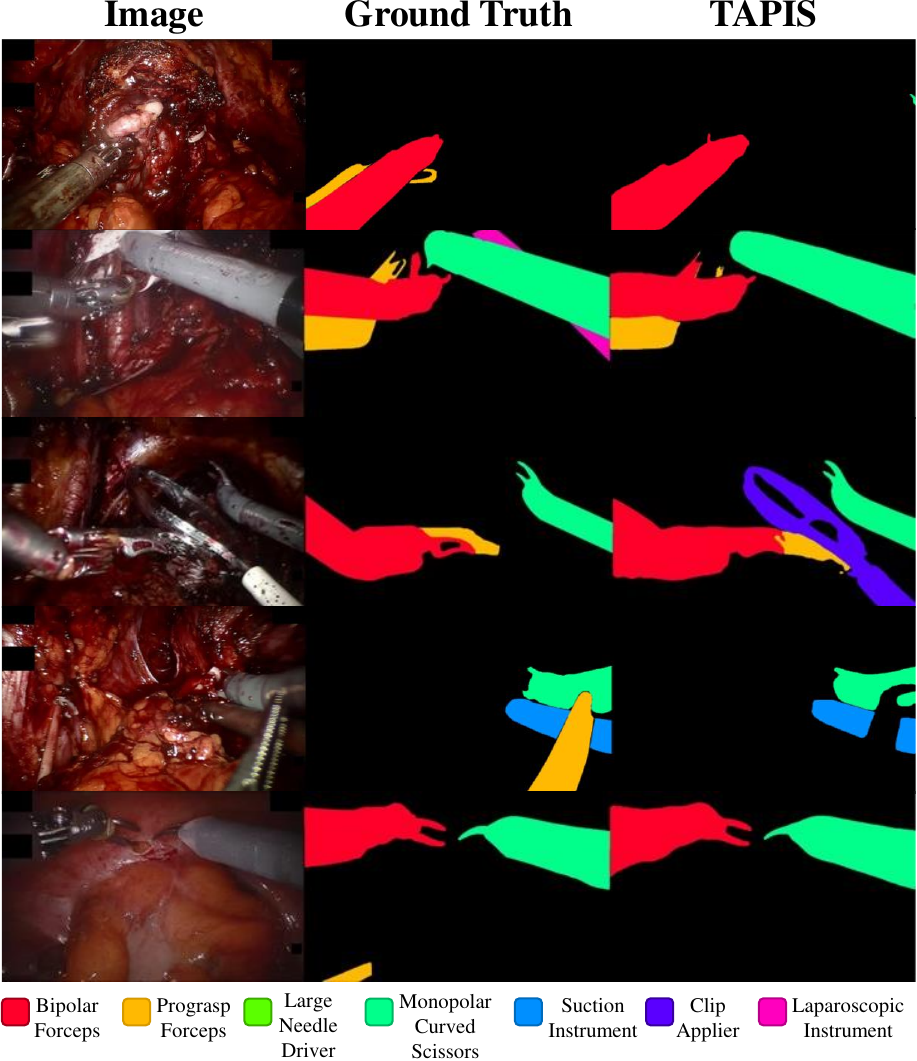}
    \caption{\textbf{Qualitative instance segmentation performance} comparison between TAPIS and corresponding ground truth. We show examples of TAPIS' error modes. The figure is best viewed in color.}
    \label{fig:visu_segm_bad}
\end{figure*}

Figures~\ref{fig:visu_segm_excellent}, \ref{fig:visu_segm_good}, and \ref{fig:visu_segm_bad} depict qualitative segmentation results from TAPIS on the GraSP dataset. As mentioned previously, TAPIS can perform accurate segmentation of complex images. In addition, Figure~\ref{fig:visu_segm_excellent} shows that TAPIS can segment instances of varying sizes, including bigger instrument arms and small tips. We show examples of inaccurate predictions in Figure~\ref{fig:visu_segm_bad}. We conduct an error mode analysis and observe cases where two instrument instances are predicted as one, given a high degree of overlap between them (see first row, upper left corner). Second, we observe cases where an instrument is classified as another, given a problematic view of the instrument. For example, the fifth row confuses Bipolar Forceps with Prograsp Forceps. These instruments are similar and difficult to distinguish from a profile view. 
Third, we observe cases where it wrongly identifies other objects as instruments. For instance, the third row identifies an instrument for recollecting waste as a Laparoscopic Instrument. Additionally, Tables~\ref{tab:class-instance-segmentation} and \ref{tab:class-semantic-segmentation} show the per-class results on instrument segmentation explained in the Instrument Segmentation Experiments Section.

\subsection*{Phase and step recognition}

\begin{figure*}[ht]
    \centering
    \includegraphics[width=\linewidth]{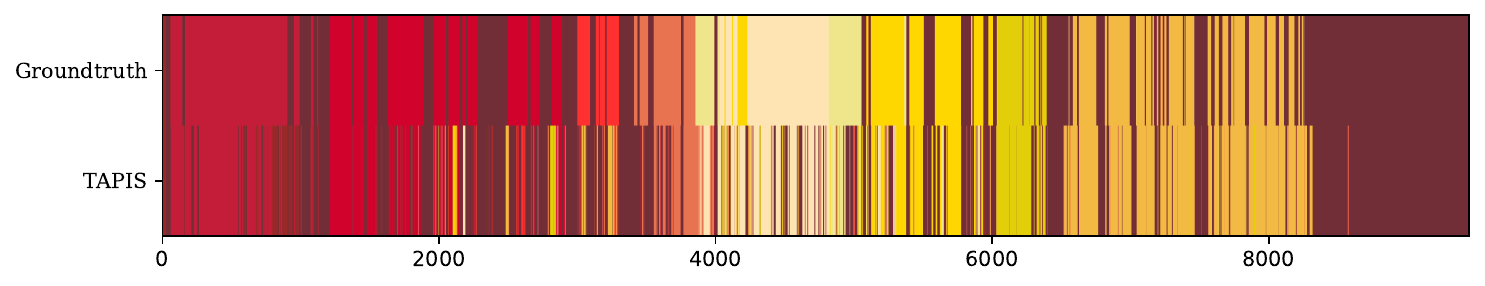}
    \caption{\textbf{Qualitative phase recognition performance} comparison between TAPIS and the Groundtruth in CASE041.}
    \label{fig:phases_TAPIS41}
\end{figure*}

\begin{figure*}[ht]
    \centering
    \includegraphics[width=\linewidth]{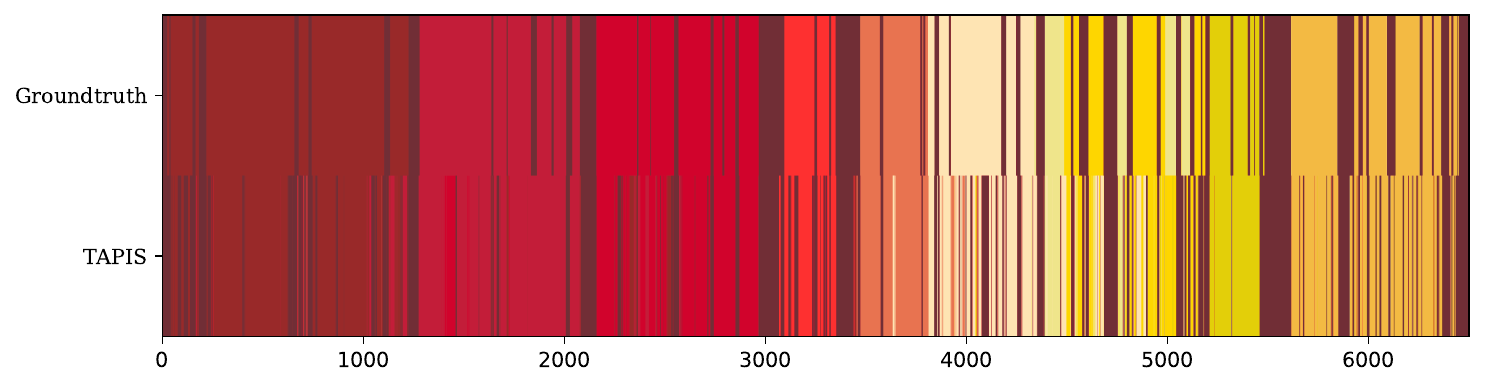}
    \caption{\textbf{Qualitative phase recognition performance} comparison between TAPIS and the Groundtruth in CASE047.}
    \label{fig:phases_TAPIS47}
\end{figure*}

On the one hand, Figures~\ref{fig:phases_TAPIS41} and~\ref{fig:phases_TAPIS47} present qualitative outcomes of TAPIS in the phase recognition task. Notably, our model encounters challenges in recognizing phases with shorter durations. In instances where recognition occurs, the model tends to identify longer time segments than their actual duration. Conversely, phases with longer durations exhibit more accurate recognition by the model.

\begin{figure*}[ht]
    \centering
    \includegraphics[width=\linewidth]{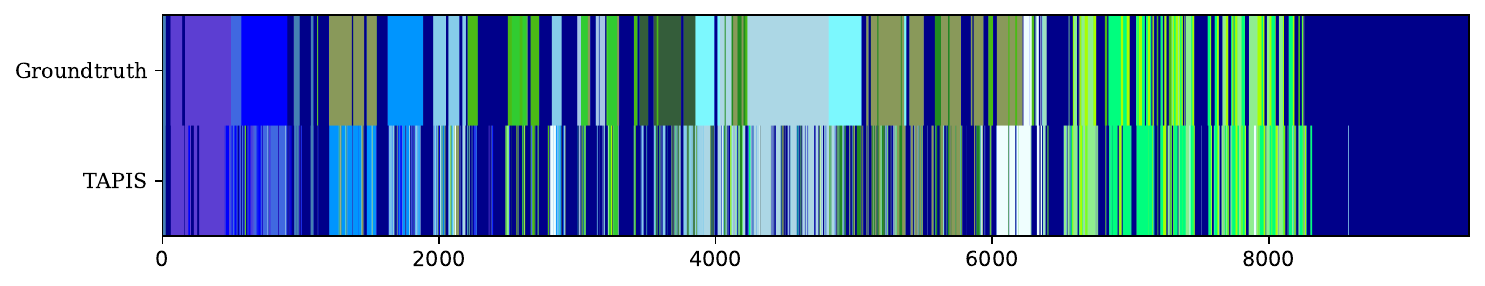}
    \caption{\textbf{Qualitative step recognition performance} comparison between TAPIS and the Groundtruth in CASE041.}
    \label{fig:steps_TAPIR41}
\end{figure*}

\begin{figure*}[ht]
    \centering
    \includegraphics[width=\linewidth]{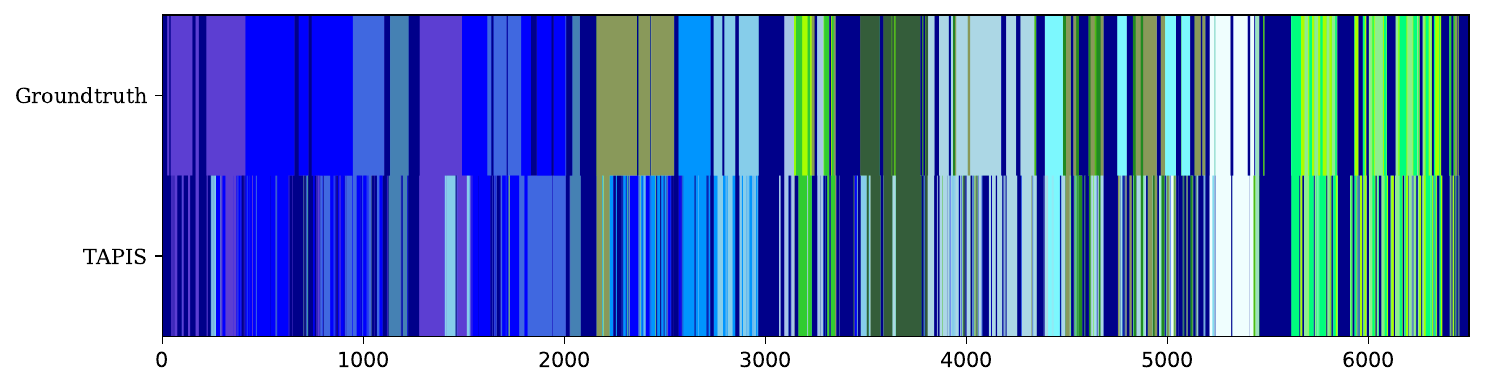}
    \caption{\textbf{Qualitative step recognition performance} comparison between TAPIS and the Groundtruth in CASE047.}
    \label{fig:steps_TAPIR47}
\end{figure*}

On the other hand, Figures~\ref{fig:steps_TAPIR41} and~\ref{fig:steps_TAPIR47} illustrate results in the steps recognition task. Notably, the model demonstrates noise in recognizing steps with longer durations while performing well in predicting short steps. Additionally, timestamps may experience shifts, indicating challenges in defining transitions between steps. Tables~\ref{tab:per_class_phase_performance} and \ref{tab:per_class_step_performance} present the per-class quantitative results on phases and steps recognition as explained in the Phases and Steps Experiments Section.

\subsection*{Action detection}

\begin{figure*}[ht]
    \centering
    \includegraphics[width=\linewidth]{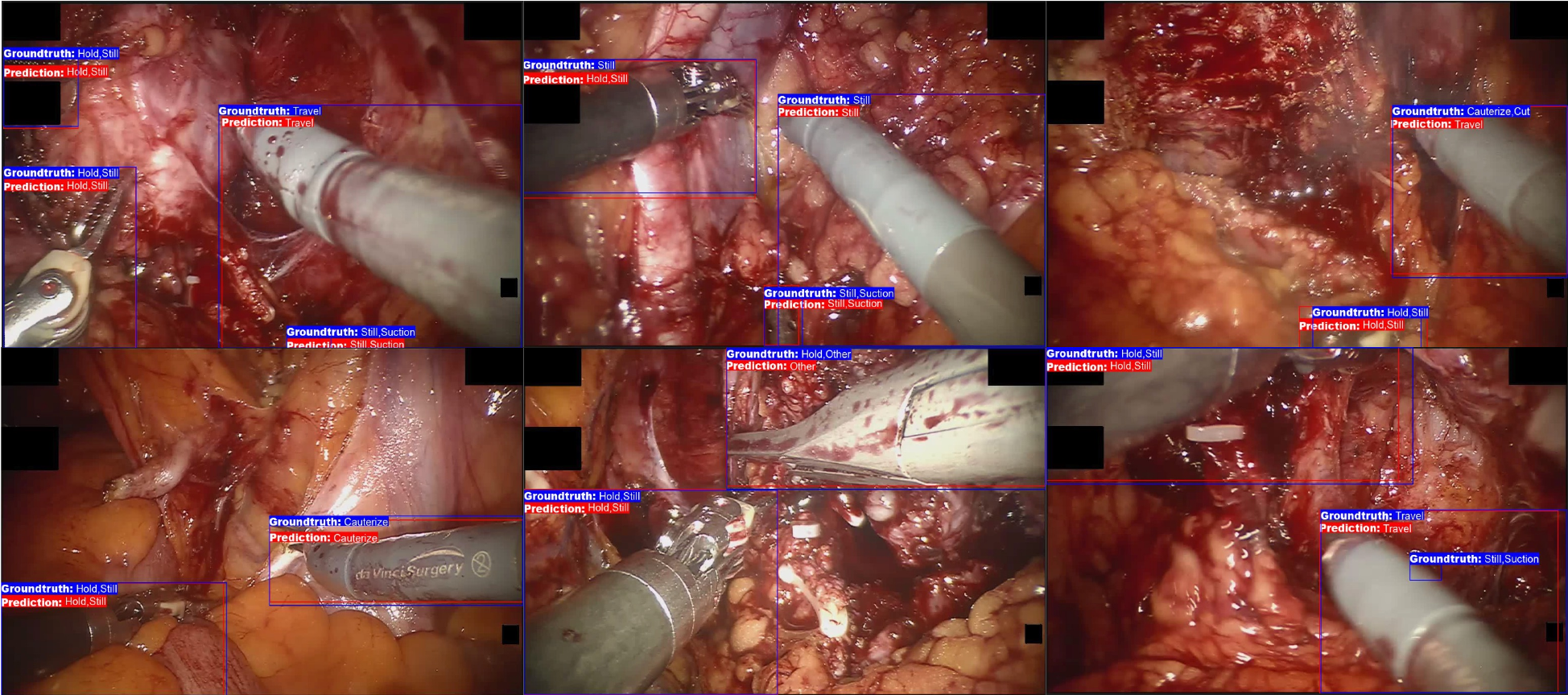}
    \caption{\textbf{Qualitative action detection performance} comparison between TAPIS and the Groundtruth. The Groundtruth boxes are marked in blue, while the ones predicted by our model are highlighted in red.}
    \label{fig:actions_qualitative}
\end{figure*}

The qualitative performance of TAPIS on the action detection task for the GraSP dataset is illustrated in Figure~\ref{fig:actions_qualitative}. The outcomes reveal consistent predictions across various action categories. TAPIS exhibits proficiency in accurately predicting multiple categories for a single instrument, showcasing its capability to comprehend surgical procedures in the short term. However, some inaccuracies are observed in the predictions, including instances of predicting incorrect categories, limiting predictions to a single category when the instrument executes multiple actions, and occasional failure to detect the instrument in the scene.

\begin{table*}[ht]
\caption{\textbf{Comparative semantic segmentation results of TAPIS in the Instrument Segmentation Task of GraSP}. Instrument categories are Bipolar Forceps (BF), Prograsp Forceps (PF), Large Needle Driver (LND), Monopolar Curved Scissors (MCS), Suction Instrument (SI), Clip Applier (CA), and Laparoscopic Grasper (LG).}
\centering
\resizebox{0.8\textwidth}{!}{
\begin{tabular}{lcccccccc}
\hline
\multirow{2}{*}{Method} & \multirow{2}{*}{$mAP@0.5IoU_{segm}$} & \multicolumn{7}{c}{Instrument Categories}                           \\ \cline{3-9} 
&                     & BF            & PF    & LND   & MCS   & SI    & CA    & LG      \\ \hline
ISINet                  & 78.29 $\pm 2.82$ & 87.58 & 63.71 & 90.31 & 96.17 & 87.14 & 73.01 & 50.10 \\
QPD Mask DINO           & 87.39 $\pm 1.75$ & 93.78 & 75.34 & 97.75 & 98.04 & 89.24 & 89.44 & 71.13 \\
TAPIS Frame (R50)       & 88.65 $\pm 1.43$ & 93.63 & 75.61 & 97.75 & 98.44 & 89.68 & 89.23 & 76.21 \\
TAPIS Full (R50)        & 87.20 $\pm 1.12$ & 93.25 & 72.73 & 97.76 & 98.48 & 88.73 & 86.64 & 72.81 \\
TAPIS Frame (SwinL)     & \textbf{91.71} $\pm 1.72$ & 95.44 & \textbf{80.60} & \textbf{99.00} & 98.73 & \textbf{92.53} & \textbf{91.27} & \textbf{84.42} \\
TAPIS Full (SwinL)      & 90.34 $\pm 1.11$ & \textbf{95.50} & 75.28 & 98.62 & \textbf{98.80} & 90.14 & 90.79 & 83.27 \\ \hline
\end{tabular}}
\label{tab:class-instance-segmentation}
\end{table*}

\begin{table*}[ht]
\caption{\textbf{Comparative semantic segmentation results of TAPIS in the Instrument Segmentation Task of GraSP}. Instrument categories are Bipolar Forceps (BF), Prograsp Forceps (PF), Large Needle Driver (LND), Monopolar Curved Scissors (MCS), Suction Instrument (SI), Clip Applier (CA), and Laparoscopic Grasper (LG).}
\centering
\resizebox{\textwidth}{!}{
\begin{tabular}{lccccccccccccc}
\hline
\multirow{2}{*}{Method} & \multirow{2}{*}{mIoU} & \multirow{2}{*}{IoU} & \multirow{2}{*}{cIoU} & \multicolumn{7}{c}{Instrument Categories}             \\ \cline{5-11} 
                        &                     &                       &                   & BF    & PF    & LND   & MCS   & SI    & CA    & LG    \\ \hline
TernausNet              & 41.74  $\pm 5.07$ & 24.46 $\pm 6.04$ & 16.87  $\pm 3.70$ & 29.24 & 7.28  & 6.60  & 50.20 & 24.69 & 0.00  & 0.04  \\
MF-TAPNet               & 66.63  $\pm 1.24$ & 29.23 $\pm 1.43$ & 24.98  $\pm 0.59$ & 48.35 & 9.95  & 12.97 & 67.68 & 28.68 & 1.80  & 5.46  \\
ISINet                  & 78.44  $\pm 1.13$ & 70.85 $\pm 0.00$ & 56.67  $\pm 1.46$ & 68.89 & 46.29 & 53.37 & 87.09 & 67.70 & 42.87 & 30.52 \\
QPD Mask DINO           & 83.89  $\pm 1.23$ & 82.56 $\pm 1.14$ & 74.36  $\pm 1.04$ & 81.60 & 58.10 & 89.06 & 91.80 & 76.77 & 68.27 & 54.90 \\
TAPIS Frame (R50)       & 84.81  $\pm 1.62$ & 81.34 $\pm 1.44$ & 73.48  $\pm 0.88$ & 80.61 & 55.08 & 87.04 & 91.85 & 74.58 & 70.09 & 55.09 \\
TAPIS Full (R50)        & 84.76  $\pm 1.63$ & 81.64 $\pm 1.45$ & 74.43  $\pm 0.79$ & 80.86 & 56.69 & 87.01 & 91.89 & 74.59 & 72.31 & 57.64 \\
TAPIS Frame (SwinL)     & 86.91  $\pm 1.59$ & 83.92 $\pm 0.68$ & 77.59  $\pm 0.08$ & 83.18 & 60.42 & 90.93 & 92.75 & 77.56 & 75.36 & \textbf{62.94} \\
TAPIS Full (SwinL)      & \textbf{87.05}  $\pm 1.63$ & \textbf{84.45} $\pm 0.72$ & \textbf{78.82}  $\pm 0.88$ & \textbf{83.92} & \textbf{62.00} & \textbf{91.18} & \textbf{92.90} & \textbf{77.96} & \textbf{78.50} & 65.26 \\ \hline
\end{tabular}}
\label{tab:class-semantic-segmentation}
\end{table*}

\begin{table*}[t]
\caption{\textbf{Comparative results of TAPIS in Atomic Action Detection}. Action categories are Cauterize (Cau), Close (Clo), Cut (Cut), Grasp (Gra), Hold (Hol), Open (Open), Open Something (O.Sm), Pull (Pull), Push (Pus), Release (Rel), Still (Sti), Suction (Suc), Travel (Tra) and Other (Oth).}
\centering
\resizebox{\textwidth}{!}{
\begin{tabular}{lcccccccccccccccc}
\hline
\multirow{2}{*}{Region Proposal Method} & \multirow{2}{*}{$mAP@0.5IoU_{box}$} & \multicolumn{14}{c}{Atomic Actions Categories}                                                              \\ \cline{3-16} 
                        &                     & Cau   & Clo   & Cut   & Gra   & Hol   & Ope  & O.Sm  & Pul   & Pus   & Rel  & Sti   & Suc   & Tra   & Oth   \\ \hline
Deformable DETR         & 28.54 $\pm 0.68$ & 59.70 & 12.44 & 29.15 & 9.56  & 53.45 & \textbf{7.95} & 3.67  & 11.60 & 27.95 & \textbf{6.75} & 59.24 & 51.03 & 58.61 & 8.44  \\
DINO                    & 31.37 $\pm 0.50$ & \textbf{60.97} & \textbf{15.73} & \textbf{30.06} & 12.03 & 53.28 & 5.52 & 20.60 & 15.08 & 27.74 & 5.54 & 57.47 & 47.72 & 61.88 & 25.61 \\
TAPIS Frame (R50)       & 31.24 $\pm 1.84$ & 54.94 & 14.00 & 18.78 & \textbf{13.04} & 66.20 & 3.56 & 9.08  & 15.20 & 28.74 & 2.18 & 71.37 & 57.07 & 60.64 & 22.57 \\
TAPIS Frame (SwinL)     & \textbf{35.46} $\pm 0.34$ & 60.26        & 13.93        & 22.43        & 13.77        & \textbf{71.64}        & 5.31         & \textbf{22.02}        & \textbf{17.60}        & \textbf{31.46}        & 5.69         & \textbf{76.80}        & \textbf{65.97} & \textbf{64.27}        & \textbf{27.51}        \\ \hline
\end{tabular}}
\label{tab:class-action-detection}
\end{table*}

\begin{table*}[ht]
\centering
\caption{\textbf{Comparative results of TAPIS in the Phase Recognition Task of GraSP}. Phase categories are Idle (0), Left pelvic isolated lymphadenectomy (1), Right pelvic isolated lymphadenectomy (2), Developing the Space of Retzius (3), Ligation of the deep dorsal venous complex (4), Bladder neck identification and transection (5), Seminal vesicle dissection (6), Development of the plane between the prostate and rectum (7), Prostatic pedicle control (8), Severing of the prostate from the urethra (9), and Bladder neck reconstruction (10).}
\resizebox{\textwidth}{!}{
\begin{tabular}{c|cccccccccccc}
\hline
\textbf{Method}  & \textbf{mAP (\%)} & \textbf{0} & \textbf{1} & \textbf{2} & \textbf{3} & \textbf{4} & \textbf{5} & \textbf{6} & \textbf{7} & \textbf{8} & \textbf{9} & \textbf{10} \\
\hline
SlowFast          & $68.44 \pm 0.10$ & $72.89$ & $67.83$ & $66.63$ & 80.53 & $82.37$ & $75.49$ & $72.59$ & \textbf{29.30} & $56.29$ & $71.78$ & $75.02$ \\
TAPIS-VST         & $70.60 \pm 1.59$ & $69.15$ &\textbf{81.23} & \textbf{78.08} & $76.49$ & $79.45$ & \textbf{80.84} & \textbf{73.73} & $26.80$ & $54.27$ & $77.17$ & \textbf{79.44} \\
TAPIS             & $\textbf{71.36} \pm 1.33$ & \textbf{73.16} & $79.86$ & $73.57$ & $\textbf{81.76}$ & \textbf{86.91} & $78.84$ & $71.88$ & $24.76$ & \textbf{56.66} & \textbf{80.42} & $77.11$ \\
\hline
\end{tabular}
}
\label{tab:per_class_phase_performance}
\end{table*}

\begin{table*}[ht]
\centering
\caption{\textbf{Comparative results of TAPIS in the Step Recognition Task of GraSP}. Phase categories are Idle (0), Identification and dissection of the Iliac vein and artery (1), Cutting and dissection of the external iliac vein’s lymph node (2), Obturator nerve and vessel path identification, dissection, and cutting of the obturator lymph nodes (3), Insert the lymph nodes in retrieval bags (4), Prevessical dissection (5), Ligation of the dorsal venous complex (6), Prostate dissection until the levator ani (7), Seminal vesicle dissection (8), Dissection of Denonviliers’ fascia (9), Cut the tissue between the prostate and the urethra (10), Hold prostate (11), Insert prostate in retrieval bag (12), Pass suture to the urethra (13), Pass suture to the bladder neck (14), Pull suture (15), Tie suture (16), Suction (17), Cut suture or tissue (18), Cut between the prostate and bladder neck (19), and Vascular pedicle control (20). }
\resizebox{\textwidth}{!}{%
\begin{tabular}{c|cccccccccccccccccccccc}
\hline
\textbf{Method}  & \textbf{mAP (\%)} & \textbf{0} & \textbf{1} & \textbf{2} & \textbf{3} & \textbf{4} & \textbf{5} & \textbf{6} & \textbf{7} & \textbf{8} & \textbf{9} & \textbf{10} & \textbf{11} & \textbf{12} & \textbf{13} & \textbf{14} & \textbf{15} & \textbf{16} & \textbf{17} & \textbf{18} & \textbf{19} & \textbf{20} \\
\hline
SlowFast          & $43.86 \pm 1.79$ & $70.15$ & $63.50$ & $30.73$ & $38.93$ & $79.18$ & $52.68$ & $29.61$ & $9.60$ & $65.41$ & $22.56$ & $81.62$ & $1.10$ & $54.41$ & $38.30$ & $44.37$ & $29.31$ & $59.94$ & $7.90$ & $36.15$ & $58.09$ & $47.62$ \\
TAPIS-VST         & $47.17 \pm 2.65$ & $70.34$ & $57.01$ & $31.36$ & $31.64$ & $78.92$ & $55.98$ & $45.20$ & $\textbf{11.39}$ & $\textbf{74.97}$ & $26.01$ & $\textbf{81.68}$ & $\textbf{7.22}$ & $79.52$ & $36.12$ & $45.29$ & $33.58$ & $49.52$ & $8.30$ & $38.56$ & $\textbf{79.23}$ & $48.98$ \\
TAPIS             & $\textbf{50.74} \pm 2.53$ & $\textbf{73.11}$ & $\textbf{66.26}$ & $\textbf{36.36}$ & $\textbf{37.93}$ & $\textbf{80.85}$ & $\textbf{67.14}$ & $\textbf{46.68}$ & $10.11$ & $71.81$ & $\textbf{27.10}$ & $78.24$ & $0.51$ & $\textbf{89.81}$ & $\textbf{37.9}3$ & $\textbf{46.99}$ & $\textbf{40.47}$ & $\textbf{67.88}$ & $\textbf{11.70}$ & $\textbf{44.23}$ & $76.47$ & $\textbf{53.35}$ \\
\hline
\end{tabular}%
}
\label{tab:per_class_step_performance}
\end{table*}

\end{document}